\documentclass[10pt,twocolumn,letterpaper]{article}

\usepackage[pagenumbers]{cvpr}

\usepackage{multirow}
\usepackage{color, colortbl}
\usepackage{xcolor,amsmath}
\usepackage{amssymb}
\usepackage{pifont}
\usepackage{paralist}
\usepackage{ulem}
\usepackage{diagbox}
\usepackage{lscape}
\usepackage{pdflscape}
\usepackage{rotating}
\usepackage{enumitem}
\usepackage[toc,page]{appendix} 
\usepackage{minitoc} 
\usepackage{tocloft}

\newcommand{\toccolor}{\hypersetup{linkcolor=black}}
\newcommand{\defaultcolor}{\hypersetup{linkcolor=red}}
\definecolor{emerald}{rgb}{0.31, 0.78, 0.37}
\definecolor{coralred}{rgb}{1.0, 0.25, 0.25}
\definecolor{mygray}{rgb}{0.906, 0.902, 0.902}

\definecolor{cvprblue}{rgb}{0.21,0.49,0.74}
\usepackage[pagebackref,breaklinks,colorlinks,allcolors=cvprblue]{hyperref}
\usepackage{amsmath}
\usepackage{amsfonts}
\usepackage{overpic}

\def\paperID{***} 
\def\confName{CVPR}
\def\confYear{2025}

\title{ALoRE: Efficient Visual Adaptation via Aggregating Low Rank Experts}

\author{
Sinan Du$^{1,2}$\thanks{Equal contribution.~~\textsuperscript{\dag}Corresponding authors.}\quad Guosheng Zhang$^{2*}$\quad Keyao Wang$^{2}$\quad Yuanrui Wang$^{1}$\quad Haixiao Yue$^{2}$ \\
Gang Zhang$^{2}$\quad Errui Ding$^{2}$\quad Jingdong Wang$^{2}$\quad Zhengzhuo Xu$^{1\dagger}$\quad Chun Yuan$^{1\dagger}$ \\
\textsuperscript{1} Tsinghua University~~\textsuperscript{2}Department of Computer Vision Technology (VIS), Baidu Inc
}

\begin{document}
\maketitle
\begin{abstract}
Parameter-efficient transfer learning (PETL) has become a promising paradigm for adapting large-scale vision foundation models to downstream tasks. Typical methods primarily leverage the intrinsic low rank property to make decomposition, learning task-specific weights while compressing parameter size. However, such approaches predominantly manipulate within the original feature space utilizing a single-branch structure, which might be suboptimal for decoupling the learned representations and patterns. In this paper, we propose \textbf{ALoRE}, a novel PETL method that reuses the hypercomplex parameterized space constructed by Kronecker product to \underline{\textbf{A}}ggregate \underline{\textbf{Lo}}w \underline{\textbf{R}}ank \underline{\textbf{E}}xperts using a multi-branch paradigm, disentangling the learned cognitive patterns during training. Thanks to the artful design, ALoRE maintains negligible extra parameters and can be effortlessly merged into the frozen backbone via re-parameterization in a sequential manner, avoiding additional inference latency. We conduct extensive experiments on \textbf{24} image classification tasks using various backbone variants. Experimental results demonstrate that ALoRE outperforms the full fine-tuning strategy and other state-of-the-art PETL methods in terms of performance and parameter efficiency. For instance, ALoRE obtains \textbf{3.06\%} and \textbf{9.97\%} Top-1 accuracy improvement on average compared to full fine-tuning on the FGVC datasets and VTAB-1k benchmark by only updating \textbf{0.15M} parameters.
\end{abstract}

\addtocontents{toc}{\protect\setcounter{tocdepth}{-1}} 
\section{Introduction}
\label{sec:intro}
Vision foundation models~\cite{vit,mae,swin,swinv2,convnet,mlp-mixer,deit,scaling} pre-trained on massive datasets have been demonstrated to possess powerful feature extraction and modeling capabilities, which can be directly generalized to downstream tasks through fine-tuning. Nevertheless, full fine-tuning poses two significant challenges: (1) Updating the parameters of the entire model requires expensive computational resources, time costs, and sufficiently large datasets to avoid overfitting; (2) The storage overhead caused by the enormous number of model parameters makes it impractical to save a completely new set of model weights for each downstream task. A simple solution is linear probing~\cite{moco}, which only updates the last head layer but yields inferior performance compared to the full fine-tuning strategy.

\begin{figure}[t]
    \centering
    \includegraphics[width=0.96\linewidth]{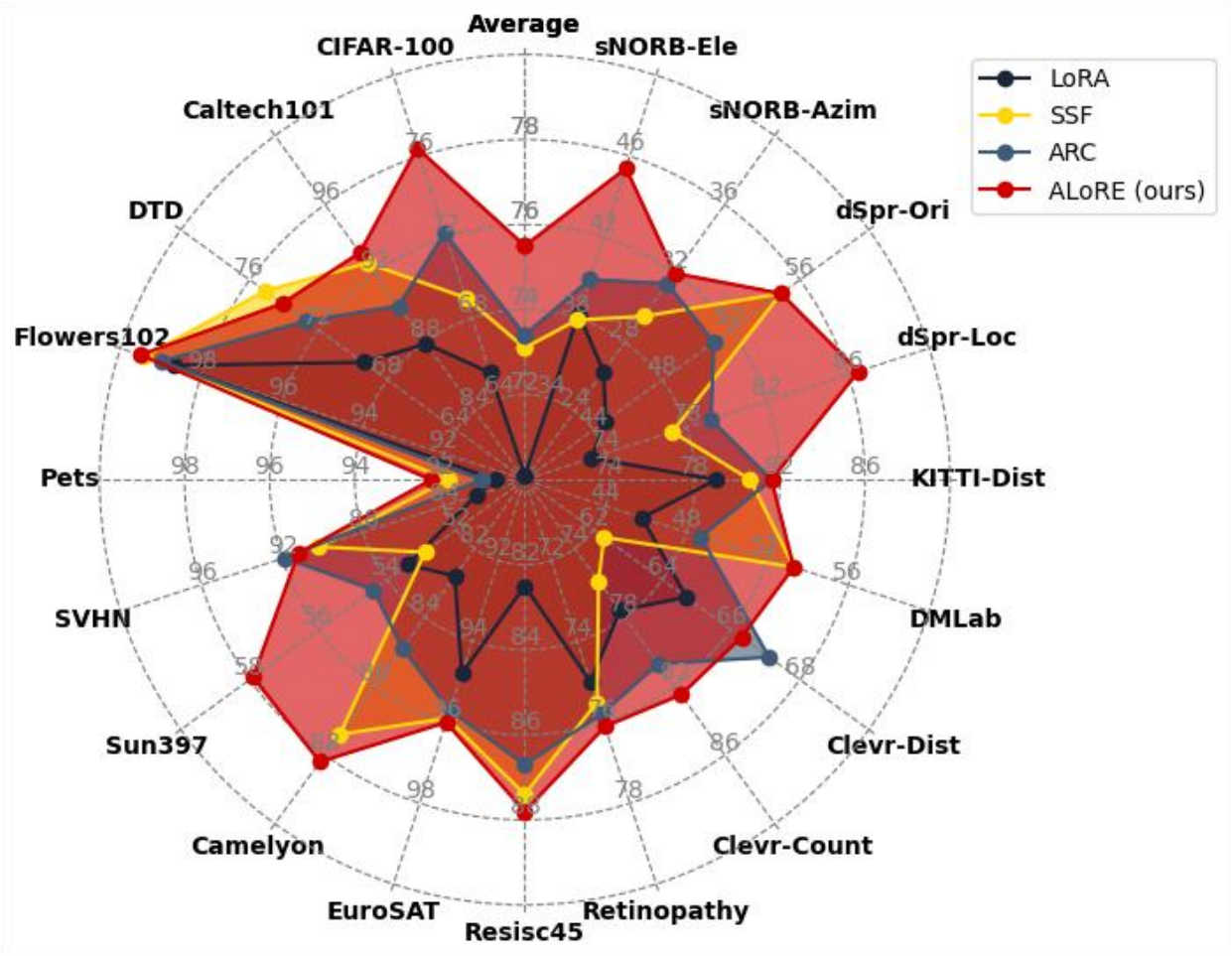}
    \caption{We present the comparisons of performance with other existing PETL methods with the ViT-B/16 model on the VTAB-1k benchmark. Our ALoRE achieves state-of-the-art Top-1 accuracy (\%) on average in a broad range of 19 downstream tasks.}
    \label{fig:radar}
    \vspace{-15pt}
\end{figure}

To strike a balance between performance and computation overhead, recent research efforts have focused on parameter-efficient transfer learning (PETL), which updates a minimal number of parameters to achieve comparable or even superior performance to full fine-tuning. Inspired by the successful practice in NLP~\cite{compacter,adapter,lora,power,prefix,autoprompt}, visual adapter-based and prompt-based fine-tuning methods are transferred to computer vision. Adapter-based methods~\cite{adaptformer,compacter,adapter,restuning} leverage a non-linear bottleneck to reduce the dimension of the representation into a lower rank. Prompt-based methods~\cite{vpt} insert learnable soft prompts to interact with visual tokens through the self-attention mechanism. LoRA~\cite{lora} inserts trainable low rank matrices to approximate the updates of weights. However, typical methods (Adapter-based~\cite{adaptformer,arc,adapter,restuning} and LoRA-like~\cite{lora,adalora}) predominantly capitalize on the inherent low rank property~\cite{aghajanyan2020intrinsic,li2018measuring} to conduct decomposition within the \textbf{original dense space} with a \textbf{single-branch} module, which might impede the purification of acquired representations and the decoupling of feature patterns, thus restricting the performance of visual adaptation tasks. \textit{Considering that a visual concept can encompass various observation perspectives, such as geometric shape, color, texture, topology, and semantics, we aim to innovatively disentangle the learned cognitive patterns within the PETL modules.}

Motivated by the multi-head attention mechanism in Transformer~\cite{transformer} to \textit{disentangle modes without introducing explicit constraints or extra parameters}, ALoRE leverages Kronecker product~\cite{wen2020batchensemble,zhang2021beyond} to construct a \textbf{hypercomplex parameterized space} while concurrently reusing the new feature space and performing low rank decompositions, thereby establishing a \textbf{multi-branch} structure for aggregating multiple low rank experts. With the design of multiplexing, the parameter increment resulting from the number of branches is \textbf{negligible}. Moreover, we dismiss the non-linear component in the original adapter~\cite{adapter}, allowing ALoRE to perform re-parameterization~\cite{repvgg} in a \textbf{sequential} (unlike \textit{parallel} in LoRA~\cite{lora}) manner.

\begin{figure}[t]
    \centering
    \includegraphics[width=0.96\linewidth]{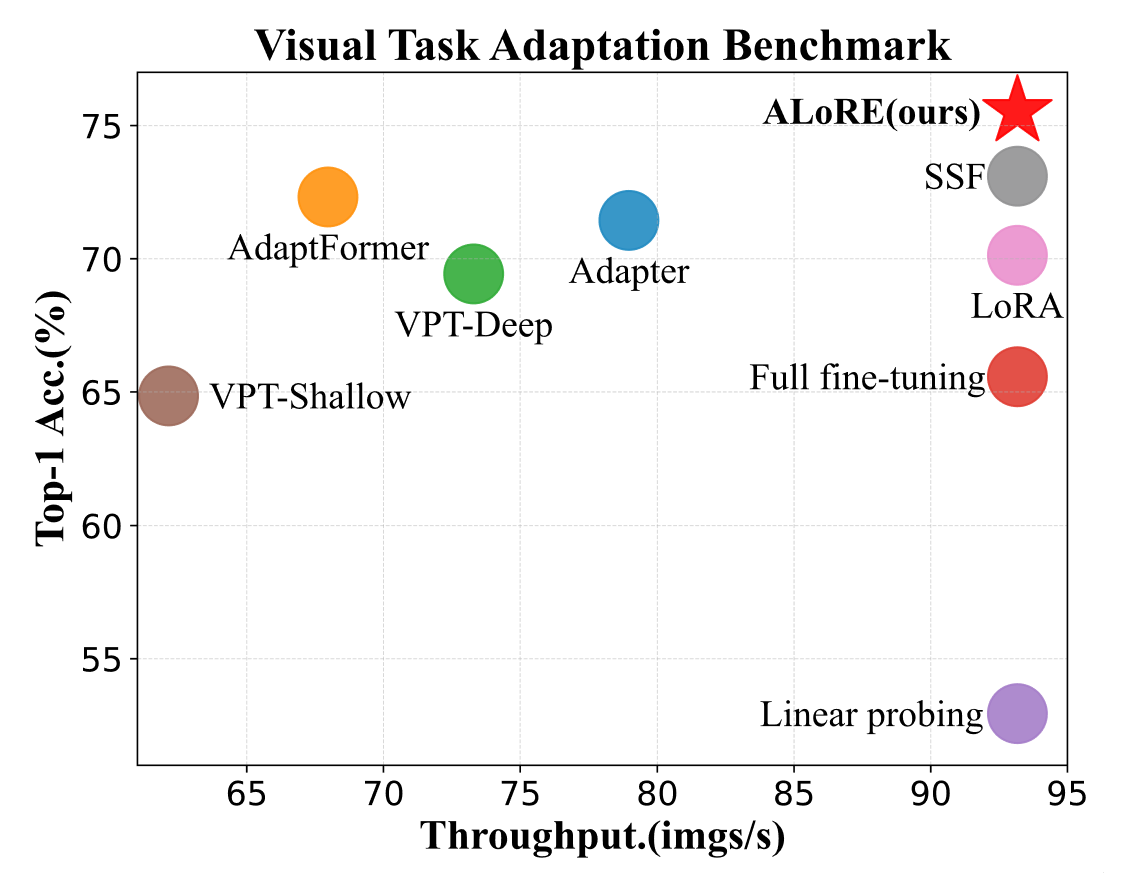}
    \caption{The comparisons of performance and throughput during inference. Our ALoRE achieves the theoretical optimal throughput by maintaining the desirable property of re-parameterization.}
    \label{fig:throughput}
    \vspace{-10pt}
\end{figure}

As illustrated in Figure~\ref{fig:radar} and~\ref{fig:throughput}, ALoRE achieves state-of-the-art performance and optimal throughput with the assistance of the desirable property of re-parameterization. 

Our key contributions are summarized as follows:
\begin{compactitem}
    \item We revisit the topic of visual adaptation from a novel perspective of disentanglement motivated by the multi-head attention mechanism, differing from prevalent methods that focus on designing specific structures.
    \item Diverging from the prevailing approaches of decomposition within the primitive dense space (\textit{single-branch}), our proposed ALoRE reuses the hypercomplex parameterized space constructed by the Kronecker product to aggregate low rank experts in a \textit{multi-branch} manner, which concurrently prevents a linear increase in parameter size with the number of branches.
    \item We dismiss the non-linear component in the original adapter, formulating a new solid baseline for re-parameterizable methods, which is free of additional latency during inference.
    \item We conduct extensive experiments on 24 downstream tasks using various backbone variants, demonstrating that our method surpasses the full fine-tuning strategy and other PETL methods in performance, parameter efficiency, and scalability.
\end{compactitem}
\section{Related Work} 
\label{sec:related}

\subsection{Parameter-Efficient Transfer Learning} 
\label{sec:petl}
As the explosion in model size~\cite{vit,swin,swinv2,deit} incurs an unbearable computation cost when fine-tuning the whole model, PETL for large language models (LLMs) has been initially explored in the field of NLP~\cite{adapter}, which demonstrates that comparable or even superior performance can be achieved by only updating few trainable parameters. Inspired by the successful practice of NLP~\cite{compacter,adapter,lora,prefix,pre,autoprompt}, numerous profound works~\cite{adaptformer,arc,vpt,restuning,convpass,ssf,rep,losa,bitfit,noah} have emerged in the computer vision community. Adapter-based methods insert a non-linear bottleneck structure~\cite{adaptformer,adapter,restuning} with a residual~\cite{resnet} connection into the backbone to learn task-specific representations, mainly focusing on the placement of adapters~\cite{adaptformer,adapter,restuning} or the special architecture design~\cite{conv-adapter,convpass,mona,conv-lora} for vision tasks. VPT~\cite{vpt} proposes to insert soft trainable prompts into the input space, which interact with visual tokens by the self-attention module. Res-Tuning-Bypass~\cite{restuning} leverages a ladder-side bypass~\cite{lst} to enable the objective of memory efficiency at the cost of performance. Compacter~\cite{compacter} is the pioneering work to introduce hypercomplex layers~\cite{wen2020batchensemble,zhang2021beyond} to replace trivial linear layers in the adapter, extremely compressing the parameter size. Regarding re-parameterizable PETL methods, LoRA~\cite{lora} approximates the updates of weights through training low rank matrices, which are parallel to the backbone module and can be merged during inference. ARC~\cite{arc}, SSF~\cite{ssf} and RepAdapter~\cite{rep} are capable of re-parameterizing the sequential structures. Differing from them in perspective, the motivation of ALoRE originates from disentangling the learned cognitive patterns using a multi-branch structure to aggregate low rank experts.

\subsection{Model Re-parameterization}
\label{sec:rep}
Model re-parameterization~\cite{acnet,repmlp,dbbnet,repvgg,repmlpnet,diracnets} has become a commonly adopted technique for designing efficient deep neural networks. RepVGG~\cite{repvgg} merges convolutions of different kernel sizes in a multi-branch structure into a single module, significantly reducing the computational cost during inference. Inspired by RepVGG~\cite{repvgg}, ACNet~\cite{acnet}, RepMLPNet~\cite{repmlpnet}, and DBBNet~\cite{dbbnet} expand the capacity of models through structural re-parameterization. Similarly, ALoRE only exploits linear transformation to interact with other frozen modules in the backbone sequentially, which can be merged through re-parameterization.
\begin{figure*}[t!]
    \centering
    \includegraphics[width=0.8\textwidth]{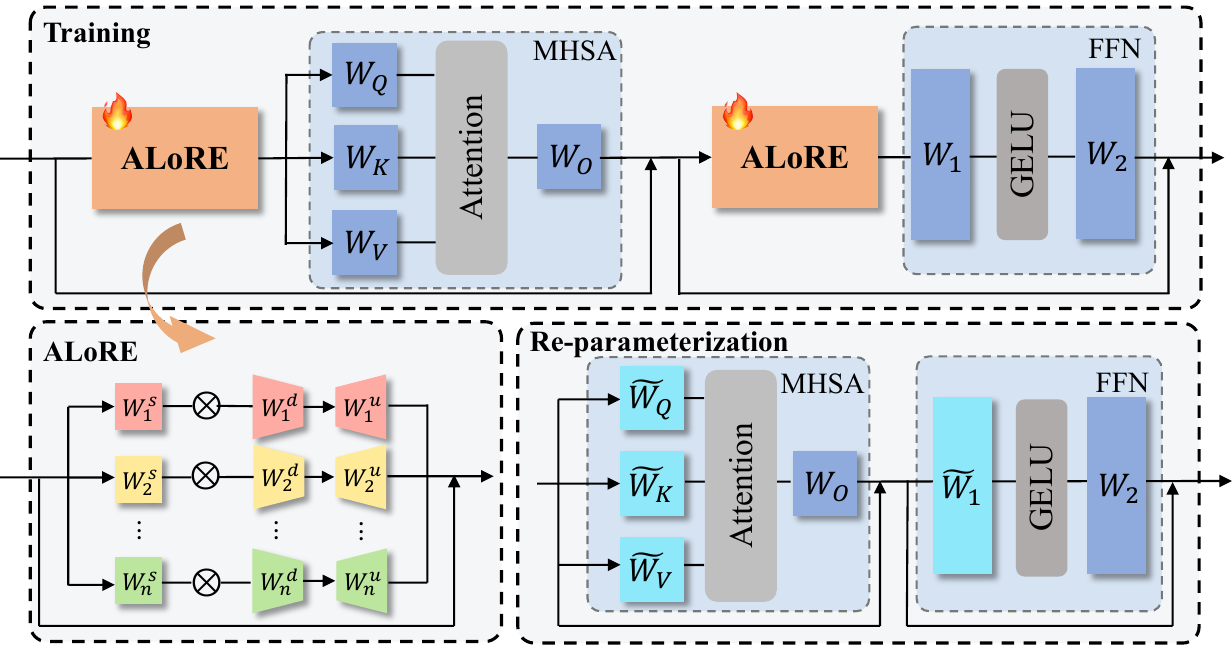}
    \caption{Illustration of Aggregated Low Rank Experts method. The ALoRE block can be reduced to a simple linear transformation layer after training. Afterward, the reduced linear weights can be re-parameterized into the first projection layer of the nearest module.}
    \label{fig:method}
    \vspace{-10pt}
\end{figure*}

\section{Method}
\label{sec:method}

\subsection{Preliminaries}
\label{sec:preliminaries}

\textbf{Kronecker Product.} The Kronecker product between matrix $A \in \mathbb{R}^{m\times n}$ and $B \in \mathbb{R}^{p\times q}$, denoted by $A \otimes B \in \mathbb{R}^{mp\times nq}$, can be formulated as:
\begin{equation}
\small
    A \otimes B = \begin{pmatrix}
        a_{11}B & \cdots & a_{1n}B \\
        \vdots & \ddots & \vdots \\
        a_{m1}B & \cdots & a_{mn}B \\
    \end{pmatrix},
\end{equation}
where $a_{ij}$ denotes the element positioned at the $i$-th row and $j$-th column of matrix $A$.

\hspace{-0.4cm}\textbf{Vision Transformer.} The plain Vision Transformer (ViT)~\cite{vit} is composed of a patch embedding layer and $L$ stacked transformer encoder blocks. Given an RGB image $I \in \mathbb{R}^{3 \times H \times W}$, the patch embedding layer initially divides it into $N\times N$ non-overlapping image patches, and then these patches are flattened as visual tokens with a learnable classification token prepended and positional embedding added. Afterward, the input tokens $X_0 \in \mathbb{R}^{(N^2+1)\times d}$, where $d$ is the hidden dimension, are fed into subsequent $L$-layer encoder blocks, each consisting of LayerNorm (LN), Multi-Head Self-Attention (MHSA) and Feed-Forward Network (FFN). MHSA is leveraged to capture the interrelationships among diverse image regions, which serves as a fundamental component within transformer architectures. The forward process of the $l$-th encoder block can be defined as:
\begin{equation}
\small
\begin{split}
    &{X_l^\prime} = {\rm MHSA}({\rm LN}({X_{l-1}})) + {X_{l-1}}, \\
    &{X_l} = {\rm FFN}({\rm LN}({X_l^\prime})) + {X_l^\prime},
\end{split}
\end{equation}
where $X_l$ denotes the output of the $l$-th block, $X_{l-1}$ indicates the input tokens of the $l$-th layer and the output of previous layers, $X_l^\prime$ denotes the intermediate representations.

\hspace{-0.4cm}Specifically, MHSA can be formulated as:
\begin{equation}
\small
    \begin{split}
        &{\rm MHSA}(Q,K,V) = [{\rm head}_1,\ \cdots,\ {\rm head}_{n_h}]W^O, \\
        &{\rm head}_i = {\rm Attention}(QW_i^Q, KW_i^K, VW_i^V), \\
        &{\rm Attention}(Q,K,V) = {\rm softmax}(\frac{QK^T}{\sqrt{d_k}})V,
    \end{split}
\end{equation}
where [·,·] denotes concatenation operation, $W_i^Q \in \mathbb{R}^{d\times \frac{d}{n_h}}$, $W_i^K \in \mathbb{R}^{d\times \frac{d}{n_h}}$, $W_i^V \in \mathbb{R}^{d\times \frac{d}{n_h}}$ and $W^O \in \mathbb{R}^{d\times d}$ are the projection matrices, $n_h$ means the number of heads, $d_k$ indicates the reduced dimension of each head. The FFN consists of a two-layer MLP and can be defined as:
\begin{equation}
\small
    {\rm FFN}(X) = {\rm GELU}(XW_1+b_1)W_2 + b_2,
\end{equation}
where $W_1 \in \mathbb{R}^{d\times 4d}$ and $W_2 \in \mathbb{R}^{4d\times d}$ are linear projections, $b_1 \in \mathbb{R}^{4d}$ and $b_2 \in \mathbb{R}^{d}$ are bias items.

\hspace{-0.4cm}\textbf{Visual Adapter.} Visual adapter~\cite{adapter} is a bottleneck structure with a non-linear activation function and residual connection with few trainable parameters
. It can be defined as:
\begin{equation}
\small
    f(X;\theta)=X + \sigma(XW^d + b^d)W^u + b^u,
\end{equation}
where $W^d \in \mathbb{R}^{d\times r}$ ($r << d$), $\sigma$, and $W^u \in \mathbb{R}^{r\times d}$ denote the down-projection, non-linear activation function and up-projection, $b^d \in \mathbb{R}^r$ and $b^u \in \mathbb{R}^d$ represent bias items.

\subsection{Aggregating Low Rank Experts}

\textbf{The design of ALoRE.} \textit{A straightforward approach to aggregate low rank experts would be to duplicate the original non-linear visual adapter multiple times in parallel within the backbone to formulate a multi-branch structure. However, it leads to a linear increase in the parameter size with the number of branches and introduces additional computational overhead during inference.} Therefore, we seek the assistance of the Kronecker product mentioned in Section~\ref{sec:preliminaries} to circumvent the restrictions imposed by the direct superposition. The architecture of ALoRE is illustrated in Figure~\ref{fig:method}. Starting from a simple linear transformation $W^\prime \in \mathbb{R}^{d\times d}$, we leverage Kronecker product to decompose it into a scale weight matrix $W_i^s \in \mathbb{R}^{n\times n}$ and a hypercomplex weight matrix $W_i^e \in \mathbb{R}^{\frac{d}{n}\times \frac{d}{n}}$. Previous work~\cite{chung2020rethinking,zhang2020revisiting} observes that redundant features are introduced into the pre-trained model and sharing adapters across layers~\cite{adapterfusion,adapterdrop} contributes to a negligible performance degradation for transfer learning. Motivated by these findings, the scale weight matrix $W_i^s$ is shared by all experts of the same index across different layers to capture the common patterns useful for adapting to the target task. Leverage the intrinsic low rank property~\cite{aghajanyan2020intrinsic,li2018measuring} in the hypercomplex space, the hypercomplex weight matrix $W_i^e$ is then subjected to low rank decomposition to obtain multiple adapters. $W^\prime$ is learned by a sum of Kronecker products. To achieve module re-parameterization, we omit the non-linear component in the previous visual adapter~\cite{adaptformer,adapter} and avoid extra inference latency compared to the original backbone. The process of decomposition can be defined as:
\begin{equation}
\small
    W^\prime = \sum_{i=1}^nW_i^s\otimes W_i^e=\sum_{i=1}^nW_i^s \otimes(W_i^dW_i^u),
\end{equation}
where $n$ is a user-defined hyper-parameter and denotes the number of adapters or experts, $W_i^d \in \mathbb{R}^{\frac{d}{n}\times r}$ and $W_i^u \in \mathbb{R}^{r\times \frac{d}{n}}$ represent the down-projection layer and up-projection layer respectively in the $i$-th expert, $\otimes$ denotes the Kronecker product. Formally, the key difference from the previous approaches~\cite{adaptformer,adapter,lora}, which involve a single decomposition in the original feature space, is that we directly reuse this hypercomplex feature space to perform multiple low rank decompositions. This enables us to exponentially increase the number of branches. Thanks to the artful design, we only introduce extra $n^3$ ($n$ is very small) parameters compared with the original adapter-based methods but multiple low rank adapters with a multi-branch structure. Detailed parameter size analyses and comparisons are presented in appendix~\ref{app:param}. The forward process of our ALoRE can be defined as:
\vspace{-5pt}
\begin{equation}
\small
    \begin{split}
        &{\rm ALoRE}(X) = X\sum_{i=1}^nW_i^s \otimes(W_i^dW_i^u) + X, \\
        &{X_l^\prime} = {\rm MHSA}({\rm ALoRE}_{\rm MHSA}({\rm LN}(X_{l-1}))) + {X_{l-1}}, \\
        &{X_l} = {\rm FFN}({\rm ALoRE}_{\rm FFN}({\rm LN}(X_l^\prime))) + {X_l^\prime},
    \end{split}
\end{equation}
where ${\rm ALoRE}_{\rm MHSA}$ and ${\rm ALoRE}_{\rm FFN}$ are two independent modules placed before MHSA and FFN.

\hspace{-0.4cm}\textbf{Re-parameterization.} Thanks to the purely linear transformation design of our ALoRE module, we can effortlessly merge it into the nearest pre-trained linear layer via re-parameterization during inference. Considering the sequential re-parameterization, we can formulate as follows:
\begin{equation}
\small
    \begin{split}
        &\hspace{0.4cm}{\rm ALoRE}(X)W_0+b_0 \\
        &= X\big(\sum_{i=1}^nW_i^s \otimes(W_i^dW_i^u) + I\big)W_0 + b_0, \\
        &= XW_{\rm ALoRE}W_0+b_0, \\
        &= X\widetilde{W} + \widetilde{b},
    \end{split}
\end{equation}
where $W_0$ and $b_0$ are the nearest linear projection weight matrices in the pre-trained backbone, $I$ denotes an identity matrix, $W_{\rm ALoRE}$ can be constructed by $\sum_{i=1}^nW_i^s \otimes(W_i^dW_i^u) + I$, and $\widetilde{W}$ and $\widetilde{b}$ represent the final weight matrices for inference, posing no shape modification.

\hspace{-0.4cm}\textbf{Discussion.} Typical methods (adapter-based~\cite{adaptformer,adapter,restuning,rep} and LoRA-like~\cite{glora,lora,adalora}) primarily leverage the intrinsic low rank characteristic~\cite{aghajanyan2020intrinsic,li2018measuring} and directly perform decomposition on the original dense feature space. Despite significantly reducing the parameter count and achieving acceptable performance, the direct parameter updates in a single-branch learning way tend to limit the decoupling ability of the model for learned feature patterns. For instance, in the case of image understanding tasks, a visual entity consists of diverse components, including foreground and background, edge and subject. A robust visual modeling process should operate at multiple semantic levels, whereas assigning complex tasks to a compressed feature space within a single-branch network can lead to confounding learned representation patterns, thereby constraining the expressive capabilities of the model. Hence, we leverage the Kronecker product to construct a hypercomplex parameterized space to substitute the original feature space. We then reuse the new parameter space to perform low rank decomposition, ultimately obtaining multiple experts that facilitate the decoupling of learned representation patterns and minimizing extra parameter costs. We will conduct a visual analysis in Section~\ref{subsec:vis} to provide further insights.
\section{Experiments} 
\label{sec:experiments}

\subsection{Experimental setups}
\label{sec:setups}

\textbf{Datasets.} To evaluate the \textit{effectiveness}, \textit{efficiency}, and \textit{scalability} of our proposed ALoRE, we conduct experiments on two visual adaptation benchmarks consisting of 24 datasets. The details of the datasets are provided below:

\begin{table*}[t]
  \centering
  \setlength{\tabcolsep}{3pt}
  \resizebox{0.95\linewidth}{!}{
    \begin{tabular}{c|c|c|c|c|c|cc}
    \toprule[2pt]
    \diagbox{\textbf{Method}}{\textbf{Dataset}}  & \textbf{CUB-200-2011} & \textbf{NABirds} & \textbf{Oxford Flowers} & \textbf{Stanford Dogs} & \textbf{Stanford Cars} & \textbf{Mean}  & \textbf{Params.(M)} \\
    \midrule[1pt]
    \multicolumn{8}{c}{\textit{Traditional methods}} \\
    Full fine-tuning~\cite{vpt} & 87.3  & 82.7  & 98.8  & 89.4  & 84.5  & 88.54 & 85.98 \\
    Linear probing~\cite{vpt} & 85.3  & 75.9  & 97.9  & 86.2  & 51.3  & 79.32 & 0.18 \\
    \midrule[1pt]
    \multicolumn{8}{c}{\textit{Parameter-efficient transfer learning methods}} \\
    Adapter~\cite{adapter} & 87.1  & 84.3  & 98.5  & 89.8  & 68.6  & 85.66 & 0.41 \\
    Bitfit~\cite{bitfit} & 88.4  & 84.2  & 98.8  & 91.2  & 79.4  & 88.40  & 0.28 \\
    VPT-Shallow~\cite{vpt} & 86.7  & 78.8  & 98.4  & 90.7  & 68.7  & 84.66  & 0.25 \\
    VPT-Deep~\cite{vpt} & 88.5 & 84.2 & 99.0 & 90.2 & 83.6  & 89.10 & 0.85 \\
    \midrule[1pt]
    \multicolumn{8}{c}{\textit{Structural re-parameterization methods}} \\
    LoRA~\cite{lora} & 88.3 & 85.6  & 99.2   & 91.0  & 83.2  & 89.46 & 0.44 \\
    SSF~\cite{ssf} & 89.5 & 85.7 & \textbf{99.6} & 89.6  & 89.2  & 90.72 & 0.39 \\
    ARC~\cite{arc} & 88.5 & 85.3  & 99.3  & \textbf{91.9} & 85.7 & 90.14 & 0.25 \\
    RepAdapter~\cite{rep} & 89.4 & 85.9  & \uline{99.5}  & 90.6 & 85.9 & 90.30 & 0.34 \\
    \rowcolor[rgb]{ .906,  .902,  .902} ALoRE\({\rm{}_{attn}}\) & \uline{89.6}  & \uline{86.5} & \uline{99.5} & \uline{91.7}  & \textbf{90.0}  & \uline{91.46} & 0.22 \\
    \rowcolor[rgb]{ .906,  .902,  .902} ALoRE  & \textbf{89.9}  & \textbf{87.1}  & \textbf{99.6}  & \textbf{91.9}  & \uline{89.5}  & \textbf{91.60} & 0.27 \\
    \bottomrule[2pt]
    \end{tabular}
    }
    \caption{Performance and parameter efficiency comparisons on five FGVC datasets with ViT-B/16 pre-trained on ImageNet-21K. We strive to utilize the result reported in the previous paper to the best ability and reproduce RepAdapter~\cite{rep} based on the officially released code, as it has not been evaluated on this benchmark. The \textbf{bold} font shows the best Top-1 accuracy and the \uline{underline} font shows the second-best Top-1 accuracy in the structural re-parameterization methods.}
    \label{tab:fgvc}
    \vspace{-15pt}
\end{table*}

\textit{FGVC.} Following the default setups in VPT~\cite{vpt}, we conduct experiments on five Fine-Grained Visual Classification (FGVC) datasets, which comprises CUB-200-2011~\cite{cub}, NABirds~\cite{birds}, Oxford Flowers~\cite{flowers}, Stanford Dogs~\cite{dogs}, and Stanford Cars~\cite{cars}.

\textit{VTAB-1k.} We also evaluate our ALoRE approach on VTAB-1k benchmark~\cite{vtab}, which contains 19 visual classification tasks from diverse image domains. These tasks are categorized into three groups: \textit{Natural} group includes images captured by standard cameras; \textit{Specialized} group contains images captured by advanced equipment, \textit{e.g.}, remote sensing and medical cameras; \textit{Structured} group consists of synthesized images from simulated environments, \textit{e.g.}, 3D depth estimation and object counting. This benchmark comprises downstream tasks that are distinct from typical image domains, and each task is limited to only 1000 training images, rendering the benchmark highly challenging.

\hspace{-0.4cm}\textbf{Models.} In order to minimize the interference of experimental factors and ensure a fair comparison with previous work, we follow VPT~\cite{vpt} and mainly choose ViT-B/16~\cite{vit} model pre-trained on ImageNet-21K~\cite{imagenet21k} as the initialization for fine-tuning. We also conduct experiments on different backbone variants with varying model sizes: ViT-Base/Large/Huge. In addition, we extend our method to different model architectures: Swin Transformer (Swin-B)~\cite{swin}, ConvNeXt-B~\cite{convnet} and AS-MLP-B~\cite{asmlp,mlp-mixer}, the former is hierarchical transformer-based, and the latter two are CNN-based and MLP-based respectively. Specific pre-trained backbones will be presented in the appendix~\ref{app:weights}.

\hspace{-0.4cm}\textbf{Baselines.} Apart from the traditional fine-tuning methods like full fine-tuning and linear probing, we divide existing tuning methods into two categories based on whether they introduce additional latency during inference. (i) Classic PETL methods: Adapter~\cite{adapter}, AdaptFormer~\cite{adaptformer}, VPT~\cite{vpt}, Convpass~\cite{convpass}, Compacter~\cite{compacter}, NOAH~\cite{noah} and Res-Tuning~\cite{restuning}; (ii) Structural re-parameterization methods:  LoRA~\cite{lora}, SSF~\cite{ssf}, ARC~\cite{arc} and RepAdapter~\cite{rep}.

\hspace{-0.4cm}\textbf{Implementation Details.} We follow the default setups of VPT~\cite{vpt} to ensure a fair evaluation of the effectiveness of our method. For the FGVC datasets, we preprocess the image using random resize crop to $224 \times 224$ and random horizontal flip. For the VTAB-1k benchmark~\cite{vtab}, we only resize the image to $224 \times 224$. We employ the AdamW~\cite{adamw} optimizer to fine-tune models for 100 epochs and the cosine decay strategy to schedule the learning rate with a linear warm-up for 10 epochs. Further implementation details will be presented in the appendix~\ref{app:impl}.

\subsection{Experimental Comparisons}
\label{sec:exp_comparions}

\begin{table*}[t]
  \centering
  \setlength{\tabcolsep}{3pt}
  \resizebox{0.98\linewidth}{!}{
    \begin{tabular}{c|ccccccc|cccc|cccccccc|ccc}
    \toprule[2pt]
    & \multicolumn{7}{c|}{\textbf{Natural}}      
    & \multicolumn{4}{c|}{\textbf{Specialized}}      
    & \multicolumn{8}{c|}{\textbf{Structured}}         &  & & \\
    \midrule[1pt]
    \diagbox{\textbf{Method}}{\textbf{Dataset}} & \begin{sideways}\textbf{CIFAR-100}\end{sideways} & \begin{sideways}\textbf{Caltech101}\end{sideways} & \begin{sideways}\textbf{DTD}\end{sideways} & \begin{sideways}\textbf{Flowers102}\end{sideways} & \begin{sideways}\textbf{Pets}\end{sideways} & \begin{sideways}\textbf{SVNH}\end{sideways} & \begin{sideways}\textbf{Sun397}\end{sideways} & \begin{sideways}\textbf{Camelyon}\end{sideways} & \begin{sideways}\textbf{EuroSAT}\end{sideways} & \begin{sideways}\textbf{Resisc45}\end{sideways} & \begin{sideways}\textbf{Retinopathy}\end{sideways} &  \begin{sideways}\textbf{Clevr-Count}\end{sideways} & \begin{sideways}\textbf{Clevr-Dist}\end{sideways} & \begin{sideways}\textbf{DMLab}\end{sideways} & \begin{sideways}\textbf{KITTI-Dist}\end{sideways} & \begin{sideways}\textbf{dSpr-Loc}\end{sideways} & \begin{sideways}\textbf{dSpr-Ori}\end{sideways} & \begin{sideways}\textbf{sNORB-Azim}\end{sideways} & \begin{sideways}\textbf{sNORB-Ele}\end{sideways} &  \begin{sideways}\textbf{All Mean}\end{sideways} &  \begin{sideways}\textbf{Group Mean}\end{sideways} & \begin{sideways}\textbf{Params(M)}\end{sideways} \\
    \midrule[1pt]
    \multicolumn{23}{c}{\textit{Traditional methods}} \\
    Full fine-tuning~\cite{vpt} & 68.9  & 87.7  & 64.3  & 97.2  & 86.9  & 87.4  & 38.8  & 79.7  & 95.7  & 84.2  & 73.9  & 56.3  & 58.6  & 41.7  & 65.5  & 57.5  & 46.7  & 25.7  & 29.1  & 65.57 & 68.97 & 85.84 \\
    Linear probing~\cite{vpt} & 63.4  & 85.0    & 63.2  & 97.0    & 86.3  & 36.6  & 51.0    & 78.5  & 87.5  & 68.6  & 74.0    & 34.3  & 30.6  & 33.2  & 55.4  & 12.5  & 20.0    & 9.6   & 19.2  & 52.94  & 57.64 & 0.04 \\
    \midrule[1pt]
    \multicolumn{23}{c}{\textit{Parameter-efficient transfer learning methods}} \\
    Adapter~\cite{adapter} & 69.2  & 90.1  & 68.0  & 98.8  & 89.9    & 82.8  & 54.3  & 84.0  & 94.0    & 81.9  & 75.5  & 80.9  & 65.3  & 48.6  & 78.3  & 74.8  & 48.5  & 29.9  & 41.6  & 71.44  & 73.86 & 0.16 \\
    AdapterFormer~\cite{adaptformer} & 70.8  & 91.2  & 70.5  & 99.1  & 90.9    & 86.6  & 54.8  & 83.0  & 95.8    & 84.4  & 76.3  & 81.9  & 64.3  & 49.3  & 80.3  & 76.3  & 45.7  & 31.7  & 41.1  & 72.32  & 74.75 & 0.16 \\
    VPT-Shallow~\cite{vpt} & 77.7 & 86.9 & 62.6 & 97.5 & 87.3 & 74.5 & 51.2 & 78.2 & 92.0 & 75.6 & 72.9 & 50.5 & 58.6 & 40.5 & 67.1 & 68.7 & 36.1 & 20.2 & 34.1 & 64.85 & 67.82 & 0.11 \\
    VPT-Deep~\cite{vpt} & 78.8 & 90.8 & 65.8 & 98.0 & 88.3 & 78.1 & 49.6 & 81.8 & 96.1 & 83.4 & 68.4 & 68.5 & 60.0 & 46.5 & 72.8 & 73.6 & 47.9 & 32.9 & 37.8 & 69.43 & 71.97 & 0.60 \\
    Compacter~\cite{compacter} & 71.9  & 89.0  & 69.7  & 99.1  & 90.7    & 82.7  & 56.1  & 86.0  & 93.5    & 82.4  & 75.3  & 80.2  & 63.4  & 47.4  & 77.2  & 78.1  & 53.5  & 27.3  & 39.8  & 71.75  & 74.18 & 0.62 \\
    Convpass~\cite{convpass} & 72.3 & 91.2 & 72.2 & 99.2 & 90.9 & 91.3 & 54.9 & 84.2 & 96.1 & 85.3 & 75.6 & 82.3 & 67.9 & 51.3 & 80.0 & 85.9 & 53.1 & 36.4 & 44.4 & 74.45 & 76.56 & 0.33 \\
    NOAH~\cite{noah} & 69.6 & 92.7 & 70.2 & 99.1 & 90.4 & 86.1 & 53.7 & 84.4 & 95.4 & 83.9 & 75.8 & 82.8 & 68.9 & 49.9 & 81.7 & 81.8 & 48.3 & 32.8 & 44.2 & 73.25 & 75.48 & 0.42 \\
    Res-Tuning~\cite{restuning} & 75.2 & 92.7 & 71.9 & 99.3 & 91.9 & 86.7 & 58.5 & 86.7 & 95.6 & 85.0 & 74.6 & 80.2 & 63.6 & 50.6 & 80.2 & 85.4 & 55.7 & 31.9 & 42.0  & 74.10 & 76.32 & 0.55 \\
    \midrule[1pt]
    \multicolumn{23}{c}{\textit{Structural re-parameterization methods}} \\
    BitFit~\cite{bitfit} & 72.8 & 87.0 & 59.2 & 97.5 & 85.3 & 59.9 & 51.4 & 78.7 & 91.6 & 72.9 & 69.8 & 61.5 & 55.6 & 32.4 & 55.9 & 66.6 & 40.0 & 15.7 & 25.1 & 62.05 & 65.22 & 0.10 \\
    LoRA~\cite{lora} & 65.3 & 87.9 & 69.4 & 98.7 & 90.7 & 82.4 & 53.4 & 82.8 & 94.8 & 82.5 & 75.0 & 77.6 & 64.7 & 45.8 & 79.0 & 73.3 & 44.7 & 26.3 & 38.2 & 70.10 & 72.74 & 0.29 \\
    SSF~\cite{ssf} & 69.0 & \uline{92.6} & \textbf{75.1} & \uline{99.4} & 91.8 & 90.2 & 52.9 & \uline{87.4} & \uline{95.9} & \uline{87.4} & 75.5 & 75.9 & 62.3 & \textbf{53.3} & \uline{80.6} & 77.3 & \uline{54.9} & 29.5 & 37.9 & 73.10 & 75.69 & 0.24 \\
    ARC~\cite{arc} & 72.2  & 90.1  & 72.7 & 99.0 & 91.0 & \textbf{91.9} & 54.4 & 84.9  & 95.7  & 86.7 & 75.8 & 80.7  & \uline{67.1}  & 48.7  & \textbf{81.6} & 79.2 & 51.0 & 31.4  & 39.9 & 73.40 & 75.78 & 0.13\\
    RepAdapter~\cite{rep} & 72.4  & 91.6  & 71.0 & 99.2 & 91.4 & 90.7 & 55.1 & 85.3  & \uline{95.9}  & 84.6 & 75.9 & \uline{82.3}  & \textbf{68.0}  & 50.4  & 79.9 & 80.4 & 49.2 & \textbf{38.6}  & 41.0 & 73.84 & 76.10 & 0.22 \\
    \rowcolor[rgb]{ .906,  .902,  .902} ALoRE\({\rm{}_{attn}}\) & \uline{75.2}  & \uline{92.6}  & 73.6 & \uline{99.4} & \textbf{92.3} & 90.3 & \uline{57.5} & 87.0  & 95.5  & 87.0 & \textbf{76.4} & 81.6  & 65.9  & \uline{52.6}  & 79.5 & \uline{84.3} & \textbf{55.1} & 31.9 & \textbf{45.7} & \uline{74.92} & \uline{77.18} & 0.07 \\
    \rowcolor[rgb]{ .906,  .902,  .902} ALoRE & \textbf{76.4}  & \textbf{93.2}  & \uline{74.1} & \textbf{99.5} & \uline{92.2} & \uline{91.2} & \textbf{57.9} & \textbf{88.2}  & \textbf{96.0}  & \textbf{87.8} & \uline{76.1} & \textbf{82.5}  & 66.3  & \textbf{53.3}  & \textbf{81.6} & \textbf{86.5} & \uline{54.9} & \uline{32.0}  & \uline{45.4} & \textbf{75.54} & \textbf{77.78} & 0.15 \\
    \bottomrule[2pt]
    \end{tabular}
    }
    \caption{Performance and parameter efficiency comparisons on the VTAB-1k benchmark with ViT-B/16 pre-trained on ImageNet-21K. We strive to utilize the results reported in the previous paper to the best of our ability. ``Group Mean'' denotes the average Top-1 accuracy of the three subgroups. ``All Mean'' denotes the average Top-1 accuracy of 19 downstream tasks.}
    \label{tab:vtab1k}
    \vspace{-10pt}
\end{table*}
\begin{table*}
\centering
\begin{subtable}[t]{0.495\linewidth}
    \centering
    \caption{ViT-Large}
    \label{tab:vit-large}
    \resizebox{1\columnwidth}{!}{
    \begin{tabular}{c|c|c|c|ccc}
    \toprule[2pt]
    \diagbox{\textbf{Method}}{\textbf{Dataset}} & \textbf{Natural} & \textbf{Specialized} & \textbf{Structured} & \textbf{All Mean} & \textbf{Group Mean} & \textbf{Params(M)} \\
    \midrule[1pt]
    Full fine-tuning~\cite{arc} & 74.7  & 83.8  & 48.1  & 65.43 & 68.87  & 303.40 \\
    Linear probing~\cite{arc} & 70.9  & 69.1  & 25.8  & 51.49 & 55.23 & 0.05 \\
    \midrule[1pt]
    Adapter~\cite{adapter} & 68.6  & 73.5  & 29.0  & 52.92  & 56.99 & 2.38 \\
    VPT-Deep~\cite{vpt} & 82.5 & 83.9  & 54.1 & 70.85 & 73.51 & 0.49 \\
    \midrule[1pt]
    LoRA~\cite{lora} & 81.4 & 85.0  & 57.3 & 72.00 & 74.55 & 0.74 \\
    ARC~\cite{arc} & 82.3 & 85.6 & 57.3 & 72.46 & 75.06 & 0.18 \\
    RepAdapter~\cite{adapter} & \uline{84.0} & \uline{86.3} & \uline{60.1} & \uline{74.43} & \uline{76.81} & 0.79 \\
    \rowcolor[rgb]{ .906,  .902,  .902} ALoRE & \textbf{85.0} & \textbf{87.5} & \textbf{61.3} & \textbf{75.55} & \textbf{77.93} & 0.39 \\
    \bottomrule[2pt]
    \end{tabular}
    }
\end{subtable}
\begin{subtable}[t]{0.495\linewidth}
    \centering
    \caption{ViT-Huge}
    \label{tab:vit-huge}
    \resizebox{1\columnwidth}{!}{
    \begin{tabular}{c|c|c|c|ccc}
    \toprule[2pt]
    \diagbox{\textbf{Method}}{\textbf{Dataset}} & \textbf{Natural} & \textbf{Specialized} & \textbf{Structured} & \textbf{All Mean} & \textbf{Group Mean} & \textbf{Params(M)} \\
    \midrule[1pt]
    Full fine-tuning~\cite{arc} & 70.9  & 83.6  & 46.0  & 63.07 & 66.81  & 630.90 \\
    Linear probing~\cite{arc} & 67.9  & 79.0  & 26.1  & 52.66 & 57.68 & 0.06 \\
    \midrule[1pt]
    Adapter~\cite{adapter} & 68.1  & 76.4  & 24.5    & 51.52  & 56.36 & 5.78 \\
    VPT-Deep~\cite{vpt} & 77.9 & 83.3  & 52.2 & 68.23 & 71.14 & 0.96 \\
    \midrule[1pt]
    LoRA~\cite{lora} & 77.1 & 83.5  & 55.4 & 69.28 & 71.96 & 1.21 \\
    ARC~\cite{arc} & \uline{79.1} & \textbf{84.8} & 53.7 & 69.62 & 72.54 & 0.22 \\
    RepAdapter~\cite{rep} & 77.9 & 83.2 & \uline{57.6} & \uline{70.48} & \uline{72.91} & 1.33 \\
    \rowcolor[rgb]{ .906,  .902,  .902} ALoRE & \textbf{80.0} & \uline{84.6} & \textbf{59.0} & \textbf{72.11} & \textbf{74.51} & 0.66 \\
    \bottomrule[2pt]
    \end{tabular}
    }
\end{subtable}
\caption{Performance and parameter efficiency comparisons on the VTAB-1k using ViT-L/16 and ViT-H/14 pre-trained on ImageNet-21K.}
\label{tab:scale}
\vspace{-12pt}
\end{table*}

\textbf{Performance and parameter efficiency trade-off.} As shown in Table~\ref{tab:fgvc} and Table~\ref{tab:vtab1k}, we compare the performance of our ALoRE and other baselines in terms of Top-1 accuracy (\%) on FGVC datasets and VTAB-1k benchmark. Our ALoRE method demonstrates state-of-the-art performance and high parameter efficiency. In order to ensure fairness in comparison, that is, to keep the parameter size usage of each method as consistent as possible, we also develop a variant of ALoRE called ALoRE\({\rm{}_{attn}}\), which incorporates the module we have designed solely before the multi-head self-attention module, thus further reducing the parameter size by half. In comparison to the PETL methods mentioned in the table, full fine-tuning attains inferior performance across various downstream tasks, despite updating a considerable amount of parameters. This observation highlights the significance of fine-tuning strategies to avoid overfitting in environments where training data is scarce. 

\hspace{-0.4cm}\textbf{Scalability.} As presented in Table~\ref{tab:scale} and Table~\ref{tab:backbone}, We have extended the design of ALoRE to larger-scale backbones of different underlying architectures and various pre-training strategies. Experimental results demonstrate our method consistently outperforms classic PETL methods and other state-of-the-art re-parameterization methods, highlighting the scalability and generality of our approach.

\subsection{Ablation Studies}
With the purpose of gaining valuable insights into our ALoRE method, in Table~\ref{tab:structure} and ~\ref{tab:ablation}, we conduct systematic ablation studies and analyses to identify its crucial properties. All experimental results are obtained by evaluating the pre-trained ViT-B/16 model on the VTAB-1k benchmark.

\vspace{-5pt}
\begin{table}[h]
    \centering
    \resizebox{1\columnwidth}{!}{
    \begin{tabular}{c|c|c|c|ccc}
        \toprule[2pt]
        \diagbox{\textbf{Method}}{\textbf{Dataset}} & \textbf{Natural} & \textbf{Specialized} & \textbf{Structured} & \textbf{All Mean} & \textbf{Group Mean} & \textbf{Params(M)} \\
        \midrule[1pt]
        stacked non-linear & 80.73 & 85.93 & 60.38 & 73.26 & 75.68 & 0.66 \\
        stacked linear & 81.37 & 86.25 & 60.91 & 73.79 & 76.18 & 0.66 \\
        \rowcolor[rgb]{ .906,  .902,  .902} ALoRE & \textbf{83.51} & \textbf{87.04} & \textbf{62.81} & \textbf{75.54} & \textbf{77.78} & 0.15 \\
        \bottomrule[2pt]
    \end{tabular}
    }
    \caption{Ablations of structure design in ALoRE, where "stacked non-linear" denotes directly replicating the original adapter multiple times in parallel within the model to aggregate low rank experts, which introduces parameters that grow linearly with the number of branches. "stacked linear" means dismissing the non-linear component in original adapter as a purely linear module.}
    \label{tab:structure}
    \vspace{-10pt}
\end{table}
\begin{table*}
\centering
    \begin{subtable}[t]{0.495\linewidth}
        \centering
        \caption{Swin-Base}
        \label{tab:swin-base}
        \resizebox{1.0\columnwidth}{!}{
        \begin{tabular}{c|c|c|c|ccc}
        \toprule[2pt]
        \diagbox{\textbf{Method}}{\textbf{Dataset}} & \textbf{Natural} & \textbf{Specialized} & \textbf{Structured} & \textbf{All Mean} & \textbf{Group Mean} & \textbf{Params(M)} \\
        \midrule[1pt]
        Full fine-tuning~\cite{arc} & 79.1  & 86.2  & 59.7  & 72.42 & 74.99  & 86.90 \\
        Linear probing~\cite{arc} & 73.5  & 80.8  & 33.5  & 58.22 & 62.62 & 0.05 \\
        \midrule[1pt]
        VPT-Shallow~\cite{vpt} & 79.9 & 82.4 & 37.8 & 62.71 & 66.71 & 0.05 \\
        VPT-Deep~\cite{vpt} & 76.8 & 84.5  & 53.4 & 68.59 & 71.59 & 0.22 \\
        \midrule[1pt]
        ARC~\cite{arc} & 79.0 & 86.6 & 59.9 & 73.01 & 75.60 & 0.16 \\
        RepAdapter~\cite{rep} & \uline{82.8} & \uline{87.2} & \uline{61.2} & \uline{74.61} & \uline{77.04} & 0.39 \\
        \rowcolor[rgb]{ .906,  .902,  .902} ALoRE & \textbf{83.6} & \textbf{89.2} & \textbf{61.5} & \textbf{75.47} & \textbf{78.09} & 0.19 \\
        \bottomrule[2pt]
        \end{tabular}
        }
    \end{subtable}
    \begin{subtable}[t]{0.495\linewidth}
        \centering
        \caption{ConvNeXt-Base}
        \label{tab:conv-base}
        \resizebox{1.0\columnwidth}{!}{
        \begin{tabular}{c|c|c|c|ccc}
        \toprule[2pt]
        \diagbox{\textbf{Method}}{\textbf{Dataset}} & \textbf{Natural} & \textbf{Specialized} & \textbf{Structured} & \textbf{All Mean} & \textbf{Group Mean} & \textbf{Params(M)} \\
        \midrule[1pt]
        Full fine-tuning & 77.5  & 84.2  & 59.1  & 71.16 & 73.59  & 87.56 \\
        Linear probing & 76.3  & 82.8  & 35.8  & 60.61 & 64.95 & 0.04 \\
        \midrule[1pt]
        Adapter~\cite{adapter} & 78.8 & 79.2 & 57.3 & 69.84 & 71.77 & 0.67 \\
        Adaptformer~\cite{adaptformer} & 83.1 & 85.5  & 59.7 & 73.74 & 76.09 & 0.34 \\
        \midrule[1pt]
        LoRA~\cite{lora} & 79.2 & 83.4  & 53.8 & - & 72.13 & 0.40 \\
        SSF~\cite{ssf} & \uline{83.8} & \uline{87.0} & \uline{64.6} & \uline{76.38} & \uline{78.46} & 0.27 \\
        \rowcolor[rgb]{ .906,  .902,  .902} ALoRE & \textbf{85.1} & \textbf{88.3} & \textbf{66.1} & \textbf{77.75} & \textbf{79.81} & 0.29 \\
        \bottomrule[2pt]
        \end{tabular}
        }
    \end{subtable}
    \caption{Performance and parameter efficiency comparisons on the VTAB-1k benchmark using Swin-Base and ConvNeXt-Base pre-trained on ImageNet-21K. ``-`` denotes the missing value in the previous paper. Due to the lack of comparisons on CNN-like backbones, We implement the fine-tuning methods with ConvNeXt-Base from scratch, except LoRA, which is taken from~\cite{rep}.}
    \label{tab:backbone}
    \vspace{-5pt}
\end{table*}
\begin{table*}
    \centering
    \vspace{-5pt}
    \begin{subtable}[t]{0.22\linewidth}
        \centering
        \caption{Rank dimensions.}
        \label{tab:abl_rank}
    \scalebox{0.75}{
        \begin{tabular}{ccc}
            \toprule
            \textbf{Rank} & \textbf{Acc.} & \textbf{Params.} \\         \midrule
            1 &  73.90 & 0.04 \\
            2  & 74.58 & 0.07 \\
            \rowcolor[rgb]{ .906,  .902,  .902}4 & \textbf{75.54} & 0.15 \\
            8 & 75.39 & 0.29 \\
            16 & 74.86 & 0.59 \\
                32 & 74.35 & 1.18 \\
            \bottomrule
    \end{tabular}}
    \end{subtable}
    \hfill
    \begin{subtable}[t]{0.22\linewidth}
            \centering
            \caption{Nums of experts.}
            \label{tab:abl_nums}
        \scalebox{0.75}{
            \begin{tabular}{ccc}
                \toprule
                \textbf{Nums} & \textbf{Acc.} & \textbf{Params.} \\ \midrule
                1  & 73.81 & 0.15 \\
                2  &  74.58 & 0.15 \\
                \rowcolor[rgb]{ .906,  .902,  .902}4 & \textbf{75.54} & 0.15 \\
                8 & 75.22 & 0.15 \\
                16 & 74.85 & 0.15 \\
                    32 & 74.89 & 0.18 \\
                \bottomrule
        \end{tabular}}
    \end{subtable}
    \hfill
    \begin{subtable}[t]{0.26\linewidth}
            \centering
            \caption{ALoRE location.}
            \label{tab:abl_loc}
        \scalebox{0.75}{
            \begin{tabular}{ccc}
                \toprule
                \textbf{Location} & \textbf{Acc.} & \textbf{Params.} \\ \midrule
                Before MHA  & 74.92 & 0.07 \\
                After MHA  &  74.29 & 0.07 \\
                    Before FFN & 73.89 & 0.07  \\
                    After FFN & 74.77 & 0.07 \\
                \rowcolor[rgb]{ .906,  .902,  .902}Before MHA \& FFN & \textbf{75.54} & 0.15 \\
                    After MHA \& FFN & 75.25 & 0.15 \\
                \bottomrule
        \end{tabular}}
    \end{subtable}
    \hfill
    \hspace{0.2cm}
    \begin{subtable}[t]{0.24\linewidth}
            \centering
            \caption{Insert selection.}
            \label{tab:abl_sel}
        \scalebox{0.75}{
            \begin{tabular}{ccc}
                \toprule
                \textbf{\#Layers} & \textbf{Acc.} & \textbf{Params.} \\ \midrule
                2 & 68.76 & 0.02 \\
                4 & 71.97 & 0.05 \\
                  6 & 73.69 & 0.07 \\
                  8 & 74.39 & 0.10 \\
                  10 & 74.94 & 0.12 \\
                    \rowcolor[rgb]{ .906,  .902,  .902}12 & \textbf{75.54} & 0.15 \\
                \bottomrule
        \end{tabular}}
    \end{subtable}
    \caption{Ablations of ALoRE variants on VTAB-1k benchmark with ViT-B/16 backbone. The table showcases average accuracy across 19 downstream datasets and parameter count of different variants. The rows in \colorbox{mygray}{\textbf{gray}} background are the default settings we adopt.}
    \label{tab:ablation}
    \vspace{-10pt}
\end{table*}

\hspace{-0.4cm}\textbf{Structure design.} Formally, the structure of our proposed ALoRE is predicated upon two innovations: (1) the utilization of the Kronecker product to construct a hypercomplex parameterized space that supplants the original dense feature space within adapter; (2) the elimination of non-linear structure in the adapter to enhance inference efficiency. We present ablations on these two aspects in Table~\ref{tab:structure}. A comparison between the first and second rows reveals that the employment of a purely linear structure yields a slight improvement over the non-linear configuration while concurrently reducing inference overhead. Furthermore, an examination of the second and third rows demonstrates that performing low rank decomposition within the hypercomplex parameterized space results in a significant boost relative to the original dense feature space, with the extra benefit of achieving an exponential reduction in parameter size.

\hspace{-0.4cm}\textbf{Low rank dimensions.} We investigate the impact of the bottleneck dimension of each low rank expert in ALoRE on the performance of visual adaptation. We find that increasing the bottleneck dimension leads to a linear relationship with the parameter complexity, resulting in a significant increment in parameter redundancy and overhead. By comparing the third and sixth rows in Table~\ref{tab:abl_rank}, it can be inferred that increasing the bottleneck dimension does not always lead to improved performance, and sometimes it even results in decreased performance (75.54\% \textit{vs.} 74.35\%, 0.15M \textit{vs.} 1.18M). It is mainly due to the overfitting problem when adapting to small-scale downstream datasets. The experimental results demonstrate that a bottleneck dimension of 4 can achieve a trade-off between adaptation performance and parameter efficiency, which also confirms the effectiveness and rationality of the low rank design~\cite{aghajanyan2020intrinsic,li2018measuring} of experts.

\hspace{-0.4cm}\textbf{Number of experts.} Our ALoRE leverages the aggregation of low rank experts as its core mechanism. We investigate the impact of varying quantities of low rank experts in Table~\ref{tab:abl_nums}. When the number of low rank experts is set to 1, our method reduces to the conventional RepAdapter~\cite{rep} form, except each layer has an additional learnable scaling factor. From the first three rows of Table~\ref{tab:abl_nums}, we verify the effectiveness of our method, as increasing the number of low rank experts while maintaining the bottleneck dimension not only introduces a negligible number of parameters but also yields a 1.73\% average accuracy improvement (75.54\% \textit{vs.} 73.81\%) across 19 downstream datasets. We explain this phenomenon by positing that the fusion of multiple low rank experts in ALoRE encourages each expert to focus on different regions, corners and patterns of visual entities during network training. Similarly to increasing the bottleneck dimension, we infer from the last three rows that the optimal number of low rank experts is task-dependent, with more experts possibly leading to redundancy and confusion, resulting in saturation and performance degradation.

\hspace{-0.4cm}\textbf{ALoRE location.} In Table~\ref{tab:abl_loc}, we investigate the impact of the placement of ALoRE in the backbone on performance. The experimental results in the first two rows show that placing ALoRE module before MHA leads to improved performance compared to placing it after, whereas the opposite is true for FFN as shown in the third and fourth rows, which results in decreased performance. Moreover, using ALoRE on only one of MHA and FFN falls short of using it on both.

\begin{figure*}[!t]
    \centering
    \includegraphics[width=0.9\textwidth]{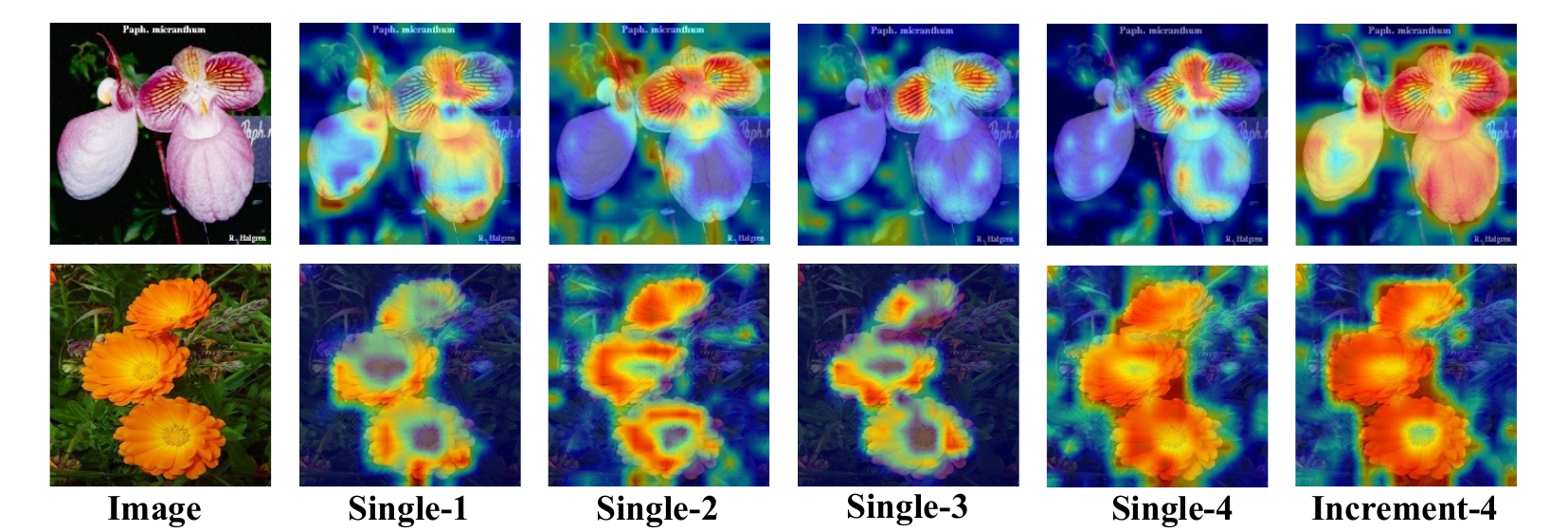}
    \vspace{-5pt}
    \caption{Visualization of attention maps with respect to different experts. ``Single-$i$`` denotes the exclusive preservation of the $i$-th expert only. ``Increment-4`` represents the aggregation of all experts.}
    \label{fig:gradcam}
    \vspace{-15pt}
\end{figure*}

\hspace{-0.4cm}\textbf{Insert selection.} We also study the effect of layers we choose to insert ALoRE module in Table~\ref{tab:abl_sel}. The numerical values presented under the column labeled "\#Layers" indicate the extent to which the ALoRE is employed across the preceding layers of the model. We find that adding the ALoRE module consistently across all subsequent layers leads to improvement in performance. Therefore, we propose that our method defaults to the inclusion of the ALoRE module in each layer.

\subsection{Visualization and Analysis}
\label{subsec:vis}
\begin{figure}[t]
    \centering
    \includegraphics[width=1.0\columnwidth]{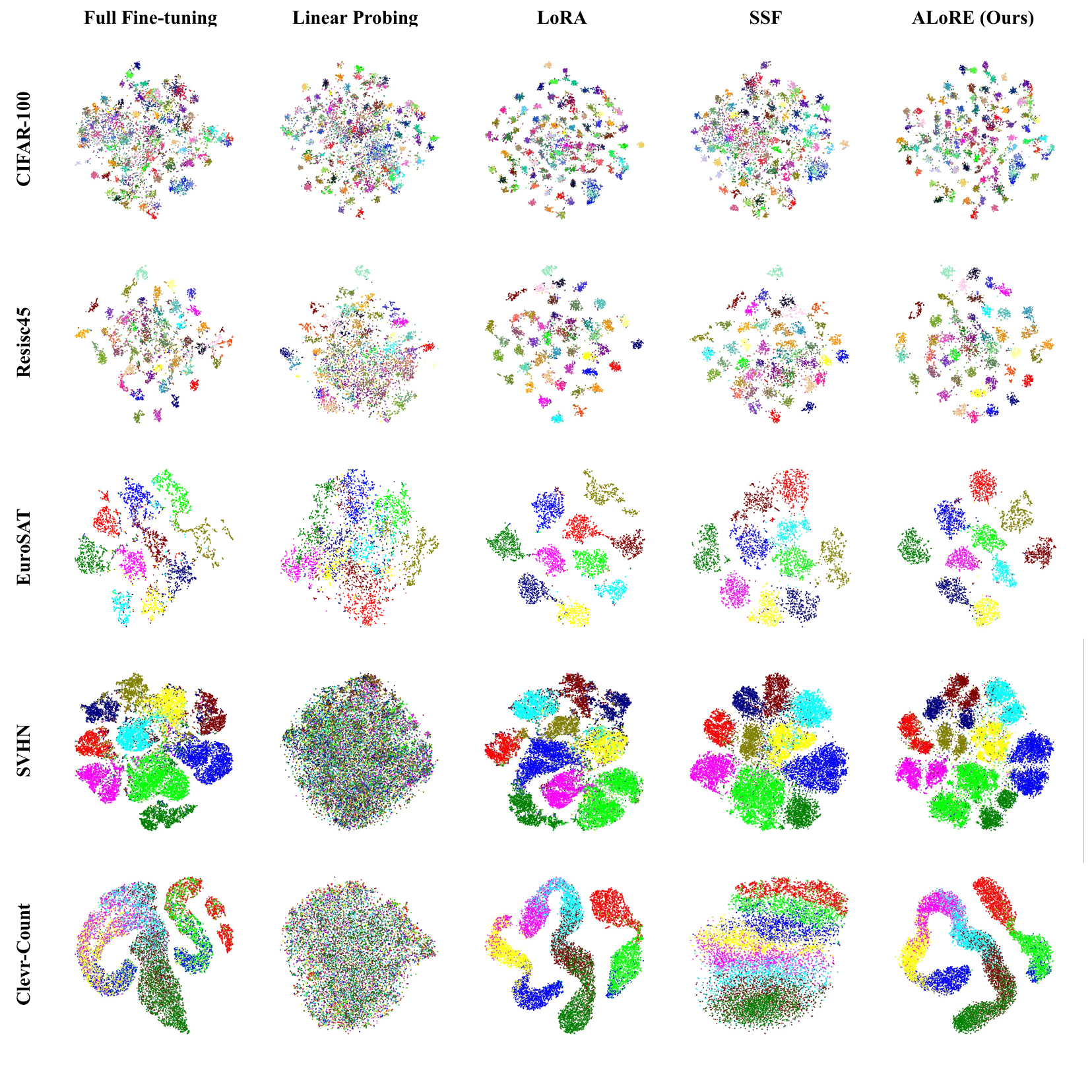}
    \caption{t-SNE visualizations of different fine-tuning methods on the VTAB-1k benchmark. We use the embeddings of [CLS] after the last transformer layer and before the classification head. All results are acquired using ViT-B/16 pre-trained on ImageNet-21K. Our ALoRE attains better manifolds and feature clustering results compared to other fine-tuning strategies.}
    \label{fig:tsne_1}
\end{figure}

To gain a more intuitive insight into the strategy of aggregating multiple low rank experts employed in our ALoRE method and to comprehend the role of individual experts in visual adaptation tasks, we visualize the attention maps in Figure~\ref{fig:gradcam} and identify the role of each expert in appendix Table~\ref{tab:app_rep}. We observe that in different tasks, each expert tends to focus on different edge regions and various main body parts of the visual entities. This observation aligns with the human cognitive process of perceiving a novel class entity, as different levels of visual sensory systems attend to distinct concepts. The last column in Figure~\ref{fig:gradcam} represents the attention map obtained by aggregating all the experts. Based on this, we can infer that the strategy of aggregating multiple experts enables the synthesis of different visual representation patterns obtained by each expert. By distributing these diverse representation patterns to different experts for learning during the training process, we can achieve a decoupling of learned features, thereby enhancing the performance of visual adaptation tasks. As shown in Figure~\ref{fig:tsne_1}, we visualize the learned manifolds and feature clustering of different fined-tuning methods on the VTAB-1k benchmark via the t-SNE~\cite{tsne} algorithm. Our method obtains better manifolds and feature clustering results. More visualization results are presented in appendix~\ref{app:vis}.
\section{Conclusion} \label{sec:conclusion}
In this paper, we introduce ALoRE, an efficient PETL approach that employs hypercomplex parameterized space constructed through the Kronecker Product for low rank factorization, and revisit visual adaptation from the perspective of decoupling learned representation patterns. Our method achieves improved performance over other state-of-the-art PETL methods by aggregating low rank experts in a multi-branch paradigm, while introducing negligible additional parameters. Due to the purely linear structure of the design, ALoRE can be re-parameterized to avoid extra inference latency. 
Extensive experiments on 24 downstream image classification tasks with backbones of various scales, model families, and pre-training strategies demonstrate the effectiveness and scalability of our ALoRE, building a solid baseline for re-parameterizable PETL methods. 
\clearpage
{
\small
\bibliographystyle{ieeenat_fullname}
\bibliography{main}

\begin{thebibliography}{81}
\providecommand{\natexlab}[1]{#1}
\providecommand{\url}[1]{\texttt{#1}}
\expandafter\ifx\csname urlstyle\endcsname\relax
  \providecommand{\doi}[1]{doi: #1}\else
  \providecommand{\doi}{doi: \begingroup \urlstyle{rm}\Url}\fi

\bibitem[Aghajanyan et~al.(2020)Aghajanyan, Zettlemoyer, and Gupta]{aghajanyan2020intrinsic}
Armen Aghajanyan, Luke Zettlemoyer, and Sonal Gupta.
\newblock Intrinsic dimensionality explains the effectiveness of language model fine-tuning.
\newblock \emph{arXiv preprint arXiv:2012.13255}, 2020.

\bibitem[Beattie et~al.(2016)Beattie, Leibo, Teplyashin, Ward, Wainwright, K{\"u}ttler, Lefrancq, Green, Vald{\'e}s, Sadik, et~al.]{dmlab}
Charles Beattie, Joel~Z Leibo, Denis Teplyashin, Tom Ward, Marcus Wainwright, Heinrich K{\"u}ttler, Andrew Lefrancq, Simon Green, V{\'\i}ctor Vald{\'e}s, Amir Sadik, et~al.
\newblock Deepmind lab.
\newblock \emph{arXiv preprint arXiv:1612.03801}, 2016.

\bibitem[Chavan et~al.(2023)Chavan, Liu, Gupta, Xing, and Shen]{glora}
Arnav Chavan, Zhuang Liu, Deepak Gupta, Eric Xing, and Zhiqiang Shen.
\newblock One-for-all: Generalized lora for parameter-efficient fine-tuning.
\newblock \emph{arXiv preprint arXiv:2306.07967}, 2023.

\bibitem[Chen et~al.(2022{\natexlab{a}})Chen, Tao, Zhang, Wang, Ye, Wang, Hu, and Savvides]{conv-adapter}
Hao Chen, Ran Tao, Han Zhang, Yidong Wang, Wei Ye, Jindong Wang, Guosheng Hu, and Marios Savvides.
\newblock Conv-adapter: Exploring parameter efficient transfer learning for convnets.
\newblock \emph{arXiv preprint arXiv:2208.07463}, 2022{\natexlab{a}}.

\bibitem[Chen et~al.(2022{\natexlab{b}})Chen, Ge, Tong, Wang, Song, Wang, and Luo]{adaptformer}
Shoufa Chen, Chongjian Ge, Zhan Tong, Jiangliu Wang, Yibing Song, Jue Wang, and Ping Luo.
\newblock Adaptformer: Adapting vision transformers for scalable visual recognition.
\newblock In \emph{Advances in Neural Information Processing Systems (NeurIPS)}, 2022{\natexlab{b}}.

\bibitem[Chen* et~al.(2021)Chen*, Xie*, and He]{mocov3}
Xinlei Chen*, Saining Xie*, and Kaiming He.
\newblock An empirical study of training self-supervised vision transformers.
\newblock \emph{arXiv preprint arXiv:2104.02057}, 2021.

\bibitem[Cheng et~al.(2017)Cheng, Han, and Lu]{resisc45}
Gong Cheng, Junwei Han, and Xiaoqiang Lu.
\newblock Remote sensing image scene classification: Benchmark and state of the art.
\newblock \emph{Proceedings of the IEEE}, pages 1865--1883, 2017.

\bibitem[Chung et~al.(2020)Chung, Fevry, Tsai, Johnson, and Ruder]{chung2020rethinking}
Hyung~Won Chung, Thibault Fevry, Henry Tsai, Melvin Johnson, and Sebastian Ruder.
\newblock Rethinking embedding coupling in pre-trained language models.
\newblock \emph{arXiv preprint arXiv:2010.12821}, 2020.

\bibitem[Cimpoi et~al.(2014)Cimpoi, Maji, Kokkinos, Mohamed, and Vedaldi]{dtd}
Mircea Cimpoi, Subhransu Maji, Iasonas Kokkinos, Sammy Mohamed, and Andrea Vedaldi.
\newblock Describing textures in the wild.
\newblock In \emph{Proceedings of the IEEE conference on computer vision and pattern recognition}, pages 3606--3613, 2014.

\bibitem[Deng et~al.(2009)Deng, Dong, Socher, Li, Li, and Fei-Fei]{imagenet}
Jia Deng, Wei Dong, Richard Socher, Li-Jia Li, Kai Li, and Li Fei-Fei.
\newblock Imagenet: A large-scale hierarchical image database.
\newblock In \emph{Proceedings of the IEEE/CVF Conference on Computer Vision and Pattern Recognition}, pages 248--255. Ieee, 2009.

\bibitem[Ding et~al.(2019)Ding, Guo, Ding, and Han]{acnet}
Xiaohan Ding, Yuchen Guo, Guiguang Ding, and Jungong Han.
\newblock Acnet: Strengthening the kernel skeletons for powerful cnn via asymmetric convolution blocks.
\newblock In \emph{Proceedings of the IEEE/CVF international conference on computer vision}, pages 1911--1920, 2019.

\bibitem[Ding et~al.(2021{\natexlab{a}})Ding, Xia, Zhang, Chu, Han, and Ding]{repmlp}
Xiaohan Ding, Chunlong Xia, Xiangyu Zhang, Xiaojie Chu, Jungong Han, and Guiguang Ding.
\newblock Repmlp: Re-parameterizing convolutions into fully-connected layers for image recognition.
\newblock \emph{arXiv preprint arXiv:2105.01883}, 2021{\natexlab{a}}.

\bibitem[Ding et~al.(2021{\natexlab{b}})Ding, Zhang, Han, and Ding]{dbbnet}
Xiaohan Ding, Xiangyu Zhang, Jungong Han, and Guiguang Ding.
\newblock Diverse branch block: Building a convolution as an inception-like unit.
\newblock In \emph{Proceedings of the IEEE/CVF conference on computer vision and pattern recognition}, pages 10886--10895, 2021{\natexlab{b}}.

\bibitem[Ding et~al.(2021{\natexlab{c}})Ding, Zhang, Ma, Han, Ding, and Sun]{repvgg}
Xiaohan Ding, Xiangyu Zhang, Ningning Ma, Jungong Han, Guiguang Ding, and Jian Sun.
\newblock Repvgg: Making vgg-style convnets great again.
\newblock In \emph{Proceedings of the IEEE/CVF conference on computer vision and pattern recognition}, pages 13733--13742, 2021{\natexlab{c}}.

\bibitem[Ding et~al.(2022)Ding, Chen, Zhang, Han, and Ding]{repmlpnet}
Xiaohan Ding, Honghao Chen, Xiangyu Zhang, Jungong Han, and Guiguang Ding.
\newblock Repmlpnet: Hierarchical vision mlp with re-parameterized locality.
\newblock In \emph{Proceedings of the IEEE/CVF conference on computer vision and pattern recognition}, pages 578--587, 2022.

\bibitem[Dong et~al.(2024)Dong, Yan, Lin, and Wang]{arc}
Wei Dong, Dawei Yan, Zhijun Lin, and Peng Wang.
\newblock Efficient adaptation of large vision transformer via adapter re-composing.
\newblock \emph{Advances in Neural Information Processing Systems}, 36, 2024.

\bibitem[Dosovitskiy et~al.(2021)Dosovitskiy, Beyer, Kolesnikov, Weissenborn, Zhai, Unterthiner, Dehghani, Minderer, Heigold, Gelly, Uszkoreit, and Houlsby]{vit}
Alexey Dosovitskiy, Lucas Beyer, Alexander Kolesnikov, Dirk Weissenborn, Xiaohua Zhai, Thomas Unterthiner, Mostafa Dehghani, Matthias Minderer, Georg Heigold, Sylvain Gelly, Jakob Uszkoreit, and Neil Houlsby.
\newblock An image is worth 16x16 words: Transformers for image recognition at scale.
\newblock \emph{ICLR}, 2021.

\bibitem[Fei-Fei et~al.(2006)Fei-Fei, Fergus, and Perona]{caltech101}
Li Fei-Fei, Robert Fergus, and Pietro Perona.
\newblock One-shot learning of object categories.
\newblock \emph{IEEE transactions on pattern analysis and machine intelligence}, pages 594--611, 2006.

\bibitem[Gebru et~al.(2017)Gebru, Krause, Wang, Chen, Deng, and Fei-Fei]{cars}
Timnit Gebru, Jonathan Krause, Yilun Wang, Duyun Chen, Jia Deng, and Li Fei-Fei.
\newblock Fine-grained car detection for visual census estimation.
\newblock In \emph{Proceedings of the AAAI Conference on Artificial Intelligence}, 2017.

\bibitem[Geiger et~al.(2013)Geiger, Lenz, Stiller, and Urtasun]{kitti}
Andreas Geiger, Philip Lenz, Christoph Stiller, and Raquel Urtasun.
\newblock Vision meets robotics: The kitti dataset.
\newblock \emph{The International Journal of Robotics Research}, pages 1231--1237, 2013.

\bibitem[Graham(2015)]{retinopathy}
Ben Graham.
\newblock Kaggle diabetic retinopathy detection competition report.
\newblock \emph{University of Warwick}, 22:\penalty0 17, 2015.

\bibitem[He et~al.(2016)He, Zhang, Ren, and Sun]{resnet}
Kaiming He, Xiangyu Zhang, Shaoqing Ren, and Jian Sun.
\newblock Deep residual learning for image recognition.
\newblock In \emph{Proceedings of the IEEE conference on computer vision and pattern recognition}, pages 770--778, 2016.

\bibitem[He et~al.(2020)He, Fan, Wu, Xie, and Girshick]{moco}
Kaiming He, Haoqi Fan, Yuxin Wu, Saining Xie, and Ross Girshick.
\newblock Momentum contrast for unsupervised visual representation learning.
\newblock In \emph{Proceedings of the IEEE/CVF conference on computer vision and pattern recognition}, pages 9729--9738, 2020.

\bibitem[He et~al.(2021)He, Chen, Xie, Li, Doll{\'a}r, and Girshick]{mae}
Kaiming He, Xinlei Chen, Saining Xie, Yanghao Li, Piotr Doll{\'a}r, and Ross Girshick.
\newblock Masked autoencoders are scalable vision learners.
\newblock \emph{arXiv:2111.06377}, 2021.

\bibitem[Helber et~al.(2019)Helber, Bischke, Dengel, and Borth]{eurosat}
Patrick Helber, Benjamin Bischke, Andreas Dengel, and Damian Borth.
\newblock Eurosat: A novel dataset and deep learning benchmark for land use and land cover classification.
\newblock \emph{IEEE Journal of Selected Topics in Applied Earth Observations and Remote Sensing}, pages 2217--2226, 2019.

\bibitem[Henderson et~al.(2021)Henderson, Ruder, et~al.]{compacter}
James Henderson, Sebastian Ruder, et~al.
\newblock Compacter: Efficient low-rank hypercomplex adapter layers.
\newblock In \emph{Advances in Neural Information Processing Systems}, 2021.

\bibitem[Houlsby et~al.(2019)Houlsby, Giurgiu, Jastrzebski, Morrone, De~Laroussilhe, Gesmundo, Attariyan, and Gelly]{adapter}
Neil Houlsby, Andrei Giurgiu, Stanislaw Jastrzebski, Bruna Morrone, Quentin De~Laroussilhe, Andrea Gesmundo, Mona Attariyan, and Sylvain Gelly.
\newblock Parameter-efficient transfer learning for nlp.
\newblock In \emph{International Conference on Machine Learning}, pages 2790--2799. PMLR, 2019.

\bibitem[Hu et~al.(2021)Hu, Shen, Wallis, Allen-Zhu, Li, Wang, Wang, and Chen]{lora}
Edward~J Hu, Yelong Shen, Phillip Wallis, Zeyuan Allen-Zhu, Yuanzhi Li, Shean Wang, Lu Wang, and Weizhu Chen.
\newblock Lora: Low-rank adaptation of large language models.
\newblock \emph{arXiv preprint arXiv:2106.09685}, 2021.

\bibitem[Jia et~al.(2022)Jia, Tang, Chen, Cardie, Belongie, Hariharan, and Lim]{vpt}
Menglin Jia, Luming Tang, Bor-Chun Chen, Claire Cardie, Serge Belongie, Bharath Hariharan, and Ser-Nam Lim.
\newblock Visual prompt tuning.
\newblock In \emph{Proceedings of the European conference on computer vision (ECCV)}, pages 709--727. Springer, 2022.

\bibitem[Jiang et~al.(2023)Jiang, Mao, Huang, Ma, Lv, Shen, Zhao, and Zhou]{restuning}
Zeyinzi Jiang, Chaojie Mao, Ziyuan Huang, Ao Ma, Yiliang Lv, Yujun Shen, Deli Zhao, and Jingren Zhou.
\newblock Res-tuning: A flexible and efficient tuning paradigm via unbinding tuner from backbone.
\newblock In \emph{Advances in Neural Information Processing Systems}, 2023.

\bibitem[Jie and Deng(2022)]{convpass}
Shibo Jie and Zhi-Hong Deng.
\newblock Convolutional bypasses are better vision transformer adapters.
\newblock \emph{arXiv preprint arXiv:2207.07039}, 2022.

\bibitem[Johnson et~al.(2017)Johnson, Hariharan, Van Der~Maaten, Fei-Fei, Lawrence~Zitnick, and Girshick]{clevr}
Justin Johnson, Bharath Hariharan, Laurens Van Der~Maaten, Li Fei-Fei, C Lawrence~Zitnick, and Ross Girshick.
\newblock Clevr: A diagnostic dataset for compositional language and elementary visual reasoning.
\newblock In \emph{Proceedings of the IEEE conference on computer vision and pattern recognition}, pages 2901--2910, 2017.

\bibitem[Khosla et~al.(2011)Khosla, Jayadevaprakash, Yao, and Li]{dogs}
Aditya Khosla, Nityananda Jayadevaprakash, Bangpeng Yao, and Fei-Fei Li.
\newblock Novel dataset for fine-grained image categorization: Stanford dogs.
\newblock In \emph{Proc. CVPR workshop on fine-grained visual categorization (FGVC)}. Citeseer, 2011.

\bibitem[Krizhevsky et~al.(2009)Krizhevsky, Hinton, et~al.]{cifar100}
Alex Krizhevsky, Geoffrey Hinton, et~al.
\newblock Learning multiple layers of features from tiny images.
\newblock 2009.

\bibitem[Kuehne et~al.(2011)Kuehne, Jhuang, Garrote, Poggio, and Serre]{hmdb51}
Hildegard Kuehne, Hueihan Jhuang, Est{\'\i}baliz Garrote, Tomaso Poggio, and Thomas Serre.
\newblock Hmdb: a large video database for human motion recognition.
\newblock In \emph{2011 International conference on computer vision}, pages 2556--2563. IEEE, 2011.

\bibitem[LeCun et~al.(2004)LeCun, Huang, and Bottou]{snor}
Yann LeCun, Fu~Jie Huang, and Leon Bottou.
\newblock Learning methods for generic object recognition with invariance to pose and lighting.
\newblock In \emph{Proceedings of the 2004 IEEE Computer Society Conference on Computer Vision and Pattern Recognition, 2004. CVPR 2004.}, pages II--104. IEEE, 2004.

\bibitem[Lester et~al.(2021)Lester, Al-Rfou, and Constant]{power}
Brian Lester, Rami Al-Rfou, and Noah Constant.
\newblock The power of scale for parameter-efficient prompt tuning.
\newblock In \emph{Proceedings of the 2021 Conference on Empirical Methods in Natural Language Processing}, pages 3045--3059, 2021.

\bibitem[Li et~al.(2018)Li, Farkhoor, Liu, and Yosinski]{li2018measuring}
Chunyuan Li, Heerad Farkhoor, Rosanne Liu, and Jason Yosinski.
\newblock Measuring the intrinsic dimension of objective landscapes.
\newblock \emph{arXiv preprint arXiv:1804.08838}, 2018.

\bibitem[Li and Liang(2021)]{prefix}
Xiang~Lisa Li and Percy Liang.
\newblock Prefix-tuning: Optimizing continuous prompts for generation.
\newblock \emph{arXiv preprint arXiv:2101.00190}, 2021.

\bibitem[Lian et~al.(2021)Lian, Yu, Sun, and Gao]{asmlp}
Dongze Lian, Zehao Yu, Xing Sun, and Shenghua Gao.
\newblock As-mlp: An axial shifted mlp architecture for vision.
\newblock \emph{arXiv preprint arXiv:2107.08391}, 2021.

\bibitem[Lian et~al.(2022)Lian, Zhou, Feng, and Wang]{ssf}
Dongze Lian, Daquan Zhou, Jiashi Feng, and Xinchao Wang.
\newblock Scaling \& shifting your features: A new baseline for efficient model tuning.
\newblock In \emph{Advances in Neural Information Processing Systems (NeurIPS)}, 2022.

\bibitem[Liu et~al.(2023)Liu, Yuan, Fu, Jiang, Hayashi, and Neubig]{pre}
Pengfei Liu, Weizhe Yuan, Jinlan Fu, Zhengbao Jiang, Hiroaki Hayashi, and Graham Neubig.
\newblock Pre-train, prompt, and predict: A systematic survey of prompting methods in natural language processing.
\newblock \emph{ACM Computing Surveys}, pages 1--35, 2023.

\bibitem[Liu et~al.(2021)Liu, Lin, Cao, Hu, Wei, Zhang, Lin, and Guo]{swin}
Ze Liu, Yutong Lin, Yue Cao, Han Hu, Yixuan Wei, Zheng Zhang, Stephen Lin, and Baining Guo.
\newblock Swin transformer: Hierarchical vision transformer using shifted windows.
\newblock In \emph{Proceedings of the IEEE/CVF International Conference on Computer Vision (ICCV)}, 2021.

\bibitem[Liu et~al.(2022{\natexlab{a}})Liu, Hu, Lin, Yao, Xie, Wei, Ning, Cao, Zhang, Dong, et~al.]{swinv2}
Ze Liu, Han Hu, Yutong Lin, Zhuliang Yao, Zhenda Xie, Yixuan Wei, Jia Ning, Yue Cao, Zheng Zhang, Li Dong, et~al.
\newblock Swin transformer v2: Scaling up capacity and resolution.
\newblock In \emph{Proceedings of the IEEE/CVF conference on computer vision and pattern recognition}, pages 12009--12019, 2022{\natexlab{a}}.

\bibitem[Liu et~al.(2022{\natexlab{b}})Liu, Mao, Wu, Feichtenhofer, Darrell, and Xie]{convnet}
Zhuang Liu, Hanzi Mao, Chao-Yuan Wu, Christoph Feichtenhofer, Trevor Darrell, and Saining Xie.
\newblock A convnet for the 2020s.
\newblock \emph{Proceedings of the IEEE/CVF Conference on Computer Vision and Pattern Recognition (CVPR)}, 2022{\natexlab{b}}.

\bibitem[Loshchilov and Hutter(2017)]{adamw}
Ilya Loshchilov and Frank Hutter.
\newblock Decoupled weight decay regularization.
\newblock \emph{Learning,Learning}, 2017.

\bibitem[Luo et~al.(2023)Luo, Huang, Zhou, Sun, Jiang, Wang, and Ji]{rep}
Gen Luo, Minglang Huang, Yiyi Zhou, Xiaoshuai Sun, Guannan Jiang, Zhiyu Wang, and Rongrong Ji.
\newblock Towards efficient visual adaption via structural re-parameterization.
\newblock \emph{arXiv preprint arXiv:2302.08106}, 2023.

\bibitem[Matthey et~al.(2017)Matthey, Higgins, Hassabis, and Lerchner]{dsprites}
Loic Matthey, Irina Higgins, Demis Hassabis, and Alexander Lerchner.
\newblock dsprites: Disentanglement testing sprites dataset, 2017.

\bibitem[Mercea et~al.(2024)Mercea, Gritsenko, Schmid, and Arnab]{losa}
Otniel-Bogdan Mercea, Alexey Gritsenko, Cordelia Schmid, and Anurag Arnab.
\newblock Time-, memory-and parameter-efficient visual adaptation.
\newblock \emph{arXiv preprint arXiv:2402.02887}, 2024.

\bibitem[Netzer et~al.(2011)Netzer, Wang, Coates, Bissacco, Wu, Ng, et~al.]{svhn}
Yuval Netzer, Tao Wang, Adam Coates, Alessandro Bissacco, Baolin Wu, Andrew~Y Ng, et~al.
\newblock Reading digits in natural images with unsupervised feature learning.
\newblock In \emph{NIPS workshop on deep learning and unsupervised feature learning}, page~7. Granada, Spain, 2011.

\bibitem[Nilsback and Zisserman(2008{\natexlab{a}})]{flowers}
Maria-Elena Nilsback and Andrew Zisserman.
\newblock Automated flower classification over a large number of classes.
\newblock In \emph{2008 Sixth Indian Conference on Computer Vision, Graphics \& Image Processing}, pages 722--729. IEEE, 2008{\natexlab{a}}.

\bibitem[Nilsback and Zisserman(2008{\natexlab{b}})]{flowers102}
Maria-Elena Nilsback and Andrew Zisserman.
\newblock Automated flower classification over a large number of classes.
\newblock In \emph{2008 Sixth Indian conference on computer vision, graphics \& image processing}, pages 722--729. IEEE, 2008{\natexlab{b}}.

\bibitem[Parkhi et~al.(2012)Parkhi, Vedaldi, Zisserman, and Jawahar]{pets}
Omkar~M Parkhi, Andrea Vedaldi, Andrew Zisserman, and CV Jawahar.
\newblock Cats and dogs.
\newblock In \emph{2012 IEEE conference on computer vision and pattern recognition}, pages 3498--3505. IEEE, 2012.

\bibitem[Pfeiffer et~al.(2020)Pfeiffer, Kamath, R{\"u}ckl{\'e}, Cho, and Gurevych]{adapterfusion}
Jonas Pfeiffer, Aishwarya Kamath, Andreas R{\"u}ckl{\'e}, Kyunghyun Cho, and Iryna Gurevych.
\newblock Adapterfusion: Non-destructive task composition for transfer learning.
\newblock \emph{arXiv preprint arXiv:2005.00247}, 2020.

\bibitem[Ridnik et~al.(2021)Ridnik, Ben-Baruch, Noy, and Zelnik-Manor]{imagenet21k}
Tal Ridnik, Emanuel Ben-Baruch, Asaf Noy, and Lihi Zelnik-Manor.
\newblock Imagenet-21k pretraining for the masses.
\newblock In \emph{Advances in Neural Information Processing Systems (NeurIPS)}, 2021.

\bibitem[R{\"u}ckl{\'e} et~al.(2020)R{\"u}ckl{\'e}, Geigle, Glockner, Beck, Pfeiffer, Reimers, and Gurevych]{adapterdrop}
Andreas R{\"u}ckl{\'e}, Gregor Geigle, Max Glockner, Tilman Beck, Jonas Pfeiffer, Nils Reimers, and Iryna Gurevych.
\newblock Adapterdrop: On the efficiency of adapters in transformers.
\newblock \emph{arXiv preprint arXiv:2010.11918}, 2020.

\bibitem[Shin et~al.(2020)Shin, Razeghi, Logan~IV, Wallace, and Singh]{autoprompt}
Taylor Shin, Yasaman Razeghi, Robert~L Logan~IV, Eric Wallace, and Sameer Singh.
\newblock Autoprompt: Eliciting knowledge from language models with automatically generated prompts.
\newblock \emph{arXiv preprint arXiv:2010.15980}, 2020.

\bibitem[Srivastava et~al.(2014)Srivastava, Hinton, Krizhevsky, Sutskever, and Salakhutdinov]{dropout}
Nitish Srivastava, Geoffrey Hinton, Alex Krizhevsky, Ilya Sutskever, and Ruslan Salakhutdinov.
\newblock Dropout: a simple way to prevent neural networks from overfitting.
\newblock \emph{The journal of machine learning research}, pages 1929--1958, 2014.

\bibitem[Sung et~al.(2022)Sung, Cho, and Bansal]{lst}
Yi-Lin Sung, Jaemin Cho, and Mohit Bansal.
\newblock Lst: Ladder side-tuning for parameter and memory efficient transfer learning.
\newblock \emph{Advances in Neural Information Processing Systems}, 35:\penalty0 12991--13005, 2022.

\bibitem[Tolstikhin et~al.(2021)Tolstikhin, Houlsby, Kolesnikov, Beyer, Zhai, Unterthiner, Yung, Steiner, Keysers, Uszkoreit, Lucic, and Dosovitskiy]{mlp-mixer}
Ilya Tolstikhin, Neil Houlsby, Alexander Kolesnikov, Lucas Beyer, Xiaohua Zhai, Thomas Unterthiner, Jessica Yung, Andreas Steiner, Daniel Keysers, Jakob Uszkoreit, Mario Lucic, and Alexey Dosovitskiy.
\newblock Mlp-mixer: An all-mlp architecture for vision.
\newblock \emph{arXiv preprint arXiv:2105.01601}, 2021.

\bibitem[Tong et~al.(2022)Tong, Song, Wang, and Wang]{videomae}
Zhan Tong, Yibing Song, Jue Wang, and Limin Wang.
\newblock Videomae: Masked autoencoders are data-efficient learners for self-supervised video pre-training.
\newblock \emph{Advances in neural information processing systems}, 35:\penalty0 10078--10093, 2022.

\bibitem[Touvron et~al.(2021)Touvron, Cord, Douze, Massa, Sablayrolles, and J{\'e}gou]{deit}
Hugo Touvron, Matthieu Cord, Matthijs Douze, Francisco Massa, Alexandre Sablayrolles, and Herv{\'e} J{\'e}gou.
\newblock Training data-efficient image transformers \& distillation through attention.
\newblock In \emph{International conference on machine learning}, pages 10347--10357. PMLR, 2021.

\bibitem[Van~der Maaten and Hinton(2008)]{tsne}
Laurens Van~der Maaten and Geoffrey Hinton.
\newblock Visualizing data using t-sne.
\newblock \emph{Journal of machine learning research}, 9\penalty0 (11), 2008.

\bibitem[Van~Horn et~al.(2015)Van~Horn, Branson, Farrell, Haber, Barry, Ipeirotis, Perona, and Belongie]{birds}
Grant Van~Horn, Steve Branson, Ryan Farrell, Scott Haber, Jessie Barry, Panos Ipeirotis, Pietro Perona, and Serge Belongie.
\newblock Building a bird recognition app and large scale dataset with citizen scientists: The fine print in fine-grained dataset collection.
\newblock In \emph{Proceedings of the IEEE Conference on Computer Vision and Pattern Recognition}, pages 595--604, 2015.

\bibitem[Vaswani(2017)]{transformer}
A Vaswani.
\newblock Attention is all you need.
\newblock \emph{Advances in Neural Information Processing Systems}, 2017.

\bibitem[Veeling et~al.(2018)Veeling, Linmans, Winkens, Cohen, and Welling]{camelyon}
Bastiaan~S Veeling, Jasper Linmans, Jim Winkens, Taco Cohen, and Max Welling.
\newblock Rotation equivariant cnns for digital pathology.
\newblock In \emph{Medical Image Computing and Computer Assisted Intervention--MICCAI 2018: 21st International Conference, Granada, Spain, September 16-20, 2018, Proceedings, Part II 11}, pages 210--218. Springer, 2018.

\bibitem[Wah et~al.(2011)Wah, Branson, Welinder, Perona, and Belongie]{cub}
Catherine Wah, Steve Branson, Peter Welinder, Pietro Perona, and Serge Belongie.
\newblock The caltech-ucsd birds-200-2011 dataset.
\newblock 2011.

\bibitem[Wen et~al.(2020)Wen, Tran, and Ba]{wen2020batchensemble}
Yeming Wen, Dustin Tran, and Jimmy Ba.
\newblock Batchensemble: an alternative approach to efficient ensemble and lifelong learning.
\newblock \emph{arXiv preprint arXiv:2002.06715}, 2020.

\bibitem[Xiao et~al.(2010)Xiao, Hays, Ehinger, Oliva, and Torralba]{sun397}
Jianxiong Xiao, James Hays, Krista~A Ehinger, Aude Oliva, and Antonio Torralba.
\newblock Sun database: Large-scale scene recognition from abbey to zoo.
\newblock In \emph{2010 IEEE computer society conference on computer vision and pattern recognition}, pages 3485--3492. IEEE, 2010.

\bibitem[Yin et~al.(2023)Yin, Li, and Zhang]{mona}
Dongshuo Yin, Leiyi Hu~Bin Li, and Youqun Zhang.
\newblock Adapter is all you need for tuning visual tasks.
\newblock \emph{arXiv preprint arXiv:2311.15010}, 2023.

\bibitem[Zagoruyko and Komodakis(2017)]{diracnets}
Sergey Zagoruyko and Nikos Komodakis.
\newblock Diracnets: Training very deep neural networks without skip-connections.
\newblock \emph{arXiv preprint arXiv:1706.00388}, 2017.

\bibitem[Zaken et~al.(2022)Zaken, Goldberg, and Ravfogel]{bitfit}
Elad~Ben Zaken, Yoav Goldberg, and Shauli Ravfogel.
\newblock Bitfit: Simple parameter-efficient fine-tuning for transformer-based masked language-models.
\newblock In \emph{Proceedings of the 60th Annual Meeting of the Association for Computational Linguistics (Volume 2: Short Papers)}, pages 1--9, 2022.

\bibitem[Zhai et~al.(2019)Zhai, Puigcerver, Kolesnikov, Ruyssen, Riquelme, Lucic, Djolonga, Pinto, Neumann, Dosovitskiy, et~al.]{vtab}
Xiaohua Zhai, Joan Puigcerver, Alexander Kolesnikov, Pierre Ruyssen, Carlos Riquelme, Mario Lucic, Josip Djolonga, Andre~Susano Pinto, Maxim Neumann, Alexey Dosovitskiy, et~al.
\newblock A large-scale study of representation learning with the visual task adaptation benchmark.
\newblock \emph{arXiv preprint arXiv:1910.04867}, 2019.

\bibitem[Zhai et~al.(2022)Zhai, Kolesnikov, Houlsby, and Beyer]{scaling}
Xiaohua Zhai, Alexander Kolesnikov, Neil Houlsby, and Lucas Beyer.
\newblock Scaling vision transformers.
\newblock In \emph{Proceedings of the IEEE/CVF Conference on Computer Vision and Pattern Recognition}, pages 12104--12113, 2022.

\bibitem[Zhang et~al.(2021)Zhang, Tay, Zhang, Chan, Luu, Hui, and Fu]{zhang2021beyond}
Aston Zhang, Yi Tay, Shuai Zhang, Alvin Chan, Anh~Tuan Luu, Siu~Cheung Hui, and Jie Fu.
\newblock Beyond fully-connected layers with quaternions: Parameterization of hypercomplex multiplications with $1/n $ parameters.
\newblock \emph{arXiv preprint arXiv:2102.08597}, 2021.

\bibitem[Zhang et~al.(2023)Zhang, Chen, Bukharin, He, Cheng, Chen, and Zhao]{adalora}
Qingru Zhang, Minshuo Chen, Alexander Bukharin, Pengcheng He, Yu Cheng, Weizhu Chen, and Tuo Zhao.
\newblock Adaptive budget allocation for parameter-efficient fine-tuning.
\newblock In \emph{The Eleventh International Conference on Learning Representations}, 2023.

\bibitem[Zhang et~al.(2020)Zhang, Wu, Katiyar, Weinberger, and Artzi]{zhang2020revisiting}
Tianyi Zhang, Felix Wu, Arzoo Katiyar, Kilian~Q Weinberger, and Yoav Artzi.
\newblock Revisiting few-sample bert fine-tuning.
\newblock \emph{arXiv preprint arXiv:2006.05987}, 2020.

\bibitem[Zhang et~al.(2022)Zhang, Zhou, and Liu]{noah}
Yuanhan Zhang, Kaiyang Zhou, and Ziwei Liu.
\newblock Neural prompt search.
\newblock \emph{arXiv preprint arXiv:2206.04673}, 2022.

\bibitem[Zheng et~al.(2021)Zheng, Lu, Zhao, Zhu, Luo, Wang, Fu, Feng, Xiang, Torr, et~al.]{setr}
Sixiao Zheng, Jiachen Lu, Hengshuang Zhao, Xiatian Zhu, Zekun Luo, Yabiao Wang, Yanwei Fu, Jianfeng Feng, Tao Xiang, Philip~HS Torr, et~al.
\newblock Rethinking semantic segmentation from a sequence-to-sequence perspective with transformers.
\newblock In \emph{Proceedings of the IEEE/CVF conference on computer vision and pattern recognition}, pages 6881--6890, 2021.

\bibitem[Zhong et~al.(2024)Zhong, Tang, He, Fang, and Yuan]{conv-lora}
Zihan Zhong, Zhiqiang Tang, Tong He, Haoyang Fang, and Chun Yuan.
\newblock Convolution meets lora: Parameter efficient finetuning for segment anything model.
\newblock \emph{ICLR}, 2024.

\bibitem[Zhou et~al.(2019)Zhou, Zhao, Puig, Xiao, Fidler, Barriuso, and Torralba]{ade20k}
Bolei Zhou, Hang Zhao, Xavier Puig, Tete Xiao, Sanja Fidler, Adela Barriuso, and Antonio Torralba.
\newblock Semantic understanding of scenes through the ade20k dataset.
\newblock \emph{International Journal of Computer Vision}, 127:\penalty0 302--321, 2019.

\end{thebibliography}
}

\clearpage
\addtocontents{toc}{\protect\setcounter{tocdepth}{2}} 
\clearpage
\onecolumn
\appendix

\begin{center}
    \large \textbf{ALoRE: Efficient Visual Adaptation via Aggregating Low Rank Experts} \\
    \large \textbf{Supplementary Materials} \\
\end{center}
\vspace{30pt}

\begin{center}
\toccolor
\begin{minipage}[t]{0.95\textwidth} 
\begingroup
\renewcommand{\cftsecleader}{\cftdotfill{\cftdotsep}}
\setlength{\cftsecnumwidth}{3em}
\setlength{\cftsecindent}{1.5em}
\setlength{\cftparskip}{5pt} 
\setlength{\cftbeforesecskip}{5pt} 
\cftsetindents{section}{0em}{1.7em}
\setcounter{tocdepth}{2}
\tableofcontents 
\endgroup
\end{minipage}
\defaultcolor
\end{center}
\clearpage

\section{Explanation about Common Perplexity}
\label{app:explain}

\subsection{Discrepancy with LoRA-like Approaches}
\label{app:diff_lora}

It is worth noting that the positions of our ALoRE and that of LoRA~\cite{lora} are distinct. LoRA operates in \textbf{parallel}, acting directly on the original parameters of the MHSA. In contrast, ALoRE functions as a \textbf{completely independent module}, interfacing with MHSA in a \textbf{sequential} manner. This distinction is evident in the details of the re-parameterization process. Whereas LoRA employs an \textbf{additive} approach, updating the weights as $W + \Delta W$, ALoRE utilizes a \textbf{multiplicative} formulation, modifying \textbf{not} the weight update $\Delta W$, but rather the ALoRE-independent weights $W_{ALoRE}$, which are then multiplied with $W$, namely $W_{ALoRE} * W$. This \textbf{sequential} versus \textbf{parallel} re-parameterization is an intrinsic difference that underlies the fact that ALoRE is \textit{more akin to an adapter-based approach rather than being LoRA-like} in nature.

\subsection{Inferior Performance of ViT-Huge Compared to ViT-Large}
\label{app:diff_huge_large}

A key premise of the experimental setup that we wish to emphasize is that the VTAB-1k training dataset consists of \textbf{only 1k images}, thereby situating the visual domain PETL task primarily in a few-shot learning context. This contrasts significantly with the benchmark selection for the NLP-oriented PEFT methods. A careful comparison of Table \ref{tab:vit-large} and ~\ref{tab:vit-huge} in our paper reveals that the performance of nearly all methods is inferior on the ViT-Huge backbone compared to ViT-Large. Notably, the number of trainable parameters in these methods increases with the number of layers, suggesting a potential overfitting issue stemming from model capacity expansion. This phenomenon aligns with the findings depicted in Fig. 4 of the Visual Prompt Tuning (VPT~\cite{vpt}) paper.  Our experimental setup has been meticulously aligned with prior work to ensure a fair comparison, and we have endeavored to rely on previously reported experimental results rather than conducting from-scratch reproductions to maintain reliability. This approach has led us to the conclusion that our method exhibits greater scalability. In contrast, a more comprehensive comparison of the results in Table~\ref{tab:app_large} and ~\ref{tab:app_huge} of our paper reveals that methods with learnable parameters independent of layers, such as VPT-Shallow~\cite{vpt}, or those with inherently few learnable parameters, such as BitFit~\cite{bitfit}, typically tend to less exhibit this pronounced performance degradation.

\clearpage
\section{Detailed Descriptions for the Evaluation Datasets and Setups}
\label{app:datasets_and_impl}

\subsection{Evaluation Datasets}
\label{app:datasets}

In Table~\ref{tab:app_dataset}, we present the details of the visual adaptation datasets we utilize in our experiments, including the number of classes and train/val/test splits.

\begin{table*}[h]
	\small
	\begin{center}
         \caption{The statistics of the FGVC datasets and VTAB-1k benchmark. $^{\star}$: Since there are no public train/val splits in these datasets, we follow VPT \cite{vpt} for random train/val split.}
		\scalebox{0.94}{%
			\begin{tabular}{l  |  l  |  l  | l |  l |  l}
				\toprule[1pt]
				\textbf{Dataset}   & \textbf{Description}  & \textbf{Classes}    & \textbf{Train size} & \textbf{Val size}  & \textbf{Test size} \\ 
				\midrule
				\multicolumn{6}{c}{Fine-Grained Visual Classification (FGVC)}  \\
				\cmidrule{1-6}
				CUB-200-2011~\cite{cub} & Fine-grained bird species recognition &200 &5,394$^{\star}$	&600$^{\star}$ &5,794	\\
				NABirds~\cite{birds} & Fine-grained bird species recognition &555 &21,536$^{\star}$	&2,393$^{\star}$	&24,633	\\
				Oxford Flowers~\cite{flowers}& Fine-grained flower species recognition &102 &1,020	&1,020	&6,149  \\
				Stanford Dogs~\cite{dogs}  &Fine-grained dog species recognition  &120  &10,800$^{\star}$	&1,200$^{\star}$ &8,580 \\
				Stanford Cars~\cite{cars} & Fine-grained car classification  &196   &7,329$^{\star}$	&815$^{\star}$	&8,041 \\
				\midrule
				\multicolumn{6}{c}{Visual Task Adaptation Benchmark (VTAB-1k)~\cite{vtab}} \\
				\cmidrule{1-6}
				CIFAR-100~\cite{cifar100} &\multirow{7}{*}{Natural} &100 &\multirow{7}{*}{800/1,000} &\multirow{7}{*}{200} &10,000 \\
				Caltech101~\cite{caltech101} & &102 && &6,084  \\
				DTD~\cite{dtd} & &47 && &1,880  \\
				Flowers102~\cite{flowers102} & &102 && &6,149	\\
				Pets~\cite{pets} & &37 && &3,669	\\
				SVHN~\cite{svhn} & &10 && &26,032	\\
				Sun397~\cite{sun397} & &397 && &21,750	\\
				\cmidrule{1-6}
				Patch Camelyon~\cite{camelyon} &\multirow{4}{*}{Specialized} &2 &\multirow{4}{*}{800/1,000} &\multirow{4}{*}{200} &32,768 	\\
				EuroSAT~\cite{eurosat} & &10 && &5,400		\\
				Resisc45~\cite{resisc45} & &45 && &6,300		\\
				Retinopathy~\cite{retinopathy} & &5 && &42,670		\\
				\cmidrule{1-6}
				Clevr/count~\cite{clevr} &\multirow{8}{*}{Structured} &8	& \multirow{8}{*}{800/1,000} & \multirow{8}{*}{200} &15,000			\\
				Clevr/distance~\cite{clevr} & &6 && &15,000		\\
				DMLab~\cite{dmlab} & & 6 & & &22,735			\\	
				KITTI/distance~\cite{kitti} & &4 && &711		\\  
				dSprites/location~\cite{dsprites} & &16 & & &73,728		\\
				dSprites/orientation~\cite{dsprites} & &16 && &73,728		\\
				SmallNORB/azimuth~\cite{snor} & &18 && &12,150		\\
				SmallNORB/elevation~\cite{snor} & &9 && &12,150	\\
				\bottomrule[1pt]
			\end{tabular}
		}
		
		\label{tab:app_dataset}
	\end{center}
        \vspace{-20pt}
\end{table*}

\subsection{Implementation}
\label{app:impl}

In Table~\ref{tab:app_train}, we provide the detailed configurations in our experiments. We perform grid search to select hyper-parameters such as learning rate, weight decay, dropout~\cite{dropout}, using the validation set of each task as in VPT~\cite{vpt}. Note that we use dropout~\cite{dropout} in our method before adding the residual~\cite{resnet}.

\begin{table}[htbp]
  \centering
  \caption{The implementation details of our configurations.}
  \resizebox{0.8\linewidth}{!}{
  \setlength{\tabcolsep}{30pt}
    \begin{tabular}{c|c}
    \toprule[1pt]
    Optimizer & AdamW \\
    Learning Rate & \{0.05, 0.01, 0.005, 0.001\} \\
    Weight Decay & \{0.05, 0.01, 0.005, 0.001, 0\} \\
    Batch Size & 32 \\
    Adapter Dropout & \{0.1, 0\} \\
    Learning Rate Schedule & Cosine Decay \\
    Training Epochs  & 100 \\
    Warmup Epochs & 10 \\
    \bottomrule[1pt]
    \end{tabular}
    }
  \label{tab:app_train}
\end{table}

\subsection{Specifications of Different Pre-trained Backbones}
\label{app:weights}

In Table~\ref{tab:weights}, We provide the list of pre-trained backbones used in our paper. To evaluate the scalability, we select ViT-B/16, ViT-L/16, and ViT-H/14 pre-trained on ImageNet-21K. To evaluate the versatility, we choose different backbone regimes, namely Swin-B, ConvNeXt-B, and AS-MLP-B. Regarding the pre-training strategies, we use ViT-B/16 pre-trained via MoCo-v3 and MAE, which belong to contrastive learning and masked image modeling strategies.

\setlength{\tabcolsep}{4pt}
\begin{table*}[h]
\small
\centering
\caption{Specifications of different pre-trained backbones used in our paper.}
\resizebox{0.8\textwidth}{!}{%
\setlength{\tabcolsep}{20pt}
\label{tab:weights}
\begin{tabular}{c  |  c  |  c  | c }
\toprule[1pt]
\textbf{Backbone} & \textbf{Pre-trained Objective} & \textbf{Pre-trained Dataset} & \textbf{Params.(M)} \\ & \textbf{Checkpoint} \\ 
\midrule
ViT-B/16~\cite{vit} & \multirow{3}{*}{Supervised} & \multirow{3}{*}{ImageNet-21K~\cite{imagenet21k}} & 85 \\ 
ViT-L/16~\cite{vit} &  &  & 307 \\ 
ViT-H/14~\cite{vit} &  &  & 630 \\ 
\midrule
\multirow{2}{*}{ViT-B/16~\cite{vit}} & MoCo-v3~\cite{mocov3} & \multirow{2}{*}{ImageNet-1K~\cite{imagenet}} & \multirow{2}{*}{85} \\ 
 & MAE~\cite{mae} &  &  \\ 
\midrule
Swin-B~\cite{swin} & Supervised & ImageNet-21K~\cite{imagenet21k} & 88 \\ 
\midrule
ConvNeXt-B~\cite{convnet} & Supervised & ImageNet-21K~\cite{imagenet21k} & 88 \\ 
\midrule
AS-MLP-B~\cite{asmlp} & Supervised & ImageNet-21K~\cite{imagenet21k} & 88 \\ 
\bottomrule[1pt]
\end{tabular}
}
\end{table*}

\section{Parameter Size Analysis}
\label{app:param}
In order to further elucidate the incorporation of multiple low rank experts through a multi-branch architecture while introducing negligible parameters, we have undertaken a comparative analysis of parameter magnitudes among several popular visual adaptation methods in Table~\ref{tab:app_size}. Adapter~\cite{adapter} and AdaptFormer~\cite{adaptformer} add two linear projections to each sub-layer (MHSA and FFN) during fine-tuning in a sequential and parallel manner respectively, resulting in the introduction of $4 \cdot d \cdot r \cdot L$ trainable parameters, where $d$ denotes the dimension of the model, $r$ denotes the bottleneck dimension, $L$ represents the number of layers. Due to the non-linear activation function in ~\cite{adaptformer, adapter}, it incurs equivalent overhead during inference. VPT~\cite{vpt} incorporates $m$ prompts into the input space, introducing extra $m \cdot d$ parameters for VPT-Shallow and $m \cdot d \cdot L$ parameters for VPT-Deep. LoRA~\cite{lora} introduces $2 \cdot w \cdot d \cdot r \cdot L$ trainable parameters to the attention module to approximate the updates of the weights, where $w$ denotes the number of attention projection matrices undergoing update. SSF~\cite{ssf} inserts scale and shift ($2 \cdot d$ parameters) after each operation in the model, resulting in $2 \cdot o \cdot d \cdot L$ trainable parameters, where $o$ denotes the number of operations in each layer. ARC~\cite{arc} leverages the sharing strategy across layers to compress the parameters with a compromise in performance, resulting in $2 \cdot (d \cdot r + (d + r) \cdot L)$ trainable parameters. LoRA~\cite{lora}, SSF~\cite{ssf} and ARC~\cite{arc} are capable of re-parameterization, which introduce no extra parameters during inference. Our proposed ALoRE method maintains the desirable property of re-parameterization and reuses the parameter space constructed by Kronecker product and leverages a sharing strategy across layers of the same index of expert, introducing only $n^3$ ($n$ is very small) extra parameters compared to classic low rank decomposition methods~\cite{adaptformer,adapter,lora} while obtaining a significant improvement in performance.

\begin{table*}[htbp]
  \renewcommand\arraystretch{1.3}
  \centering
  \caption{Comparisons of the additional parameter size in both fine-tuning and inference stages with other prevalent PETL methods.}
  \resizebox{1\columnwidth}{!}{
    \begin{tabular}{c|c|c|c|c|c|c|c}
    \toprule[2pt]
    \diagbox{\textbf{Stage}}{\textbf{Method}}
     & Adapter-based~\cite{adaptformer,adapter} & VPT-Shallow~\cite{vpt} & VPT-Deep~\cite{vpt} & LoRA~\cite{lora}  & SSF~\cite{ssf}   & ARC~\cite{arc} & \textbf{ALoRE (ours)} \\
    \midrule[1pt]
    Fine-Tuning &  \(4\cdot d\cdot r \cdot L\) & \(m\cdot d\)    & \({m\cdot d\cdot L}\)     & \({2\cdot w\cdot d\cdot r \cdot L}\)      & \(2\cdot o\cdot d\cdot L\)     & \(2\cdot (d\cdot r +(d + r)\cdot L)\) & \(4 \cdot d \cdot r \cdot L + n^3\)\\
    \midrule[1pt]
    Inference &  \(4\cdot d\cdot r \cdot L\) & \(m\cdot d\)   & \({m\cdot d\cdot L}\)     & 0     & 0     & 0 & 0\\
    \bottomrule[2pt]
    \end{tabular}
    }
  \label{tab:app_size}
\end{table*}

\clearpage

\section{Detailed experimental results}
\label{app:exp}

\subsection{Different Scale Variants}
\label{app:scale}

In Table~\ref{tab:app_large} and~\ref{tab:app_huge}, we present the detailed results on the VTAB-1k benchmark with ViT-Large~\cite{vit} and ViT-Huge~\cite{vit}. 

\begin{table*}[h]
\renewcommand\arraystretch{1.3}
  \centering
  \caption{Performance and parameter efficiency comparisons on the VTAB-1k benchmark with ViT-L/16 pre-trained on ImageNet-21K. ``Group Mean'' denotes the average Top-1 accuracy of the three subgroups. ``All Mean'' denotes the average Top-1 accuracy of 19 downstream tasks.}
  \label{tab:app_large}
  \resizebox{1\columnwidth}{!}{
    \begin{tabular}{c|ccccccc|cccc|cccccccc|ccc}
    \toprule[2pt]
    & \multicolumn{7}{c|}{\textbf{Natural}}      
    & \multicolumn{4}{c|}{\textbf{Specialized}}      
    & \multicolumn{8}{c|}{\textbf{Structured}}         &  & & \\
    \midrule[1pt]
    \diagbox{\textbf{Method}}{\textbf{Dataset}} & \begin{sideways}\textbf{CIFAR-100}\end{sideways} & \begin{sideways}\textbf{Caltech101}\end{sideways} & \begin{sideways}\textbf{DTD}\end{sideways} & \begin{sideways}\textbf{Flowers102}\end{sideways} & \begin{sideways}\textbf{Pets}\end{sideways} & \begin{sideways}\textbf{SVNH}\end{sideways} & \begin{sideways}\textbf{Sun397}\end{sideways} & \begin{sideways}\textbf{Camelyon}\end{sideways} & \begin{sideways}\textbf{EuroSAT}\end{sideways} & \begin{sideways}\textbf{Resisc45}\end{sideways} & \begin{sideways}\textbf{Retinopathy}\end{sideways} &  \begin{sideways}\textbf{Clevr-Count}\end{sideways} & \begin{sideways}\textbf{Clevr-Dist}\end{sideways} & \begin{sideways}\textbf{DMLab}\end{sideways} & \begin{sideways}\textbf{KITTI-Dist}\end{sideways} & \begin{sideways}\textbf{dSpr-Loc}\end{sideways} & \begin{sideways}\textbf{dSpr-Ori}\end{sideways} & \begin{sideways}\textbf{sNORB-Azim}\end{sideways} & \begin{sideways}\textbf{sNORB-Ele}\end{sideways} &  \begin{sideways}\textbf{All Mean}\end{sideways} &  \begin{sideways}\textbf{Group Mean}\end{sideways} & \begin{sideways}\textbf{Params(M)}\end{sideways} \\
    \midrule[1pt]
    \multicolumn{23}{c}{\textit{Traditional methods}} \\
    Full fine-tuning~\cite{vpt} & 68.6 & 84.3 & 58.6 & 96.3 & 86.5 & 87.5 & 41.4 & 82.6 & 95.9 & 82.4 & 74.2 & 55.4 & 55.0 & 42.2 & 74.2 & 56.8 & 43.0 & 28.5 & 29.7 & 65.43 & 68.87 & 303.40 \\
    Linear probing~\cite{vpt} & 72.2 & 86.4 & 63.6 & 97.4 & 85.8 & 38.1 & 52.5 & 76.9 & 87.3 & 66.6 & 45.4 & 28.2 & 28.0 & 34.7 & 54.0 & 10.6 & 14.2 & 14.6 & 21.9 & 51.49 & 55.23 & 0.05 \\
    \midrule[1pt]
    \multicolumn{23}{c}{\textit{Parameter-efficient transfer learning methods}} \\
    Adapter~\cite{adapter} & 75.3 & 84.2 & 54.5 & 97.4 & 84.3 & 31.3 & 52.9 & 75.8 & 85.1 & 63.4 & 69.5 & 35.4 & 34.1 & 30.8 & 47.1 & 30.4 & 23.4 & 10.8 & 19.8 & 52.92 & 56.99 & 2.38 \\
    VPT-Shallow~\cite{vpt} & 80.6 & 88.2 & 67.1 & 98.0 & 85.9 & 78.4 & 53.0 & 79.7 & 93.5 & 73.4 & 73.1 & 41.5 & 52.5 & 32.3 & 64.2 & 48.3 & 35.3 & 21.6 & 28.8 & 62.92 & 66.41 & 0.15 \\
    VPT-Deep~\cite{vpt} & 84.1 & 88.9 & 70.8 & 98.8 & 90.0 & 89.0 & 55.9 & 82.5 & 96.6 & 82.6 & 73.9 & 63.7 & 60.7 & 46.1 & 75.7 & 83.7 & 47.4 & 18.9 & 36.9 & 70.85 & 73.51 & 0.49 \\
    \midrule[1pt]
    \multicolumn{23}{c}{\textit{Structural re-parameterization methods}} \\
    BitFit~\cite{bitfit} & 71.0 & 82.4 & 51.3 & 96.3 & 83.2 & 59.5 & 49.9 & 72.9 & 87.9 & 63.1 & 71.3 & 51.2 & 50.7 & 33.5 & 54.8 & 65.9 & 37.3 & 13.7 & 22.2 & 58.85 & 61.83 & 0.32 \\
    LoRA~\cite{lora} & 75.8 & \uline{89.8} & 73.6 & 99.1 & 90.8 & 83.2 & 57.5 & 86.0 & 95.0 & 83.4 & 75.5 & 78.1 & 60.5 & 46.7 & \textbf{81.6} & 76.7 & 51.3 & 28.0 & 35.4 & 72.00 & 74.55 & 0.74 \\
    ARC~\cite{arc} & 76.2 & 89.6 & 73.4 & 99.1 & 90.3 & \uline{90.9} & 56.5 & 85.0 & 95.7 & 85.9 & \uline{75.8} & 78.6 & 62.1 & 46.7 & 76.7 & 75.9 & \textbf{53.0} & \uline{30.2} & 35.2 & 72.46 & 75.06 & 0.18 \\
    RepAdapter~\cite{rep} & \uline{79.8} & \textbf{93.1} & \uline{75.0} & \uline{99.4} & \uline{92.9} & 87.9 & \uline{59.7} & \uline{87.1} & \uline{95.9} & \uline{86.8} & 75.4 & \uline{83.2} & \uline{63.2} & \uline{51.2} & 78.6 & \uline{82.0} & 52.0 & 29.9 & \uline{41.1} & \uline{74.43} & \uline{76.81} & 0.79 \\
    \rowcolor[rgb]{ .906,  .902,  .902} ALoRE & \textbf{80.6} & \textbf{93.1} & \textbf{75.7} & \textbf{99.6} & \textbf{93.2} & \textbf{92.7} & \textbf{60.1} & \textbf{88.8} & \textbf{96.4} & \textbf{87.7} & \textbf{77.0} & \textbf{84.2} & \textbf{63.5} & \textbf{53.4} & \uline{79.3} & \textbf{85.4} & \uline{52.3} & \textbf{31.2} & \textbf{41.3} & \textbf{75.55} & \textbf{77.93} & 0.39 \\
    \bottomrule[2pt]
    \end{tabular}
    }
\end{table*}
\vspace{-15pt}
\begin{table*}[h]
\renewcommand\arraystretch{1.3}
  \centering
  \caption{Performance and parameter efficiency comparisons on the VTAB-1k benchmark with ViT-H/14 pre-trained on ImageNet-21K. ``Group Mean'' denotes the average Top-1 accuracy of the three subgroups. ``All Mean'' denotes the average Top-1 accuracy of 19 downstream tasks.}
  \label{tab:app_huge}
  \resizebox{1\columnwidth}{!}{
    \begin{tabular}{c|ccccccc|cccc|cccccccc|ccc}
    \toprule[2pt]
    & \multicolumn{7}{c|}{\textbf{Natural}}      
    & \multicolumn{4}{c|}{\textbf{Specialized}}      
    & \multicolumn{8}{c|}{\textbf{Structured}}         &  & & \\
    \midrule[1pt]
    \diagbox{\textbf{Method}}{\textbf{Dataset}} & \begin{sideways}\textbf{CIFAR-100}\end{sideways} & \begin{sideways}\textbf{Caltech101}\end{sideways} & \begin{sideways}\textbf{DTD}\end{sideways} & \begin{sideways}\textbf{Flowers102}\end{sideways} & \begin{sideways}\textbf{Pets}\end{sideways} & \begin{sideways}\textbf{SVNH}\end{sideways} & \begin{sideways}\textbf{Sun397}\end{sideways} & \begin{sideways}\textbf{Camelyon}\end{sideways} & \begin{sideways}\textbf{EuroSAT}\end{sideways} & \begin{sideways}\textbf{Resisc45}\end{sideways} & \begin{sideways}\textbf{Retinopathy}\end{sideways} &  \begin{sideways}\textbf{Clevr-Count}\end{sideways} & \begin{sideways}\textbf{Clevr-Dist}\end{sideways} & \begin{sideways}\textbf{DMLab}\end{sideways} & \begin{sideways}\textbf{KITTI-Dist}\end{sideways} & \begin{sideways}\textbf{dSpr-Loc}\end{sideways} & \begin{sideways}\textbf{dSpr-Ori}\end{sideways} & \begin{sideways}\textbf{sNORB-Azim}\end{sideways} & \begin{sideways}\textbf{sNORB-Ele}\end{sideways} &  \begin{sideways}\textbf{All Mean}\end{sideways} &  \begin{sideways}\textbf{Group Mean}\end{sideways} & \begin{sideways}\textbf{Params(M)}\end{sideways} \\
    \midrule[1pt]
    \multicolumn{23}{c}{\textit{Traditional methods}} \\
    Full fine-tuning~\cite{vpt} & 58.7 & 86.5 & 55.0 & 96.5 & 79.7 & 87.5 & 32.5 & 83.1 & 95.5 & 81.9 & 73.8 & 47.6 & 53.9 & 37.8 & 69.9 & 53.8 & 48.6 & 30.2 & 25.8 & 63.07 & 66.81 & 630.90 \\
    Linear probing~\cite{vpt} & 64.3 & 83.6 & 65.2 & 96.2 & 83.5 & 39.8 & 43.0 & 78.0 & 90.5 & 73.9 & 73.4 & 25.6 & 24.5 & 34.8 & 59.0 & 9.5 & 15.6 & 17.4 & 22.8 & 52.66 & 57.68 & 0.06 \\
    \midrule[1pt]
    \multicolumn{23}{c}{\textit{Parameter-efficient transfer learning methods}} \\
    Adapter~\cite{adapter} & 69.4 & 84.4 & 62.7 & 97.2 & 84.2 & 33.6 & 45.3 & 77.3 & 86.6 & 70.8 & 71.1 & 28.6 & 27.5 & 29.2 & 55.2 & 10.0 & 15.2 & 11.9 & 18.6 & 51.52 & 56.36 & 5.78 \\
    VPT-Shallow~\cite{vpt} & 70.6 & 84.7 & 64.8 & 96.4 & 85.1 & 75.6 & 46.2 & 79.9 & 93.7 & 77.7 & 73.6 & 40.3 & 60.9 & 34.9 & 63.3 & 61.3 & 38.9 & 19.8 & 24.9 & 62.77 & 66.34 & 0.18 \\
    VPT-Deep~\cite{vpt} & 76.9 & 87.2 & 66.8 & 97.5 & 84.8 & 85.5 & 46.5 & 81.6 & 96.3 & 82.5 & 72.8 & 50.4 & 61.2 & 43.9 & 76.6 & 79.5 & 50.1 & 24.7 & 31.5 & 68.23 & 71.14 & 0.96 \\
    \midrule[1pt]
    \multicolumn{23}{c}{\textit{Structural re-parameterization methods}} \\
    BitFit~\cite{bitfit} & 65.7 & 84.3 & 59.9 & 96.6 & 80.6 & 60.1 & 44.9 & 79.7 & 92.8 & 71.5 & 71.6 & 52.3 & 50.4 & 31.2 & 57.7 & 65.9 & 39.7 & 16.7 & 20.2 & 60.09 & 63.65 & 0.52 \\
    LoRA~\cite{lora} & 63.0 & 89.4 & 68.1 & \uline{98.0} & 87.0 & 85.2 & 48.7 & 82.2 & 94.3 & 83.1 & \uline{74.2} & 68.6 & 65.0 & 44.8 & 76.4 & 70.8 & 48.8 & 30.4 & 38.3 & 69.28 & 71.96 & 1.21 \\
    ARC~\cite{arc} & 65.5 & 89.1 & \uline{69.9} & \uline{98.0} & 87.5 & \textbf{89.1} & 48.8 & 83.4 & 94.5 & \textbf{84.5} & \textbf{74.4} & 73.2 & \uline{66.6} & \uline{45.6} & 76.2 & 78.3 & \uline{51.2} & \uline{32.1} & 37.6 & \uline{70.82} & \uline{73.36} & 0.17 \\
    RepAdapter~\cite{rep} & \uline{66.0} & \textbf{93.1} & 66.1 & \uline{98.0} & \uline{88.4} & 84.2 & \uline{49.8} & \uline{83.7} & \uline{94.6} & 80.9 & 73.8 & \uline{74.0} & \textbf{67.2} & 43.8 & \textbf{80.7} & \uline{78.9} & 51.1 & 26.3 & \uline{38.5} & 70.48 & 72.91 & 1.33 \\
    \rowcolor[rgb]{ .906,  .902,  .902} ALoRE & \textbf{69.7} & \uline{92.4} & \textbf{70.9} & \textbf{98.6} & \textbf{89.1} & \uline{87.2} & \textbf{51.8} & \textbf{85.2} & \textbf{94.8} & \uline{84.1} & 74.0 & \textbf{74.7} & 64.8 & \textbf{47.9} & \uline{79.5} & \textbf{81.7} & \textbf{52.2} & \textbf{32.4} & \textbf{38.7} & \textbf{72.11} & \textbf{74.51} & 0.66 \\
    \bottomrule[2pt]
    \end{tabular}
    }
\end{table*}

\clearpage

\subsection{Different Backbone Variants}
\label{app:backbone}

In Table~\ref{tab:app_swin},~\ref{tab:app_conv},~\ref{tab:app_mlp}, we present the detailed results on the VTAB-1k with Swin-Base, ConvNext-Base, and AS-MLP-Base.

\vspace{-5pt}
\begin{table}[htbp]
\renewcommand\arraystretch{1.3}
  \centering
  \caption{Performance and parameter efficiency comparisons on the VTAB-1k with Swin-Base~\cite{swin}.}
  \vspace{-5pt}
  \label{tab:app_swin}
  \resizebox{0.98\columnwidth}{!}{
    \begin{tabular}{c|ccccccc|cccc|cccccccc|ccc}
    \toprule[2pt]
    & \multicolumn{7}{c|}{\textbf{Natural}}      
    & \multicolumn{4}{c|}{\textbf{Specialized}}      
    & \multicolumn{8}{c|}{\textbf{Structured}}         &  & & \\
    \midrule[1pt]
    \diagbox{\textbf{Method}}{\textbf{Dataset}} & \begin{sideways}\textbf{CIFAR-100}\end{sideways} & \begin{sideways}\textbf{Caltech101}\end{sideways} & \begin{sideways}\textbf{DTD}\end{sideways} & \begin{sideways}\textbf{Flowers102}\end{sideways} & \begin{sideways}\textbf{Pets}\end{sideways} & \begin{sideways}\textbf{SVNH}\end{sideways} & \begin{sideways}\textbf{Sun397}\end{sideways} & \begin{sideways}\textbf{Camelyon}\end{sideways} & \begin{sideways}\textbf{EuroSAT}\end{sideways} & \begin{sideways}\textbf{Resisc45}\end{sideways} & \begin{sideways}\textbf{Retinopathy}\end{sideways} &  \begin{sideways}\textbf{Clevr-Count}\end{sideways} & \begin{sideways}\textbf{Clevr-Dist}\end{sideways} & \begin{sideways}\textbf{DMLab}\end{sideways} & \begin{sideways}\textbf{KITTI-Dist}\end{sideways} & \begin{sideways}\textbf{dSpr-Loc}\end{sideways} & \begin{sideways}\textbf{dSpr-Ori}\end{sideways} & \begin{sideways}\textbf{sNORB-Azim}\end{sideways} & \begin{sideways}\textbf{sNORB-Ele}\end{sideways} &  \begin{sideways}\textbf{All Mean}\end{sideways} &  \begin{sideways}\textbf{Group Mean}\end{sideways} & \begin{sideways}\textbf{Params(M)}\end{sideways} \\
    \midrule[1pt]
    \multicolumn{23}{c}{\textit{Traditional methods}} \\
    Full fine-tuning~\cite{vpt} & 72.2 & 88.0 & 71.4 & 98.3 & 89.5 & 89.4 & 45.1 & 86.6 & 96.9 & 87.7 & 73.6 & 75.7 & 59.8 & 54.6 & 78.6 & 79.4 & 53.6 & 34.6 & 40.9 & 72.42 & 74.99 & 86.90 \\
    Linear probing~\cite{vpt} & 61.4 & 90.2 & 74.8 & 95.5 & 90.2 & 46.9 & 55.8 & 81.5 & 90.1 & 82.1 & 69.4 & 39.1 & 35.9 & 40.1 & 65.0 & 20.3 & 26.0 & 14.3 & 27.6 & 58.22 & 62.62 & 0.05 \\
    MLP-4~\cite{vpt} & 54.9 & 87.4 & 71.4 & 99.5 & 89.1 & 39.7 & 52.5 & 80.5 & 90.9 & 76.8 & 74.4 & 60.9 & 38.8 & 40.2 & 66.5 & 9.4 & 21.1 & 14.5 & 28.8 & 57.75 & 62.11 & 4.04 \\
    Partial~\cite{vpt} & 60.3 & 88.9 & 72.6 & 98.7 & 89.3 & 50.5 & 51.5 & 82.8 & 91.7 & 80.1 & 72.3 & 34.3 & 35.5 & 43.2 & 77.1 & 15.8 & 26.2 & 19.1 & 28.4 & 58.86 & 63.26 & 12.65 \\
    \midrule[1pt]
    \multicolumn{23}{c}{\textit{Parameter-efficient transfer learning methods}} \\
    VPT-Shallow~\cite{vpt} & 78.0 & 91.3 & 77.2 & 99.4 & 90.4 & 68.4 & 54.3 & 80.1 & 93.9 & 83.0 & 72.7 & 40.8 & 43.9 & 34.1 & 63.2 & 28.4 & 44.5 & 21.5 & 26.3 & 62.71 & 66.71 & 0.05 \\
    VPT-Deep~\cite{vpt} & 79.6 & 90.8 & 78.0 & 99.5 & 91.4 & 46.5 & 51.7 & 84.9 & 96.2 & 85.0 & 72.0 & 67.6 & 59.4 & 50.1 & 74.1 & 74.4 & 50.6 & 25.7 & 25.7 & 68.59 & 71.59 & 0.22 \\
    \midrule[1pt]
    \multicolumn{23}{c}{\textit{Structural re-parameterization methods}} \\
    BitFit~\cite{bitfit} & 73.1 & 86.8 & 65.7 & 97.7 & 87.5 & 56.4 & 52.3 & 80.4 & 91.6 & 76.1 & 72.5 & 47.3 & 48.5 & 34.7 & 66.3 & 57.6 & 36.2 & 17.2 & 31.6 & 62.08 & 65.60 & 0.25 \\
    ARC~\cite{arc} & 67.2 & 89.7 & 74.7 & 99.5 & 89.7 & \uline{88.5} & \uline{52.7} & \uline{88.1} & 95.9 & \uline{85.7} & \uline{77.2} & 76.5 & \uline{58.5} & 52.1 & 82.8 & \textbf{89.4} & \textbf{56.4} & 27.5 & \uline{35.1} & 73.01 & 75.60 & 0.16 \\
    RepAdapter~\cite{rep} & \textbf{73.9} & \uline{93.3} & \uline{76.7} & \uline{99.6} & \uline{92.9} & 87.9 & \textbf{55.6} & 87.4 & \uline{96.2} & \textbf{88.3} & 76.7 & \textbf{85.8} & 58.0 & \uline{53.4} & \uline{83.1} & \uline{89.1} & 54.5 & \uline{28.9} & \textbf{36.6} & \uline{74.61} & \uline{77.04} & 0.39 \\
    \rowcolor[rgb]{ .906,  .902,  .902} ALoRE & \uline{73.5} & \textbf{93.4} & \textbf{78.0} & \textbf{99.8} & \textbf{93.0} & \textbf{91.9} & \textbf{55.6} & \textbf{92.1} & \textbf{96.5} & \textbf{88.3} & \textbf{79.8} & \uline{82.1} & \textbf{58.9} & \textbf{58.1} & \textbf{85.8} & 88.9 & \uline{54.6} & \textbf{29.1} & 34.6 & \textbf{75.47} & \textbf{78.09} & 0.19 \\
    \bottomrule[2pt]
    \end{tabular}
    }
\end{table}
\vspace{-20pt}
\begin{table}[htbp]
\renewcommand\arraystretch{1.3}
  \centering
  \caption{Performance and parameter efficiency comparisons on the VTAB-1k benchmark with ConvNext-Base~\cite{convnet}.}
  \vspace{-5pt}
  \label{tab:app_conv}
  \resizebox{0.98\columnwidth}{!}{
    \begin{tabular}{c|ccccccc|cccc|cccccccc|ccc}
    \toprule[2pt]
    & \multicolumn{7}{c|}{\textbf{Natural}}      
    & \multicolumn{4}{c|}{\textbf{Specialized}}      
    & \multicolumn{8}{c|}{\textbf{Structured}}         &  & & \\
    \midrule[1pt]
    \diagbox{\textbf{Method}}{\textbf{Dataset}} & \begin{sideways}\textbf{CIFAR-100}\end{sideways} & \begin{sideways}\textbf{Caltech101}\end{sideways} & \begin{sideways}\textbf{DTD}\end{sideways} & \begin{sideways}\textbf{Flowers102}\end{sideways} & \begin{sideways}\textbf{Pets}\end{sideways} & \begin{sideways}\textbf{SVNH}\end{sideways} & \begin{sideways}\textbf{Sun397}\end{sideways} & \begin{sideways}\textbf{Camelyon}\end{sideways} & \begin{sideways}\textbf{EuroSAT}\end{sideways} & \begin{sideways}\textbf{Resisc45}\end{sideways} & \begin{sideways}\textbf{Retinopathy}\end{sideways} &  \begin{sideways}\textbf{Clevr-Count}\end{sideways} & \begin{sideways}\textbf{Clevr-Dist}\end{sideways} & \begin{sideways}\textbf{DMLab}\end{sideways} & \begin{sideways}\textbf{KITTI-Dist}\end{sideways} & \begin{sideways}\textbf{dSpr-Loc}\end{sideways} & \begin{sideways}\textbf{dSpr-Ori}\end{sideways} & \begin{sideways}\textbf{sNORB-Azim}\end{sideways} & \begin{sideways}\textbf{sNORB-Ele}\end{sideways} &  \begin{sideways}\textbf{All Mean}\end{sideways} &  \begin{sideways}\textbf{Group Mean}\end{sideways} & \begin{sideways}\textbf{Params(M)}\end{sideways} \\
    \midrule[1pt]
    Full fine-tuning & 55.3 & 95.0 & 66.8 & 99.0 & 84.2 & \uline{92.6} & 49.8 & \uline{86.6} & 92.3 & 82.3 & 75.5 & 63.9 & 57.7 & 53.3 & 79.6 & \uline{95.5} & 57.8 & \textbf{37.9} & 26.8 & 71.16 & 73.59 & 87.56 \\
    Linear probing & 65.2 & 92.5 & 75.9 & \uline{99.6} & 92.6 & 51.5 & \textbf{56.4} & 83.3 & 92.1 & 80.7 & 74.9 & 44.7 & 38.5 & 38.5 & 64.4 & 23.8 & 32.6 & 17.9 & 26.3 & 60.61 & 64.95 & 0.04 \\
    Adapter~\cite{adapter} & 70.0 & 91.4 & 75.1 & 93.1 & 82.3 & 89.6 & 50.0 & 79.0 & 85.7 & 78.5 & 73.6 & 72.0 & \uline{61.4} & 42.0 & 72.0 & 87.7 & 57.9 & 35.6 & 30.1 & 69.84 & 71.77 & 0.67 \\
    Adaptformer~\cite{adaptformer} & 73.4 & 95.1 & 77.0 & \textbf{99.7} & 92.6 & 89.3 & 54.7 & 84.2 & 94.9 & 87.4 & 75.3 & 79.5 & 58.1 & 51.1 & 79.8 & 85.5 & 60.9 & 29.3 & 33.2 & 73.74 & 76.09 & 0.34 \\
    SSF~\cite{ssf} & \uline{75.4} & \textbf{96.0} & \uline{77.5} & \textbf{99.7} & \uline{93.1} & 91.0 & 53.8 & 86.3 & \uline{95.7} & \uline{88.7} & \uline{77.2} & \textbf{91.9} & 60.0 & \uline{55.1} & \uline{83.5} & 89.2 & \uline{61.6} & \uline{36.8} & \uline{38.7} & \uline{76.38} & \uline{78.46} & 0.27 \\
    \rowcolor[rgb]{ .906,  .902,  .902}ALoRE & \textbf{76.7} & \uline{95.6} & \textbf{78.8} & \textbf{99.7} & \textbf{93.5} & \textbf{95.0} & \uline{56.2} & \textbf{89.6} & \textbf{96.2} & \textbf{89.8} & \textbf{77.8} & \uline{89.7} & \textbf{61.9} & \textbf{56.9} & \textbf{86.2} & \textbf{96.3} & \textbf{62.0} & 35.8 & \textbf{39.8} & \textbf{77.75} & \textbf{79.81} & 0.29 \\
    \bottomrule[2pt]
    \end{tabular}
    }
\end{table}
\vspace{-20pt}
\begin{table}[htbp]
\renewcommand\arraystretch{1.3}
  \centering
  \caption{Performance and parameter efficiency comparisons on the VTAB-1k benchmark with AS-MLP-Base~\cite{asmlp}.}
  \label{tab:app_mlp}
  \vspace{-5pt}
  \resizebox{0.98\columnwidth}{!}{
    \begin{tabular}{c|ccccccc|cccc|cccccccc|ccc}
    \toprule[2pt]
    & \multicolumn{7}{c|}{\textbf{Natural}}      
    & \multicolumn{4}{c|}{\textbf{Specialized}}      
    & \multicolumn{8}{c|}{\textbf{Structured}}         &  & & \\
    \midrule[1pt]
    \diagbox{\textbf{Method}}{\textbf{Dataset}} & \begin{sideways}\textbf{CIFAR-100}\end{sideways} & \begin{sideways}\textbf{Caltech101}\end{sideways} & \begin{sideways}\textbf{DTD}\end{sideways} & \begin{sideways}\textbf{Flowers102}\end{sideways} & \begin{sideways}\textbf{Pets}\end{sideways} & \begin{sideways}\textbf{SVNH}\end{sideways} & \begin{sideways}\textbf{Sun397}\end{sideways} & \begin{sideways}\textbf{Camelyon}\end{sideways} & \begin{sideways}\textbf{EuroSAT}\end{sideways} & \begin{sideways}\textbf{Resisc45}\end{sideways} & \begin{sideways}\textbf{Retinopathy}\end{sideways} &  \begin{sideways}\textbf{Clevr-Count}\end{sideways} & \begin{sideways}\textbf{Clevr-Dist}\end{sideways} & \begin{sideways}\textbf{DMLab}\end{sideways} & \begin{sideways}\textbf{KITTI-Dist}\end{sideways} & \begin{sideways}\textbf{dSpr-Loc}\end{sideways} & \begin{sideways}\textbf{dSpr-Ori}\end{sideways} & \begin{sideways}\textbf{sNORB-Azim}\end{sideways} & \begin{sideways}\textbf{sNORB-Ele}\end{sideways} &  \begin{sideways}\textbf{All Mean}\end{sideways} &  \begin{sideways}\textbf{Group Mean}\end{sideways} & \begin{sideways}\textbf{Params(M)}\end{sideways} \\
    \midrule[1pt]
    Full fine-tuning & 47.4 & 88.9 & 68.1 & 79.8 & 91.4 & 50.2 & 40.7 & 82.8 & 91.1 & 74.5 & 74.1 & 37.3 & 37.9 & 34.7 & 63.9 & 17.7 & 23.7 & 13.5 & 24.7 & 54.86 & 59.65 & 86.73 \\
    Linear probing & 9.7 & 27.2 & 14.0 & 35.1 & 9.6 & 19.6 & 4.7 & 77.2 & 76.3 & 37.2 & 73.6 & 50.2 & 61.6 & 34.9 & 39.5 & 88.1 & 17.2 & 9.0 & 21.1 & 37.14 & 41.13 & 0.05 \\
    Adapter~\cite{adapter} & 44.9 & 88.3 & 57.7 & 86.9 & 83.1 & \uline{92.0} & 33.7 & 83.8 & 93.6 & \textbf{82.2} & \uline{75.3} & 61.1 & 62.9 & \uline{53.2} & 79.2 & \uline{96.1} & 54.9 & \textbf{37.0} & 29.9 & 68.20 & 70.84 & 0.45 \\
    Adaptformer~\cite{adaptformer} & 55.9 & \uline{90.9} & 68.6 & 86.3 & 91.7 & 81.5 & \textbf{42.2} & 83.0 & 93.5 & 77.9 & 75.0 & 86.7 & \uline{63.7} & 44.5 & 80.3 & 70.5 & 58.1 & 26.5 & 36.7 & 69.14 & 71.54 & 0.22 \\
    SSF~\cite{ssf} & \uline{56.4} & \textbf{91.2} & \uline{70.4} & \uline{88.0} & \uline{92.1} & 84.9 & \uline{41.9} & \uline{84.0} & \uline{94.9} & 81.6 & 75.1 & \uline{90.2} & \textbf{64.2} & 49.9 & \uline{82.7} & 87.5 & \textbf{59.3} & 31.4 & \uline{41.4} & \uline{71.96} & \uline{74.08} & 0.26 \\
    \rowcolor[rgb]{ .906,  .902,  .902}ALoRE & \textbf{58.3} & 90.7 & \textbf{71.6} & \textbf{88.4} & \textbf{92.6} & \textbf{94.6} & 41.6 & \textbf{87.9} & \textbf{95.4} & \uline{82.0} & \textbf{77.9} & \textbf{91.4} & \textbf{64.2} & \textbf{54.2} & \textbf{84.7} & \textbf{96.5} & \uline{58.8} & \uline{33.0} & \textbf{42.5} & \textbf{74.01} & \textbf{76.09} & 0.19 \\
    \bottomrule[2pt]
    \end{tabular}
    }
\end{table}
\clearpage

\subsection{Different Pre-training Strategies}
\label{app:pretrain}
In Table ~\ref{tab:app_mae} and ~\ref{tab:app_moco}, we present the detailed results on the VTAB-1k benchmark with ViT-B/16~\cite{vit} pre-trained on MAE~\cite{mae} and MoCo-v3~\cite{mocov3} respectively.

\begin{table}[htbp]
\renewcommand\arraystretch{1.3}
\vspace{-5pt}
  \centering
  \caption{Performance and parameter efficiency comparisons on the VTAB-1k benchmark with ViT-B/16 pre-trained on MAE~\cite{mae}. ``Group Mean'' denotes the average Top-1 accuracy of the three subgroups. ``All Mean'' denotes the average Top-1 accuracy of 19 downstream tasks.}
  \label{tab:app_mae}
  \resizebox{1\columnwidth}{!}{
    \begin{tabular}{c|ccccccc|cccc|cccccccc|ccc}
    \toprule[2pt]
    & \multicolumn{7}{c|}{\textbf{Natural}}      
    & \multicolumn{4}{c|}{\textbf{Specialized}}      
    & \multicolumn{8}{c|}{\textbf{Structured}}         &  & & \\
    \midrule[1pt]
    \diagbox{\textbf{Method}}{\textbf{Dataset}} & \begin{sideways}\textbf{CIFAR-100}\end{sideways} & \begin{sideways}\textbf{Caltech101}\end{sideways} & \begin{sideways}\textbf{DTD}\end{sideways} & \begin{sideways}\textbf{Flowers102}\end{sideways} & \begin{sideways}\textbf{Pets}\end{sideways} & \begin{sideways}\textbf{SVNH}\end{sideways} & \begin{sideways}\textbf{Sun397}\end{sideways} & \begin{sideways}\textbf{Camelyon}\end{sideways} & \begin{sideways}\textbf{EuroSAT}\end{sideways} & \begin{sideways}\textbf{Resisc45}\end{sideways} & \begin{sideways}\textbf{Retinopathy}\end{sideways} &  \begin{sideways}\textbf{Clevr-Count}\end{sideways} & \begin{sideways}\textbf{Clevr-Dist}\end{sideways} & \begin{sideways}\textbf{DMLab}\end{sideways} & \begin{sideways}\textbf{KITTI-Dist}\end{sideways} & \begin{sideways}\textbf{dSpr-Loc}\end{sideways} & \begin{sideways}\textbf{dSpr-Ori}\end{sideways} & \begin{sideways}\textbf{sNORB-Azim}\end{sideways} & \begin{sideways}\textbf{sNORB-Ele}\end{sideways} &  \begin{sideways}\textbf{All Mean}\end{sideways} &  \begin{sideways}\textbf{Group Mean}\end{sideways} & \begin{sideways}\textbf{Params(M)}\end{sideways} \\
    \midrule[1pt]
    \multicolumn{23}{c}{\textit{Traditional methods}} \\
    Full fine-tuning~\cite{arc} & 24.6 & 84.2 & 56.9 & 72.7 & 74.4 & 86.6 & 15.8 & 81.8 & 94.0 & 72.3 & 70.6 & 67.0 & 59.8 & 45.2 & 75.3 & 72.5 & 47.5 & 30.2 & 33.0 & 61.28 & 64.27 & 85.84 \\
    Linear probing~\cite{arc} & 8.7 & 41.5 & 20.6 & 19.2 & 11.3 & 22.3 & 8.6 & 76.5 & 68.6 & 16.6 & 53.2 & 33.6 & 32.5 & 23.0 & 51.1 & 13.0 & 9.9 & 8.5 & 17.9 & 28.24 & 32.1 & 0.04 \\
    \midrule[1pt]
    \multicolumn{23}{c}{\textit{Parameter-efficient transfer learning methods}} \\
    Adapter~\cite{adapter} & 35.1 & 85.0 & 56.5 & 66.6 & 71.3 & 45.0 & 24.8 & 76.9 & 87.1 & 63.5 & 73.3 & 43.8 & 49.5 & 31.2 & 61.7 & 59.3 & 23.3 & 13.6 & 29.6 & 52.48 & 56.37 & 0.76 \\ 
    VPT-Shallow~\cite{vpt} & 21.9 & 76.2 & 54.7 & 58.0 & 41.3 & 16.1 & 15.1 & 74.0 & 69.5 & 58.9 & 72.7 & 40.3 & 44.7 & 27.9 & 60.5 & 11.8 & 11.0 & 12.4 & 16.3 & 41.23 & 45.79 & 0.04 \\
    VPT-Deep~\cite{vpt} & 8.2 & 55.2 & 58.0 & 39.3 & 45.2 & 19.4 & 21.9 & 77.9 & 91.0 & 45.4 & 73.6 & 39.0 & 40.9 & 30.6 & 53.9 & 21.0 & 12.1 & 11.0 & 14.9 & 39.92 & 45.07 & 0.06 \\
    \midrule[1pt]
    \multicolumn{23}{c}{\textit{Structural re-parameterization methods}} \\
    BitFit~\cite{bitfit} & 22.4 & 82.6 & 49.7 & 66.2 & 67.7 & 69.0 & 24.3 & 78.7 & 91.4 & 60.0 & 72.6 & 65.9 & 51.0 & 35.0 & 69.1 & 70.8 & 37.6 & 21.5 & 30.7 & 56.12 & 59.31 & 0.14 \\
    LoRA~\cite{lora} & 31.8 & 88.4 & 59.9 & 81.7 & \uline{85.3} & 90.3 & 23.7 & 84.2 & 92.5 & 76.2 & 75.4 & \textbf{85.9} & \uline{64.1} & 49.4 & 82.8 & 83.9 & 51.8 & 34.6 & \uline{41.3} & 67.54 & 69.89 & 0.30 \\
    ARC\({\rm{}_{attn}}\)~\cite{arc} & 34.8 & 89.3 & 62.0 & \textbf{85.9} & 84.4 & 91.1 & 24.8 & 85.8 & 93.5 & 81.3 & 75.6 & 84.0 & 63.5 & 51.2 & 83.0 & \uline{89.1} & 54.0 & 34.2 & \textbf{43.0} & 68.97 & 71.42 & 0.09 \\
    ARC~\cite{arc} & 31.3 & 89.3 & 61.2 & \textbf{85.9} & 83.1 & 91.6 & 24.4 & 86.0 & 94.0 & 80.4 & 74.8 & \uline{85.8} & \textbf{64.6} & 50.5 & 82.8 & 82.8 & 53.5 & \uline{36.3} & 39.7 & 68.32 & 70.83 & 0.13 \\
    RepAdapter~\cite{rep} & \uline{36.6} & 89.8 & 64.0 & 83.6 & 84.9 & 90.6 & 26.7 & 87.3 & 93.0 & 80.3 & 78.6 & 79.3 & 63.5 & 54.1 & 84.8 & 80.9 & 51.8 & 25.2 & 34.0 & 67.83 & 70.66 & 0.22 \\
    ALoRE\({\rm{}_{attn}}\) & \textbf{36.9} & \textbf{90.8} & \uline{64.7} & 85.2 & 85.2 & \uline{93.3} & \textbf{27.4} & \uline{88.8} & \uline{94.5} & \uline{81.7} & \uline{80.1} & 81.5 & 62.1 & \textbf{57.0} & \textbf{86.8} & 86.0 & \textbf{55.0} & 32.5 & 35.3 & \uline{69.71} & \uline{72.44} & 0.07 \\
    \rowcolor[rgb]{ .906,  .902,  .902} ALoRE & \textbf{36.9} & \uline{90.3} & \textbf{65.1} & \uline{85.7} & \textbf{86.0} & \textbf{95.2} & \uline{27.1} & \textbf{92.2} & \textbf{95.1} & \textbf{82.3} & \textbf{80.6} & 78.3 & 61.5 & \uline{56.5} & \uline{85.0} & \textbf{93.8} & \uline{54.9} & \textbf{38.6} & 27.4 & \textbf{70.13} & \textbf{73.01} & 0.15 \\
    \bottomrule[2pt]
    \end{tabular}
    }
\end{table}
\vspace{-20pt}
\begin{table}[htbp]
\renewcommand\arraystretch{1.3}
  \centering
  \caption{Performance and parameter efficiency comparisons on the VTAB-1k with ViT-B/16 pre-trained on MoCo-v3~\cite{mocov3}. ``Group Mean'' denotes the average Top-1 accuracy of the three subgroups. ``All Mean'' denotes the average Top-1 accuracy of 19 tasks.}
  \label{tab:app_moco}
  \resizebox{1\columnwidth}{!}{
    \begin{tabular}{c|ccccccc|cccc|cccccccc|ccc}
    \toprule[2pt]
    & \multicolumn{7}{c|}{\textbf{Natural}}      
    & \multicolumn{4}{c|}{\textbf{Specialized}}      
    & \multicolumn{8}{c|}{\textbf{Structured}}         &  & & \\
    \midrule[1pt]
    \diagbox{\textbf{Method}}{\textbf{Dataset}} & \begin{sideways}\textbf{CIFAR-100}\end{sideways} & \begin{sideways}\textbf{Caltech101}\end{sideways} & \begin{sideways}\textbf{DTD}\end{sideways} & \begin{sideways}\textbf{Flowers102}\end{sideways} & \begin{sideways}\textbf{Pets}\end{sideways} & \begin{sideways}\textbf{SVNH}\end{sideways} & \begin{sideways}\textbf{Sun397}\end{sideways} & \begin{sideways}\textbf{Camelyon}\end{sideways} & \begin{sideways}\textbf{EuroSAT}\end{sideways} & \begin{sideways}\textbf{Resisc45}\end{sideways} & \begin{sideways}\textbf{Retinopathy}\end{sideways} &  \begin{sideways}\textbf{Clevr-Count}\end{sideways} & \begin{sideways}\textbf{Clevr-Dist}\end{sideways} & \begin{sideways}\textbf{DMLab}\end{sideways} & \begin{sideways}\textbf{KITTI-Dist}\end{sideways} & \begin{sideways}\textbf{dSpr-Loc}\end{sideways} & \begin{sideways}\textbf{dSpr-Ori}\end{sideways} & \begin{sideways}\textbf{sNORB-Azim}\end{sideways} & \begin{sideways}\textbf{sNORB-Ele}\end{sideways} &  \begin{sideways}\textbf{All Mean}\end{sideways} &  \begin{sideways}\textbf{Group Mean}\end{sideways} & \begin{sideways}\textbf{Params(M)}\end{sideways} \\
    \midrule[1pt]
    \multicolumn{23}{c}{\textit{Traditional methods}} \\
    Full fine-tuning~\cite{arc} & 57.6 & 91.0 & 64.6 & 91.5 & 79.9 & 89.8 & 29.1 & 85.1 & 96.4 & 83.1 & 74.3 & 55.1 & 56.9 & 44.7 & 77.9 & 63.8 & 49.0 & 31.5 & 36.9 & 66.22 & 69.54 & 85.69 \\
    Linear probing~\cite{arc} & 62.9 & 85.1 & 68.8 & 87.0 & 85.8 & 41.8 & 40.9 & 80.3 & 93.6 & 77.9 & 72.6 & 42.3 & 34.8 & 36.4 & 59.2 & 10.1 & 22.7 & 12.6 & 24.7 & 54.71 & 59.64 & 0.04 \\
    \midrule[1pt]
    \multicolumn{23}{c}{\textit{Parameter-efficient transfer learning methods}} \\
    Adapter~\cite{adapter} & 73.0 & 88.2 & 69.3 & 90.7 & 87.4 & 69.9 & 40.9 & 82.4 & 93.4 & 80.5 & 74.3 & 55.6 & 56.1 & 39.1 & 73.9 & 60.5 & 40.2 & 19.0 & 37.1 & 64.82 & 68.18 & 0.98 \\
    VPT-Shallow~\cite{vpt} & 68.3 & 86.8 & 69.7 & 90.0 & 59.7 & 56.9 & 39.9 & 81.7 & 94.7 & 78.9 & 73.8 & 34.3 & 56.8 & 40.6 & 49.1 & 40.4 & 31.8 & 13.1 & 34.4 & 57.94 & 62.39 & 0.05 \\
    VPT-Deep~\cite{vpt} & 70.1 & 88.3 & 65.9 & 88.4 & 85.6 & 57.8 & 35.7 & 83.1 & 93.9 & 81.2 & 74.0 & 48.5 & 55.8 & 37.2 & 64.6 & 52.3 & 26.5 & 19.4 & 34.8 & 61.22 & 65.23 & 0.05 \\
    \midrule[1pt]
    \multicolumn{23}{c}{\textit{Structural re-parameterization methods}} \\
    BitFit~\cite{bitfit} & \textbf{65.5} & 89.2 & 62.9 & 88.9 & 80.5 & 82.7 & 40.5 & 80.9 & 95.2 & 77.7 & 70.8 & 71.4 & 59.4 & 39.8 & 77.4 & 70.2 & 49.0 & 17.5 & 42.8 & 66.44 & 69.16 & 0.14 \\
    LoRA~\cite{lora} & 58.8 & 90.8 & 66.0 & 91.8 & 88.1 & 87.6 & \uline{40.6} & 86.4 & 95.3 & 83.4 & 75.5 & 83.0 & \uline{64.6} & 51.3 & 81.9 & 83.2 & 47.5 & 32.4 & \uline{47.3} & 71.34 & 73.79 & 0.30 \\
    ARC\({\rm{}_{attn}}\)~\cite{arc} & 59.3 & 90.9 & \uline{67.7} & \textbf{93.6} & \uline{89.2} & 90.5 & 40.3 & 87.1 & 94.8 & \uline{85.4} & 75.5 & \textbf{84.0} & \textbf{64.9} & 51.5 & 83.1 & 88.2 & \textbf{53.4} & 33.0 & 46.2 & 72.56 & 74.89 & 0.09 \\
    ARC~\cite{arc} & \uline{60.0} & \uline{91.3} & \textbf{67.9} & \uline{92.8} & \textbf{89.3} & 91.4 & \textbf{40.9} & 87.5 & 95.6 & \textbf{86.1} & 75.6 & 83.0 & 64.2 & 50.2 & 80.6 & 85.0 & \uline{53.0} & \uline{34.6} & \textbf{47.4} & 72.44 & 74.89 & 0.13 \\
    RepAdapter~\cite{rep} & 57.2 & \textbf{91.9} & 66.1 & 89.7 & 87.8 & 91.6 & 37.1 & 88.9 & 96.0 & 82.4 & \uline{77.2} & 79.3 & 62.5 & 53.8 & 83.1 & 85.7 & 44.6 & 30.8 & 46.2 & 71.16 & 73.80 & 0.22 \\
    ALoRE\({\rm{}_{attn}}\) & 55.3 & \uline{91.3} & 65.6 & 87.3 & 86.7 & \uline{92.9} & 35.1 & \uline{92.7} & \textbf{96.8} & 83.0 & \textbf{80.2} & 82.5 & 63.0 & \uline{58.2} & \textbf{85.4} & \uline{92.4} & 49.4 & \uline{34.6} & 46.9 & \uline{72.60} & \uline{75.23} & 0.07 \\
    \rowcolor[rgb]{ .906,  .902,  .902} ALoRE & 52.1 & 90.8 & 65.5 & 86.9 & 86.0 & \textbf{94.8} & 33.8 & \textbf{93.1} & \uline{96.7} & 84.0 & \textbf{80.2} & \uline{83.4} & 62.1 & \textbf{59.6} & \uline{84.8} & \textbf{96.3} & 51.0 & \textbf{37.6} & 46.2 & \textbf{72.88} & \textbf{75.49} & 0.15 \\
    \bottomrule[2pt]
    \end{tabular}
    }
\end{table}

\clearpage
\subsection{Ablation Study}
\label{app:abl}
In Table ~\ref{tab:app_dim}, ~\ref{tab:app_experts}, ~\ref{tab:app_loc} and ~\ref{tab:app_sel}, we present the detailed ablation results on the VTAB-1k benchmark with ViT-B/16 model and provide a supplementary result on the experts in Table ~\ref{tab:app_rep}.

\begin{table}[htbp]
\renewcommand\arraystretch{1.3}
  \centering
  \caption{Ablation on bottleneck dimensions. ``Group Mean'' denotes the average Top-1 accuracy of the three subgroups. ``All Mean'' denotes the average Top-1 accuracy of 19 downstream tasks.}
  \label{tab:app_dim}
  \resizebox{1\columnwidth}{!}{
    \begin{tabular}{c|ccccccc|cccc|cccccccc|ccc}
    \toprule[2pt]
    & \multicolumn{7}{c|}{\textbf{Natural}}      
    & \multicolumn{4}{c|}{\textbf{Specialized}}      
    & \multicolumn{8}{c|}{\textbf{Structured}}         &  & & \\
    \midrule[1pt]
    \diagbox{\textbf{Rank}}{\textbf{Dataset}} & \begin{sideways}\textbf{CIFAR-100}\end{sideways} & \begin{sideways}\textbf{Caltech101}\end{sideways} & \begin{sideways}\textbf{DTD}\end{sideways} & \begin{sideways}\textbf{Flowers102}\end{sideways} & \begin{sideways}\textbf{Pets}\end{sideways} & \begin{sideways}\textbf{SVNH}\end{sideways} & \begin{sideways}\textbf{Sun397}\end{sideways} & \begin{sideways}\textbf{Camelyon}\end{sideways} & \begin{sideways}\textbf{EuroSAT}\end{sideways} & \begin{sideways}\textbf{Resisc45}\end{sideways} & \begin{sideways}\textbf{Retinopathy}\end{sideways} &  \begin{sideways}\textbf{Clevr-Count}\end{sideways} & \begin{sideways}\textbf{Clevr-Dist}\end{sideways} & \begin{sideways}\textbf{DMLab}\end{sideways} & \begin{sideways}\textbf{KITTI-Dist}\end{sideways} & \begin{sideways}\textbf{dSpr-Loc}\end{sideways} & \begin{sideways}\textbf{dSpr-Ori}\end{sideways} & \begin{sideways}\textbf{sNORB-Azim}\end{sideways} & \begin{sideways}\textbf{sNORB-Ele}\end{sideways} &  \begin{sideways}\textbf{All Mean}\end{sideways} &  \begin{sideways}\textbf{Group Mean}\end{sideways} & \begin{sideways}\textbf{Params(M)}\end{sideways} \\
    \midrule[1pt]
    1 & \textbf{76.4} & 92.0 & \textbf{74.2} & 99.4 & 92.2 & 88.5 & \textbf{58.0} & 86.5 & 95.4 & 86.2 & 76.5 & 80.3 & 62.1 & 49.8 & 80.0 & 82.7 & \uline{54.1} & 28.4 & 41.5 & 73.90 & 76.31 & 0.04 \\
    2 & 76.0 & 91.8 & 73.5 & 99.4 & \textbf{92.4} & 88.9 & \uline{57.9} & 87.7 & 95.7 & 86.6 & \uline{76.6} & 82.2 & 63.8 & 53.0 & 79.2 & 84.2 & 53.9 & 29.9 & 44.3 & 74.58 & 76.94 & 0.07 \\
    \rowcolor[rgb]{ .906,  .902,  .902} 4 & \textbf{76.4} & \uline{93.2} & \uline{74.1} & \uline{99.5} & 92.2 & \uline{91.2} & \uline{57.9} & 88.2 & 96.0 & 87.8 & 76.1 & \uline{82.5} & \textbf{66.3} & 53.3 & \uline{81.6} & \textbf{86.5} & \textbf{54.9} & 32.0 & \uline{45.4} & \textbf{75.54} & \textbf{77.78} & 0.15 \\
    8 & 75.1 & \textbf{93.7} & 73.6 & \uline{99.5} & \uline{92.3} & \textbf{91.6} & 57.4 & \uline{88.3} & \uline{96.2} & \uline{88.2} & \textbf{76.9} & \textbf{83.2} & 62.9 & \textbf{53.6} & \textbf{82.3} & 84.0 & 52.5 & \textbf{35.7} & \uline{45.4} & \uline{75.39} & \uline{77.72} & 0.29 \\
    16 & \uline{76.1} & 92.8 & 73.9 & \textbf{99.6} & \uline{92.3} & 90.6 & 57.5 & \textbf{88.4} & \textbf{96.4} & \textbf{88.6} & 75.8 & \textbf{83.2} & \uline{64.8} & \uline{53.5} & 81.4 & \uline{85.5} & 52.2 & 33.2 & \textbf{46.7} & 74.86 & 77.70 & 0.59 \\
    32 & 73.7 & \uline{93.2} & 72.5 & \uline{99.5} & 92.0 & 87.8 & 56.0 & 87.6 & 95.8 & 87.7 & 75.7 & 82.0 & 63.4 & 51.2 & 80.7 & 81.6 & 53.7 & \uline{34.2} & 44.4 & 74.35 & 76.73 & 1.18 \\
    \bottomrule[2pt]
    \end{tabular}
    }
    \vspace{20pt}
\end{table}
\begin{table}[htbp]
\renewcommand\arraystretch{1.3}
  \centering
  \vspace{-25pt}
  \caption{Ablation on the number of experts. ``Group Mean'' denotes the average Top-1 accuracy of the three subgroups. ``All Mean'' denotes the average Top-1 accuracy of 19 downstream tasks.}
  \label{tab:app_experts}
  \resizebox{1\columnwidth}{!}{
    \begin{tabular}{c|ccccccc|cccc|cccccccc|ccc}
    \toprule[2pt]
    & \multicolumn{7}{c|}{\textbf{Natural}}      
    & \multicolumn{4}{c|}{\textbf{Specialized}}      
    & \multicolumn{8}{c|}{\textbf{Structured}}         &  & & \\
    \midrule[1pt]
    \diagbox{\textbf{Nums}}{\textbf{Dataset}} & \begin{sideways}\textbf{CIFAR-100}\end{sideways} & \begin{sideways}\textbf{Caltech101}\end{sideways} & \begin{sideways}\textbf{DTD}\end{sideways} & \begin{sideways}\textbf{Flowers102}\end{sideways} & \begin{sideways}\textbf{Pets}\end{sideways} & \begin{sideways}\textbf{SVNH}\end{sideways} & \begin{sideways}\textbf{Sun397}\end{sideways} & \begin{sideways}\textbf{Camelyon}\end{sideways} & \begin{sideways}\textbf{EuroSAT}\end{sideways} & \begin{sideways}\textbf{Resisc45}\end{sideways} & \begin{sideways}\textbf{Retinopathy}\end{sideways} &  \begin{sideways}\textbf{Clevr-Count}\end{sideways} & \begin{sideways}\textbf{Clevr-Dist}\end{sideways} & \begin{sideways}\textbf{DMLab}\end{sideways} & \begin{sideways}\textbf{KITTI-Dist}\end{sideways} & \begin{sideways}\textbf{dSpr-Loc}\end{sideways} & \begin{sideways}\textbf{dSpr-Ori}\end{sideways} & \begin{sideways}\textbf{sNORB-Azim}\end{sideways} & \begin{sideways}\textbf{sNORB-Ele}\end{sideways} &  \begin{sideways}\textbf{All Mean}\end{sideways} &  \begin{sideways}\textbf{Group Mean}\end{sideways} & \begin{sideways}\textbf{Params(M)}\end{sideways} \\
    \midrule[1pt]
    1 & 74.2 & 92.9 & 72.8 & \uline{99.5} & 91.3 & 86.0 & 57.0 & 87.0 & 95.3 & 85.9 & 73.7 & 81.3 & 63.8 & 50.8 & 79.8 & 82.0 & 51.2 & 31.9 & \uline{45.8} & 73.81 & 76.10 & 0.15 \\
    2 & 73.7 & \textbf{93.5} & 72.0 & \textbf{99.6} & 91.7 & 88.1 & 56.9 & 87.3 & \textbf{96.2} & 86.6 & 75.6 & \textbf{82.8} & 63.2 & 53.1 & \textbf{82.2} & 83.2 & 52.1 & \textbf{33.4} & \textbf{45.9} & 74.58 & 76.87 & 0.15 \\
    \rowcolor[rgb]{ .906,  .902,  .902} 4 & 76.4 & \uline{93.2} & 74.1 & \uline{99.5} & 92.2 & \uline{91.2} & 57.9 & 88.2 & \uline{96.0} & 87.8 & 76.1 & \uline{82.5} & \textbf{66.3} & 53.3 & \uline{81.6} & \textbf{86.5} & \uline{54.9} & \uline{32.0} & 45.4 & \textbf{75.54} & \textbf{77.78} & 0.15 \\
    8 & 77.5 & 92.0 & 74.4 & \textbf{99.6} & \uline{92.8} & 89.4 & 58.1 & \textbf{89.9} & \uline{96.0} & \textbf{89.0} & \uline{77.8} & 82.2 & \uline{64.6} & 52.8 & 80.9 & 83.3 & 53.9 & \uline{32.0} & 43.0 & \uline{75.22} & \uline{77.72} & 0.15 \\
    16 & \textbf{77.8} & 91.6 & \uline{74.5} & \textbf{99.6} & 92.7 & \textbf{92.2} & \uline{58.4} & 87.7 & 95.8 & 88.3 & 77.7 & 80.1 & 62.1 & \textbf{53.8} & 78.6 & \uline{84.8} & 54.2 & 29.8 & 42.3 & 74.85 & 77.31 & 0.15 \\
    32 & \uline{77.6} & 92.3 & \textbf{74.6} & \textbf{99.6} & \textbf{93.0} & 91.0 & \textbf{59.0} & \uline{88.5} & \textbf{96.2} & \uline{88.4} & \textbf{78.3} & 79.1 & 62.9 & \uline{53.7} & 79.6 & 84.4 & \textbf{55.2} & 29.4 & 40.0 & 74.89 & 77.42 & 0.18 \\
    \bottomrule[2pt]
    \end{tabular}
    }
\end{table}
\clearpage
\begin{table}[htbp]
\renewcommand\arraystretch{1.3}
  \centering
  \caption{Ablation on the ALoRE location.}
  \label{tab:app_loc}
  \resizebox{0.9\columnwidth}{!}{
    \begin{tabular}{c|ccccccc|cccc|cccccccc|ccc}
    \toprule[2pt]
    & \multicolumn{7}{c|}{\textbf{Natural}}      
    & \multicolumn{4}{c|}{\textbf{Specialized}}      
    & \multicolumn{8}{c|}{\textbf{Structured}}         &  & & \\
    \midrule[1pt]
    \diagbox{\textbf{Location}}{\textbf{Dataset}} & \begin{sideways}\textbf{CIFAR-100}\end{sideways} & \begin{sideways}\textbf{Caltech101}\end{sideways} & \begin{sideways}\textbf{DTD}\end{sideways} & \begin{sideways}\textbf{Flowers102}\end{sideways} & \begin{sideways}\textbf{Pets}\end{sideways} & \begin{sideways}\textbf{SVNH}\end{sideways} & \begin{sideways}\textbf{Sun397}\end{sideways} & \begin{sideways}\textbf{Camelyon}\end{sideways} & \begin{sideways}\textbf{EuroSAT}\end{sideways} & \begin{sideways}\textbf{Resisc45}\end{sideways} & \begin{sideways}\textbf{Retinopathy}\end{sideways} &  \begin{sideways}\textbf{Clevr-Count}\end{sideways} & \begin{sideways}\textbf{Clevr-Dist}\end{sideways} & \begin{sideways}\textbf{DMLab}\end{sideways} & \begin{sideways}\textbf{KITTI-Dist}\end{sideways} & \begin{sideways}\textbf{dSpr-Loc}\end{sideways} & \begin{sideways}\textbf{dSpr-Ori}\end{sideways} & \begin{sideways}\textbf{sNORB-Azim}\end{sideways} & \begin{sideways}\textbf{sNORB-Ele}\end{sideways} &  \begin{sideways}\textbf{All Mean}\end{sideways} &  \begin{sideways}\textbf{Group Mean}\end{sideways} & \begin{sideways}\textbf{Params(M)}\end{sideways} \\
    \midrule[1pt]
    Before MHA & 75.2 & 92.6 & 73.6 & \uline{99.4} & \uline{92.3} & \uline{90.3} & 57.5 & 87.0 & 95.5 & 87.0 & 76.4 & 81.6 & \uline{65.9} & 52.6 & 79.5 & 84.3 & \textbf{55.1} & 31.9 & \textbf{45.7} & 74.92 & 77.18 & 0.07 \\
    After MHA & 75.5 & 92.5 & \uline{73.7} & \textbf{99.5} & \textbf{92.5} & 87.3 & 57.8 & 87.1 & \uline{96.0} & 86.6 & \uline{76.6} & \uline{82.7} & 62.7 & 52.1 & 79.6 & 81.9 & 54.3 & 31.2 & 41.9 & 74.29 & 76.69 & 0.07 \\
    Before FFN & 75.8 & 91.7 & 73.3 & \textbf{99.5} & 92.1 & 88.8 & 57.7 & 86.9 & 95.5 & 86.9 & 76.2 & 80.0 & 61.5 & 49.8 & 80.5 & 81.4 & 53.5 & 30.3 & 42.4 & 73.89 & 76.34 & 0.07 \\
    After FFN & \textbf{76.9} & 92.2 & 72.6 & \textbf{99.5} & 92.2 & 85.7 & \uline{58.0} & 87.0 & \textbf{96.2} & \textbf{96.5} & \textbf{76.7} & 81.7 & 65.4 & 50.0 & 80.9 & 82.4 & 54.4 & 29.6 & 42.6 & 74.77 & 77.48 & 0.07 \\
    \rowcolor[rgb]{ .906,  .902,  .902} Before MHA \& FFN & \uline{76.4} & \uline{93.2} & \textbf{74.1} & \textbf{99.5} & 92.2 & \textbf{91.2} & 57.9 & \uline{88.2} & \uline{96.0} & \uline{87.8} & 76.1 & 82.5 & \textbf{66.3} & \textbf{53.3} & \textbf{81.6} & \textbf{86.5} & 54.9 & \uline{32.0} & \uline{45.4} & \textbf{75.54} & \textbf{77.78} & 0.15 \\
    After MHA \& FFN & 76.2 & \textbf{93.7} & 73.6 & \textbf{99.5} & 92.2 & 88.8 & \textbf{58.3} & \textbf{88.4} & \uline{96.0} & 87.0 & \textbf{76.7} & \textbf{83.7} & 63.8 & \uline{53.2} & \uline{81.0} & \uline{85.6} & \uline{55.0} & \textbf{32.5} & 44.8 & \uline{75.25} & \uline{77.54} & 0.15 \\
    \bottomrule[2pt]
    \end{tabular}
    }
\end{table}
\begin{table}[htbp]
\renewcommand\arraystretch{1.3}
  \centering
  \vspace{-20pt}
  \caption{Ablation on the insert selection.}
  \label{tab:app_sel}
  \resizebox{0.9\columnwidth}{!}{
    \begin{tabular}{c|ccccccc|cccc|cccccccc|ccc}
    \toprule[2pt]
    & \multicolumn{7}{c|}{\textbf{Natural}}      
    & \multicolumn{4}{c|}{\textbf{Specialized}}      
    & \multicolumn{8}{c|}{\textbf{Structured}}         &  & & \\
    \midrule[1pt]
    \diagbox{\textbf{\#Layers}}{\textbf{Dataset}} & \begin{sideways}\textbf{CIFAR-100}\end{sideways} & \begin{sideways}\textbf{Caltech101}\end{sideways} & \begin{sideways}\textbf{DTD}\end{sideways} & \begin{sideways}\textbf{Flowers102}\end{sideways} & \begin{sideways}\textbf{Pets}\end{sideways} & \begin{sideways}\textbf{SVNH}\end{sideways} & \begin{sideways}\textbf{Sun397}\end{sideways} & \begin{sideways}\textbf{Camelyon}\end{sideways} & \begin{sideways}\textbf{EuroSAT}\end{sideways} & \begin{sideways}\textbf{Resisc45}\end{sideways} & \begin{sideways}\textbf{Retinopathy}\end{sideways} &  \begin{sideways}\textbf{Clevr-Count}\end{sideways} & \begin{sideways}\textbf{Clevr-Dist}\end{sideways} & \begin{sideways}\textbf{DMLab}\end{sideways} & \begin{sideways}\textbf{KITTI-Dist}\end{sideways} & \begin{sideways}\textbf{dSpr-Loc}\end{sideways} & \begin{sideways}\textbf{dSpr-Ori}\end{sideways} & \begin{sideways}\textbf{sNORB-Azim}\end{sideways} & \begin{sideways}\textbf{sNORB-Ele}\end{sideways} &  \begin{sideways}\textbf{All Mean}\end{sideways} &  \begin{sideways}\textbf{Group Mean}\end{sideways} & \begin{sideways}\textbf{Params(M)}\end{sideways} \\
    \midrule[1pt]
    2 & 73.3 & 91.7 & 71.7 & \uline{99.4} & 91.7 & 83.8 & 56.8 & 83.9 & 92.8 & 82.3 & 75.2 & 51.6 & 57.4 & 45.0 & 74.5 & 76.5 & 45.2 & 22.7 & 31.0 & 68.76 & 71.75 & 0.02 \\
    4 & 74.9 & 91.7 & 72.9 & \uline{99.4} & 91.4 & 84.8 & 56.8 & 85.6 & 94.1 & 83.2 & 75.6 & 73.1 & 62.1 & 47.9 & 80.3 & 82.5 & 48.2 & 25.9 & 37.0 & 71.97 & 74.48 & 0.05 \\
    6 & 74.8 & 92.2 & 73.5 & \uline{99.4} & 91.3 & 90.1 & \uline{57.1} & 86.5 & 95.5 & 85.3 & 75.2 & 79.0 & 63.9 & 50.5 & 80.3 & \uline{84.9} & 50.7 & 27.7 & 42.3 & 73.69 & 76.05 & 0.07 \\
    8 & 74.3 & \uline{93.0} & \uline{74.1} & \uline{99.4} & 91.7 & 89.9 & 56.2 & \uline{88.5} & 95.8 & 86.1 & 75.3 & 81.1 & 64.7 & 51.4 & 80.5 & 84.4 & 51.6 & 30.2 & 45.2 & 74.39 & 76.74 & 0.10 \\
    10 & \uline{75.0} & \uline{93.0} & \textbf{74.5} & \textbf{99.5} & \uline{91.9} & \uline{91.0} & 56.3 & \textbf{88.7} & \textbf{96.2} & \uline{87.2} & \textbf{76.8} & \uline{81.5} & \uline{65.0} & \uline{51.7} & \uline{81.4} & 83.1 & \uline{52.2} & \textbf{32.3} & \textbf{46.6} & \uline{74.94} & \uline{77.32} & 0.12 \\
    \rowcolor[rgb]{ .906,  .902,  .902} 12 & \textbf{76.4} & \textbf{93.2} & \uline{74.1} & \textbf{99.5} & \textbf{92.2} & \textbf{91.2} & \textbf{57.9} & 88.2 & \uline{96.0} & \textbf{87.8} & \uline{76.1} & \textbf{82.5} & \textbf{66.3} & \textbf{53.3} & \textbf{81.6} & \textbf{86.5} & \textbf{54.9} & \uline{32.0} & \uline{45.4} & \textbf{75.54} & \textbf{77.78} & 0.15 \\
    \bottomrule[2pt]
    \end{tabular}
    }
\end{table}
\begin{table}[htbp]
\renewcommand\arraystretch{1.3}
  \centering
  \caption{Insights into each expert's role in the inference stage. ``Group Mean'' denotes the average Top-1 accuracy of the three subgroups. ``All Mean'' denotes the average Top-1 accuracy of 19 downstream tasks. ``Increment-$i$'' signifies the retention of a sequential range of experts from the first to the $i$-th expert. ``Single-$i$'' denotes the exclusive preservation of the $i$-th expert only, while ``Sliced-$i$'' implies the exclusion of the $i$-th expert while retaining the remaining experts.}
  \label{tab:app_rep}
  \resizebox{0.9\columnwidth}{!}{
    \begin{tabular}{c|ccccccc|cccc|cccccccc|cc}
    \toprule[2pt]
    & \multicolumn{7}{c|}{\textbf{Natural}}      
    & \multicolumn{4}{c|}{\textbf{Specialized}}      
    & \multicolumn{8}{c|}{\textbf{Structured}}         &  &  \\
    \midrule[1pt]
    \diagbox{\textbf{Type}}{\textbf{Dataset}} & \begin{sideways}\textbf{CIFAR-100}\end{sideways} & \begin{sideways}\textbf{Caltech101}\end{sideways} & \begin{sideways}\textbf{DTD}\end{sideways} & \begin{sideways}\textbf{Flowers102}\end{sideways} & \begin{sideways}\textbf{Pets}\end{sideways} & \begin{sideways}\textbf{SVNH}\end{sideways} & \begin{sideways}\textbf{Sun397}\end{sideways} & \begin{sideways}\textbf{Camelyon}\end{sideways} & \begin{sideways}\textbf{EuroSAT}\end{sideways} & \begin{sideways}\textbf{Resisc45}\end{sideways} & \begin{sideways}\textbf{Retinopathy}\end{sideways} &  \begin{sideways}\textbf{Clevr-Count}\end{sideways} & \begin{sideways}\textbf{Clevr-Dist}\end{sideways} & \begin{sideways}\textbf{DMLab}\end{sideways} & \begin{sideways}\textbf{KITTI-Dist}\end{sideways} & \begin{sideways}\textbf{dSpr-Loc}\end{sideways} & \begin{sideways}\textbf{dSpr-Ori}\end{sideways} & \begin{sideways}\textbf{sNORB-Azim}\end{sideways} & \begin{sideways}\textbf{sNORB-Ele}\end{sideways} &  \begin{sideways}\textbf{All Mean}\end{sideways} &  \begin{sideways}\textbf{Group Mean}\end{sideways} \\
    \midrule[1pt]
    Increment-1 & 63.2 & 91.0 & 64.9 & 99.0 & 89.8 & 83.4 & 52.0 & 81.8 & 92.9 & 67.6 & 74.4 & 44.9 & 40.3 & 27.2 & 70.9 & 42.4 & 36.6 & 14.1 & 30.0 & 61.39 & 65.03 \\
    Increment-2 & 72.2 & \uline{93.1} & 72.0 & 99.3 & 91.4 & 89.6 & 56.4 & 87.3 & 95.6 & 84.2 & 76.0 & 74.7 & 57.3 & 48.8 & 80.6 & 75.6 & 47.7 & 25.6 & 40.2 & 71.99 & 74.71 \\
    Increment-3 & 74.5 & \textbf{93.2} & 73.3 & \textbf{99.5} & 91.9 & 90.6 & 57.5 & 87.9 & 95.4 & 86.8 & 75.7 & \uline{79.1} & 61.1 & \uline{51.9} & \uline{81.2} & 83.1 & 51.3 & 30.0 & 41.4 & 73.97 & 76.42 \\
    \rowcolor[rgb]{ .906,  .902,  .902} Increment-4 & \textbf{76.4} & \textbf{93.2} & \textbf{74.1} & \textbf{99.5} & \textbf{92.2} & \textbf{91.2} & \textbf{57.9} & \textbf{88.2} & \textbf{96.0} & \textbf{87.8} & \uline{76.1} & \textbf{82.5} & \textbf{66.3} & \textbf{53.3} & \textbf{81.6} & \textbf{86.5} & \textbf{54.9} & \textbf{32.0} & \textbf{45.4} & \textbf{75.54} & \textbf{77.78} \\
    \midrule
    Single-1 & 63.2 & 91.0 & 64.9 & 99.0 & 89.8 & 83.4 & 52.0 & 81.8 & 92.9 & 67.6 & 74.4 & 44.9 & 40.3 & 27.2 & 70.9 & 42.4 & 36.6 & 14.1 & 30.0 & 61.39 & 65.03 \\
    Single-2 & 63.5 & 89.3 & 65.1 & 99.2 & 90.2 & 81.2 & 51.6 & 79.1 & 92.3 & 67.3 & 74.0 & 38.5 & 46.9 & 27.1 & 65.3 & 34.5 & 20.3 & 9.3 & 25.7 & 58.96 & 62.92 \\
    Single-3 & 58.0 & 88.7 & 62.3 & 99.2 & 90.0 & 82.0 & 53.9 & 83.9 & 92.0 & 66.2 & 69.1 & 46.2 & 20.8 & 25.7 & 68.4 & 39.0 & 23.9 & 11.9 & 30.4 & 58.51 & 62.47 \\
    Single-4 & 57.9 & 90.1 & 65.0 & 98.8 & 88.3 & 81.1 & 53.5 & 84.9 & 90.8 & 64.7 & 73.4 & 44.1 & 39.8 & 29.0 & 67.5 & 43.4 & 20.7 & 10.3 & 30.9 & 59.70 & 63.52 \\
    \midrule
    Sliced-1 & 74.7 & 92.9 & 73.0 & \textbf{99.5} & 91.5 & 90.3 & \uline{57.6} & 87.5 & \uline{95.7} & \uline{86.9} & 75.4 & 70.5 & 62.1 & 51.5 & 80.9 & 83.9 & \uline{52.6} & 28.9 & 40.8 & 73.48 & 76.02 \\
    Sliced-2 & \uline{75.3} & \uline{93.1} & 73.1 & 99.3 & 91.7 & \uline{90.7} & \uline{57.6} & 87.9 & 95.3 & 86.4 & 76.0 & 72.2 & \uline{64.7} & 51.6 & 81.0 & 83.3 & 52.3 & 30.1 & 43.8 & 73.98 & 76.42 \\
    Sliced-3 & 75.0 & \uline{93.1} & \uline{73.6} & \uline{99.4} & \uline{92.1} & 90.6 & 57.5 & \uline{88.1} & 95.5 & 86.5 & \textbf{76.3} & 78.8 & 54.8 & 51.8 & 80.9 & \uline{84.8} & 51.9 & \uline{30.6} & \uline{44.7} & \uline{73.99} & \uline{76.47} \\
    Sliced-4 & 74.5 & \textbf{93.2} & 73.3 & \textbf{99.5} & 91.9 & 90.6 & 57.5 & 87.9 & 95.4 & 86.8 & 75.7 & \uline{79.1} & 61.1 & \uline{51.9} & \uline{81.2} & 83.1 & 51.3 & 30.0 & 41.4 & 73.97 & 76.42 \\
    \bottomrule[2pt]
    \end{tabular}
    }
\end{table}

\clearpage
\subsection{Video Classification and Semantic Segmentation}
\label{app:other_tasks} 
\textbf{}{Video Classification.} Given the constraints in computational resources and time, we follow the experimental setup outlined in AdaptFormer~\cite{adaptformer} and select the relatively smaller HMDB51~\cite{hmdb51} video classification dataset for our evaluation. The HMDB51 dataset comprises 3.5k training samples and 1.5k validation samples, spanning 51 distinct classes. We report the Top-1 accuracy on the validation set, utilizing the ViT-B/16~\cite{vit} pre-trained on MAE-Video~\cite{videomae}, as specified in~\cite{adaptformer}, with ALoRE employing the optimal configuration presented in the paper. We did not conduct a hyperparameter search.
\begin{table}[h]
    \centering
    \caption{Experimental results on the video classification task.}
    \begin{tabular}{c|c|c}
        \toprule
         Method & Accuracy & Params (M) \\
         \midrule
         Full fine-tuning & 46.41 & 86.04 \\
         Linear probing & 49.84 & 0.07 \\
         VPT~\cite{vpt} & 52.67 & 0.08 \\
         AdaptFormer~\cite{adaptformer} & 51.81 & 0.15 \\
         RepAdapter~\cite{rep} & 55.67 & 0.15 \\
         \rowcolor[rgb]{ .906,  .902,  .902} ALoRE (ours) & 57.76 & 0.15 \\
         \bottomrule
    \end{tabular}
    \label{tab:video}
\end{table}
\vspace{-10pt}

\paragraph{Semantic Segmentation.} We select the ADE20K~\cite{ade20k} dataset as our benchmark, adhering to the conventional practice of utilizing the 20k/2k split for the training and validation sets, respectively. We report the mIoU-SS and mIoU-MS scores as the primary evaluation metrics. For the model architecture, we have employed ViT-L/14~\cite{vit} and SETR~\cite{setr}, which are well-established and widely adopted frameworks in semantic segmentation. Given the complexity of the task at hand, we have opted to increase the dimensionality of the ALoRE component to 8.
\begin{table}[h]
    \centering
    \caption{Experimental results on the semantic segmentation task.}
    \begin{tabular}{c|c|c|c}
        \toprule
         Method & mIoU-SS & mIoU-MS & Params (M) \\
         \midrule
         Full fine-tuning & 48.31 & 50.07 & 318.31 \\
         Head Only & 35.12 & 37.46 & 13.18 \\
         BitFit~\cite{bitfit} & 43.40 & 45.33 & 13.46 \\
         VPT~\cite{vpt} & 42.11 & 44.06 & 13.43 \\
         RepAdapter~\cite{rep} & 44.44 & 46.71 & 13.82 \\
         \rowcolor[rgb]{ .906,  .902,  .902} ALoRE (ours) & 46.09 & 48.14 & 13.83 \\
         \bottomrule
    \end{tabular}
    \label{tab:semantic_seg}
\end{table}
\vspace{-10pt}

\subsection{Memory and Time cost of Training and Inference}
\label{app:cost}
\begin{figure*}[h]
  \centering
  \subfloat[Training time (batch/s)]{\includegraphics[width=0.24\textwidth]{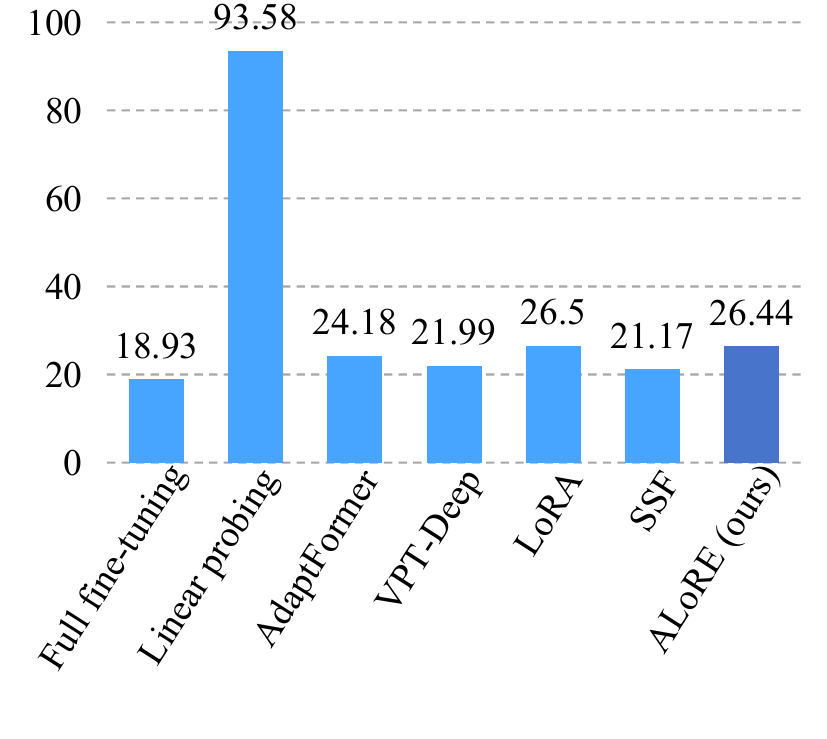}\vspace{-8pt}}\hfill
  \subfloat[Training memory (G)]{\includegraphics[width=0.24\textwidth]{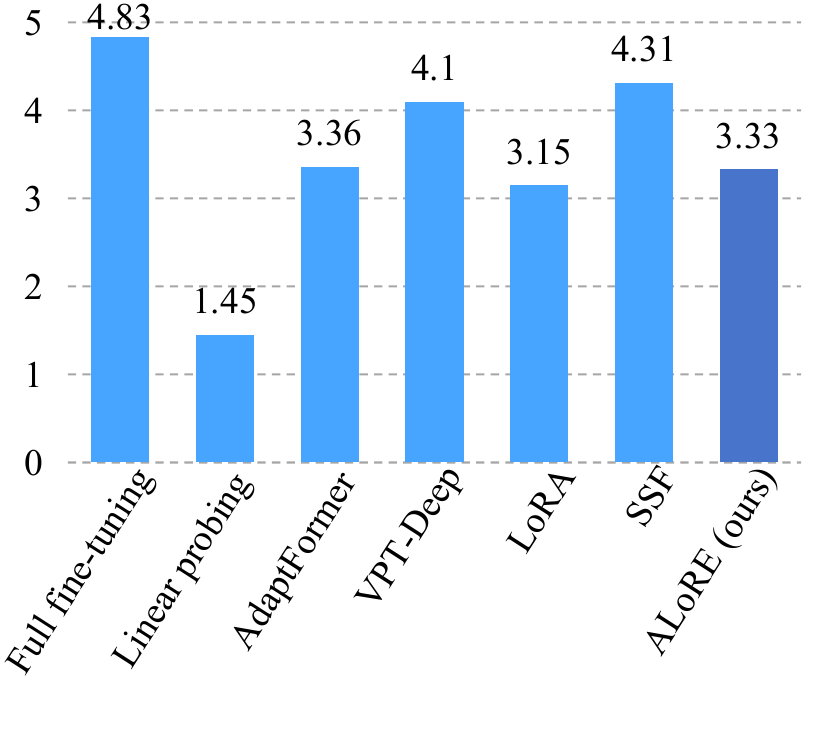}\vspace{-8pt}}\hfill
  \subfloat[Test time (batch/s)]{\includegraphics[width=0.24\textwidth]{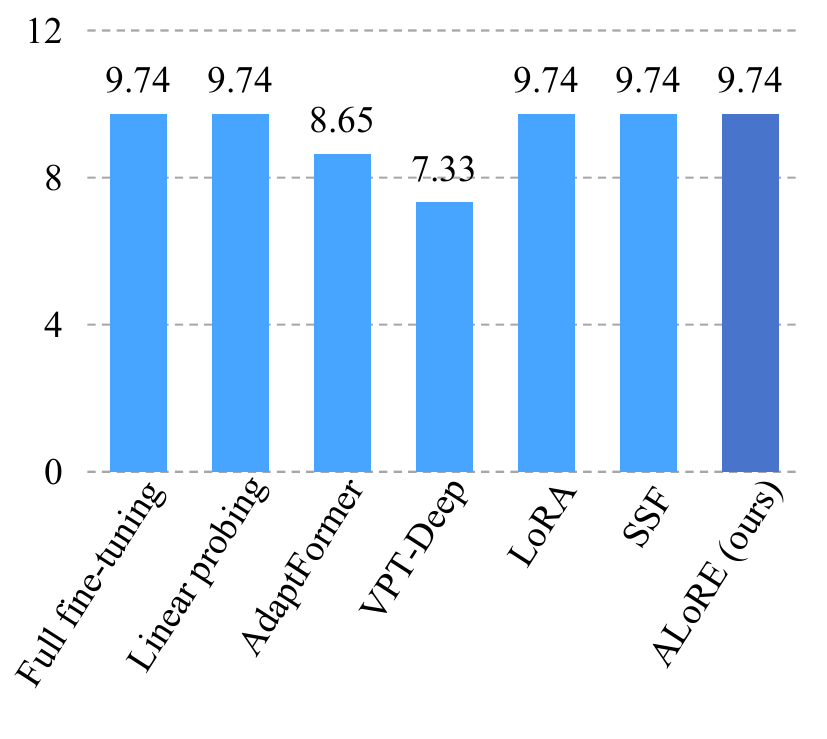}\vspace{-8pt}}\hfill
  \subfloat[Test memory (G)]{\includegraphics[width=0.24\textwidth]{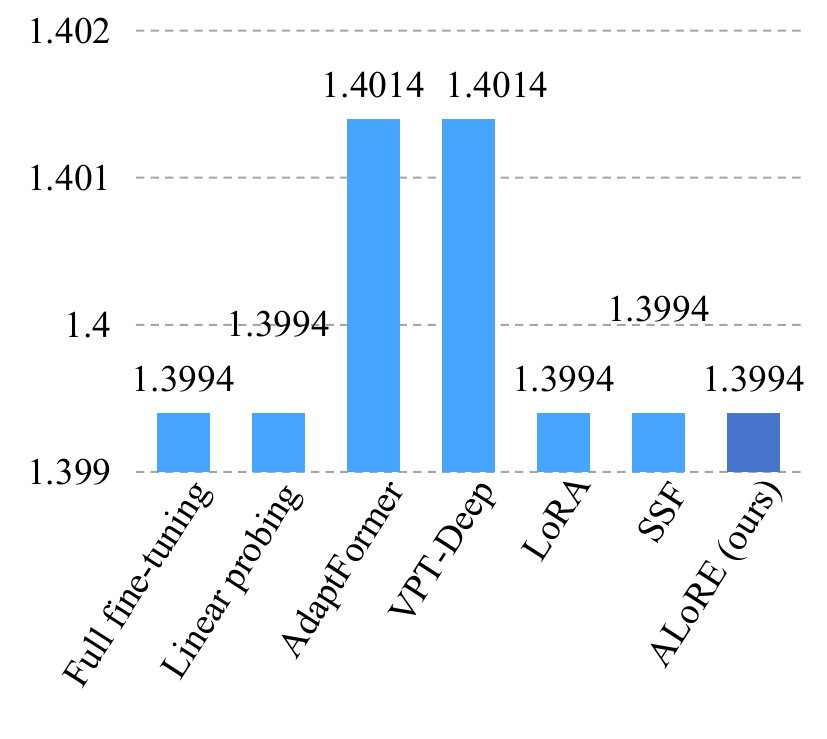}\vspace{-8pt}}
  \vspace{-6pt}
  \caption{Computation cost of fine-tuning methods, measured on Tesla P40 with a batch size of 16.}
  \label{fig:computation cost}
  \vspace{-12pt}
\end{figure*}

\paragraph{Compuation Cost.} To demonstrate the computational efficiency of our ALoRE method and highlight its advantages over other fine-tuning techniques, we compare the training and inference speeds as well as memory consumption of different methods in Figure \ref{fig:computation cost}. All results are obtained using a single Tesla P40 GPU with a batch size of 16. Our findings indicate that ALoRE exhibits comparable training cost with LoRA~\cite{lora} and faster training speed and lower memory consumption compared to SSF~\cite{ssf}, and it introduces no additional inference latency compared to VPT~\cite{vpt} and other Adapter-based.

\clearpage

\section{Visualization}
\label{app:vis}

\subsection{Attention Map}
\label{app:attn}

\begin{figure}[htbp]
    \centering
    \includegraphics[width=0.9\textwidth]{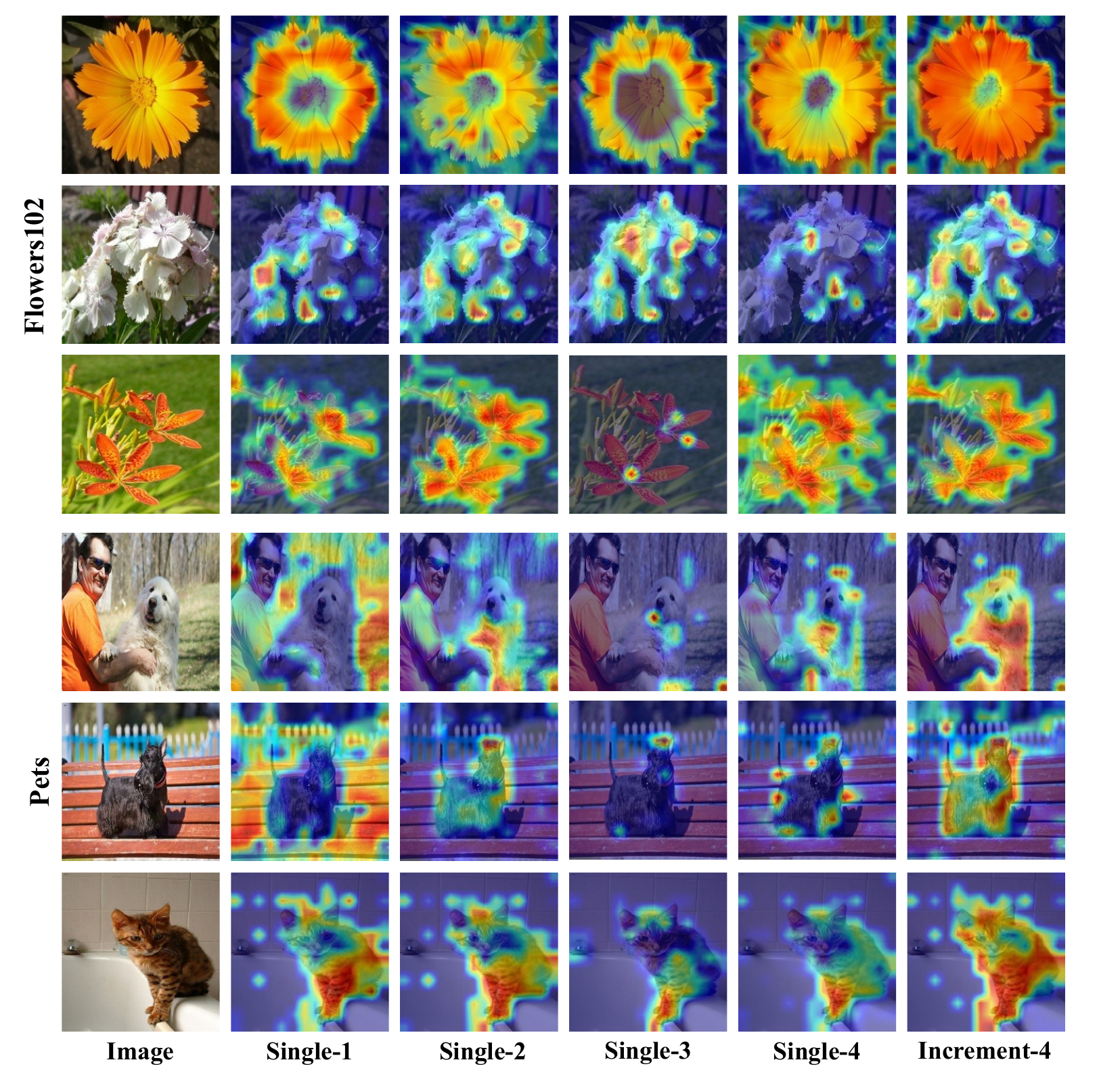}
    \caption{More visualization of attention maps with respect to different experts. ``Single-$i$`` denotes the exclusive preservation of the $i$-th expert only. ``Increment-4`` represents the aggregation of all experts.}
    \label{fig:gradcam_sup}
\end{figure}
In order to gain a more intuitive understanding of the strategy of aggregating multiple experts utilized in our proposed ALoRE method, as well as to comprehend the individual contributions of each expert in visual adaptation tasks, we employ visualizations of attention maps for each expert, as depicted in Figure~\ref{fig:gradcam_sup}. By examining Figure~\ref{fig:gradcam_sup}, we observe that different experts tend to focus on distinct edge regions and primary object parts within the visual entities across various tasks. This observation aligns with the cognitive process of humans when encountering a novel object category, wherein different levels of visual perception are attuned to different conceptual aspects. Notably, the last column in Figure~\ref{fig:gradcam_sup} represents the attention map obtained by aggregating all the experts's cognitive patterns. We can deduce from this result that the strategy of combining multiple experts allows for the extraction of diverse visual representation patterns acquired by each expert. By distributing these distinct representation patterns among different experts for learning, we achieve a decoupling of features, consequently enhancing the performance of visual adaptation tasks.

\clearpage

\section{Limitation and Broader Impacts}\label{app:limit_and_impact}

\subsection{Limitation}\label{app:limit}
Our main contribution lies in proposing the construction of a hypercomplex Parameterized space through the Kronecker product, enabling the aggregation of low rank experts in a multi-branch architecture. This multi-branch structure design facilitates the decoupling of learned feature patterns. However, the feasibility of employing alternative structures for individual experts has not been investigated. An interesting avenue for exploration is the integration of our multi-branch structure approach into multi-task learning. Investigating the applicability of this methodology in the context of multiple tasks holds significant potential.

\subsection{Broader Impacts}\label{app:impact}
Our research has far-reaching societal implications that warrant further exploration. By introducing the ALoRE method, we address the challenges posed by full fine-tuning and non-reparameterizable PETL approaches, thereby offering a more efficient solution for parameter savings during both the training and inference stages. This efficiency not only accelerates the transfer of large-scale pre-trained models to downstream tasks but also significantly reduces the computational resources required, leading to substantial cost savings and a reduction in carbon emissions.

The ability to quickly adapt pre-trained models to new tasks without the need for extensive architectural modifications is a key advantage of ALoRE. Through linear transformation and re-parameterization techniques, we eliminate the necessity of altering the deployed backbone architecture. Instead, only a set of weights needs to be replaced, making the entire process more streamlined and convenient compared to methods that introduce additional parameters, such as the widely used VPT~\cite{vpt} approach.

However, it is important to acknowledge the potential ethical implications associated with the upstream pre-training process. Pre-training large models typically relies on massive datasets, and if these datasets contain illegal or biased data, it may inadvertently perpetuate unfair biases and discriminatory patterns in the fine-tuned models. Thus, careful considerations must be taken to ensure that the pre-training data adheres to ethical guidelines and promotes fairness and inclusivity.

In summary, our work not only provides a novel solution for parameter savings and efficient model adaptation but also highlights the importance of ethical considerations in the pre-training process. The implications extend beyond the realm of visual tasks, offering potential applications in various domains and paving the way for future advancements in multi-task learning and complex task adaptation.



\end{document}